\documentclass[11pt]{article}

\usepackage[preprint]{acl}
\usepackage{amsmath}
\usepackage{times}
\usepackage{latexsym}

\usepackage[T1]{fontenc}

\usepackage[utf8]{inputenc}

\usepackage{microtype}

\usepackage{inconsolata}

\usepackage{graphicx}

\usepackage{nicefrac}
\usepackage{amsfonts}
\usepackage{booktabs}
\usepackage{float}
\usepackage{placeins}
\usepackage{changepage}
\usepackage[table]{xcolor}
\usepackage{tabularx}
\usepackage{pdflscape}
\usepackage{makecell}
\usepackage{colortbl}
\usepackage{subcaption}

\title{MVEB: Massive Video Embedding Benchmark}

\author{
\textbf{Adnan El Assadi\textsuperscript{1}},
\textbf{Roman Solomatin\textsuperscript{2,3}},
\textbf{Isaac Chung\textsuperscript{4}},
\textbf{Chenghao Xiao\textsuperscript{5}},
\textbf{Deep Shah\textsuperscript{6}},
\textbf{Manan Dey\textsuperscript{7}},
\\
\textbf{Shriya Sudhakar\textsuperscript{8}},
\textbf{Zacharie Bugaud\textsuperscript{9}},
\textbf{Wissam Siblini\textsuperscript{10}},
\textbf{Ayush Sunil Munot\textsuperscript{11}},
\textbf{Yashwanth Devavarapu\textsuperscript{12}},
\\
\textbf{Rakshitha Ireddi\textsuperscript{12}},
\textbf{Michelle Yang\textsuperscript{10}},
\textbf{Márton Kardos\textsuperscript{13}},
\textbf{Niklas Muennighoff\textsuperscript{14}},
\textbf{Kenneth Enevoldsen\textsuperscript{13}},
\\
\\
\textsuperscript{1}Harvard University,
\textsuperscript{2}SaluteDevices,
\textsuperscript{3}MIRAI,
\textsuperscript{4}Zendesk,
\textsuperscript{5}Shanghai University of Finance and Economics,
\\
\textsuperscript{6}Google LLC,
\textsuperscript{7}Salesforce,
\textsuperscript{8}Cornell University,
\textsuperscript{9}Astera Institute,
\textsuperscript{10}Independent Contributor,
\\
\textsuperscript{11}Indian Institute of Technology, Kharagpur,
\textsuperscript{12}Barclays,
\textsuperscript{13}Aarhus University,
\textsuperscript{14}Stanford University
\\
 \small{
   \textbf{Correspondence:} \href{mailto:adnanassadi56@gmail.com}{adnanassadi56@gmail.com}
 }
}

\begin{document}
\maketitle

\begin{abstract}

We introduce the \textbf{M}assive \textbf{V}ideo \textbf{E}mbedding \textbf{B}enchmark (MVEB), a 23-task benchmark for video embeddings spanning classification, zero-shot classification, clustering, pair classification, retrieval, and video-centric question answering. We evaluate 33 models and find that no single model dominates: MLLM-based embeddings lead on classification, clustering, pair classification, and QA; multimodal binding leads on retrieval and zero-shot classification; generative MLLMs without contrastive adaptation collapse on cross-modal tasks. Paired video-only vs.\ audio+video evaluations show that audio's contribution depends on dataset annotation provenance: audio helps when labels were produced from both modalities and hurts when they were produced from visuals alone, a six-point gap consistent across model families. MVEB is derived from MVEB+, a 184-task pool, and is designed to maintain task diversity while reducing evaluation cost. It integrates into the MTEB ecosystem for unified evaluation across text, image, audio, and video. We release MVEB and all 184 tasks along with code and a leaderboard at 
\url{https://github.com/embeddings-benchmark/mteb}.

\end{abstract}

\section{Introduction}

Video representations support a wide range of downstream tasks such as action recognition, text-video retrieval, and question answering. Most existing video benchmarks evaluate models on only one such capability, making it difficult to tell whether a high score reflects general-purpose video representation quality or a model that overfits to one task. The result is a fragmented landscape in which models with strong action-recognition scores routinely under-perform on retrieval, and vice versa, with no standard protocol for measuring this gap.

\begin{figure*}[t]
    \centering
    \includegraphics[width=\linewidth]{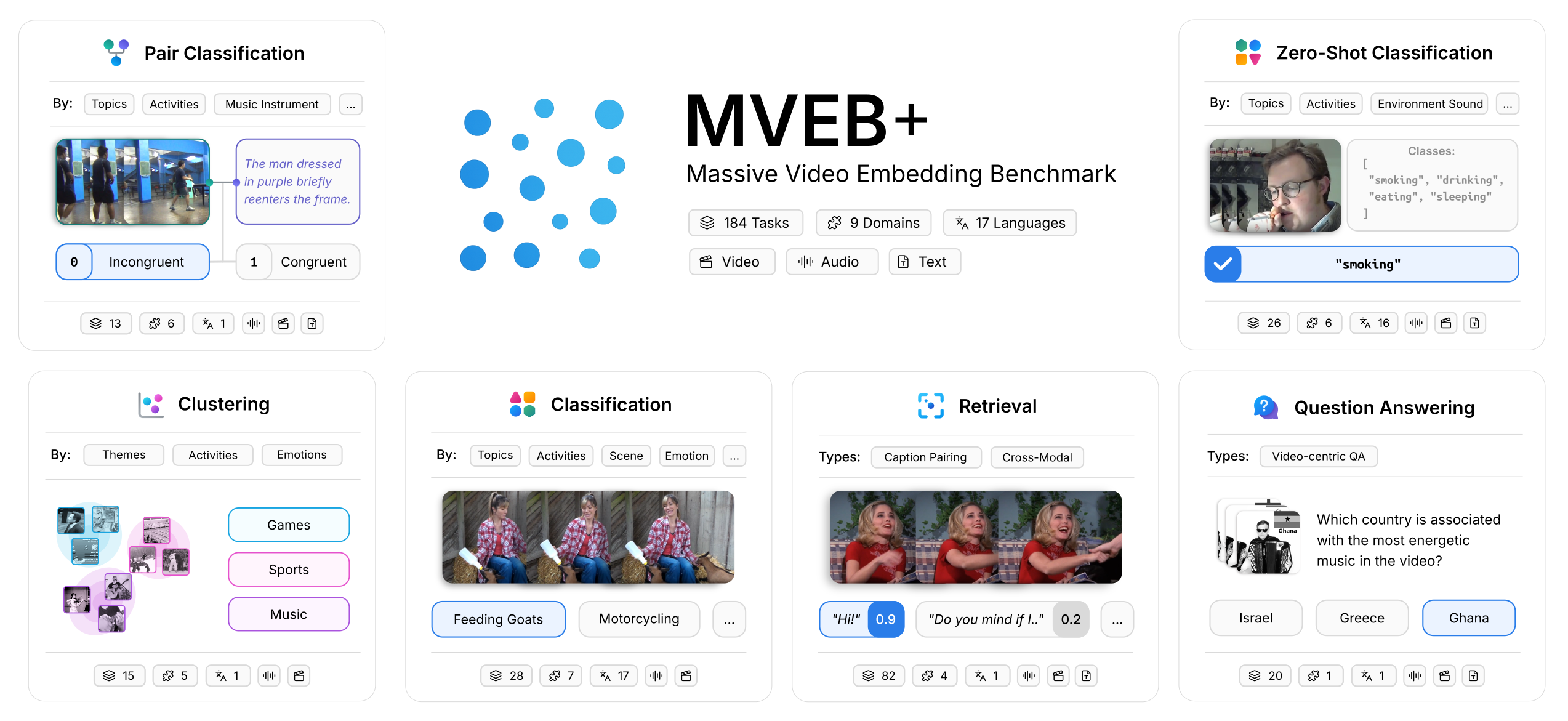}
    \caption{Overview of task types and example subtypes in MVEB+.}
    \label{fig:overview}
\end{figure*}

We introduce the \textbf{M}assive \textbf{V}ideo \textbf{E}mbedding \textbf{B}enchmark (MVEB), the video sibling in the MTEB~\cite{muennighoff2022mteb} family of embedding benchmarks alongside its multilingual extension MMTEB~\cite{enevoldsen2025mmteb}, MIEB~\cite{xiao2025mieb} for image, and MAEB~\cite{assadi2026maeb} for audio. Like its predecessors, MVEB measures general-purpose embedding quality zero-shot, with no per-task fine-tuning. MVEB systematically pairs every audio-bearing dataset with both video-only and video$+$audio variants, enabling direct measurement of when audio improves video understanding, a setting prior video benchmarks (see \S\ref{sec:related-work}) do not support.

MVEB itself is a curated 23-task subset of MVEB+ (a 184-task pool) selected by correlation-based redundancy pruning. MVEB groups tasks into six families: classification, zero-shot classification, clustering, pair classification, retrieval (over eight modality directions), and video-centric question answering. Models that cannot accept the full audio+video+text input surface are evaluated on modality-restricted leaderboards drawn from the same pool: MVEB(text, video) (19 tasks) for text-video encoders without an audio encoder, and MVEB(video) (9 tasks) for video-only encoders.

Evaluating 33 models across six embedding paradigms, we find that no single model dominates: MLLM-based embedding leads on classification, clustering, pair classification, and QA; multimodal binding leads on retrieval and zero-shot classification; and generative MLLMs used as embedders without contrastive adaptation collapse on cross-modal tasks. Two findings stand out: (i) The audio track of a video is a distinct signal whose contribution depends on the dataset's annotation provenance. Audio helps on AV-grounded datasets (labels produced from both modalities) and hurts on V-grounded ones (labels produced from visuals alone), a six-point gap consistent across task types and model families (\S\ref{sec:audio-contribution}); (ii) The modality-restricted MVEB(text, video) leaderboard reorders the top of the field: a text-video embedder without an audio path (Qwen3-VL-Embedding-8B) takes rank~1 ahead of every audio+video+text model in our roster, so the two leaderboards reward distinct model families.

To summarize, MVEB makes the following key contributions:
\begin{enumerate}
    \item A 23-task benchmark (MVEB) for video embedding models, curated from a 184-task pool (MVEB+) that combines video with text and audio modalities, with evaluations of 33 models across six embedding paradigms. We also expose MVEB(text, video) for models without an audio encoder and MVEB(video) for video-only encoders, drawn from the same pool.
    \item Actionable findings for general-purpose video embedder development: a contrastive embedding stage is a near-prerequisite for cross-modal performance, and training-data alignment matters more than backbone scale for cross-modal tasks.
    \item Integration into the MTEB ecosystem with task and model versioning, enabling community-driven contribution and long-term maintenance.
\end{enumerate}

\section{Related Work}
\label{sec:related-work}

 \textbf{Embedding Benchmarks} MTEB~\cite{muennighoff2022mteb} and its multilingual, image, and audio extensions MMTEB~\cite{enevoldsen2025mmteb}, MIEB~\cite{xiao2025mieb}, and MAEB~\cite{assadi2026maeb} have driven progress in representation learning by providing broad, regularly-maintained leaderboards with consistent protocols; MVEB completes the family for video.

 \textbf{Video Representation Benchmarks} Prior video benchmarks each cover only a slice of the surface MVEB exercises: VideoEval~\cite{li2024videoeval} (20 models, 12 tasks) is visual-only; UVRB~\cite{guo2025universalvideoretrievalgeneralizing} (16 datasets) is retrieval-only; LoVR~\cite{wu2025lovr} (467 videos) is a single long-video retrieval task; VidVec~\cite{vidvec2026} is a method paper that evaluates on 5 retrieval benchmarks. Most closely related, MMEB-V2~\cite{meng2025vlm2vecv2} extends MMEB~\cite{jiang2024vlm2vec} to video across four categories (Video Retrieval, Moment Retrieval, Video Classification, and Video QA), and MMEB-V3~\cite{huang2026mmebv3measuringperformancegaps} adds audio as a separate modality with 11 audio tasks across three categories (classification, retrieval, temporal grounding) and cross-modal retrieval directions between audio and each of text, image, and video. MVEB differs in three concrete ways: (i)~scale on video: 184 tasks in MVEB+, roughly an order of magnitude more than the video task counts in the closest prior efforts; (ii)~audio-visual coupling: MVEB systematically pairs every audio-bearing dataset with both video-only and video$+$audio variants on the same task and feeds both streams jointly to the model, whereas MMEB-V3 treats audio strictly as a separate modality with cross-modal retrieval directions and never as joint audio$+$video input; (iii)~maintenance: MVEB inherits MTEB's community-driven contribution and continuous-update model rather than shipping as a static leaderboard, avoiding the stagnation that has affected most prior video benchmarks.
\section{Benchmark Construction}
\label{sec:benchmark-construction}

MVEB is built on the MTEB ecosystem~\citep{muennighoff2022mteb}, extending its evaluation framework to video alongside the multilingual~\citep{enevoldsen2025mmteb}, image~\citep{xiao2025mieb}, and audio~\citep{assadi2026maeb} extensions. We inherit MTEB's standardized metrics, minimal task/model interface, versioned artifacts, and community-driven maintenance~\citep{chung2025maintainingmteblongterm}. MVEB evaluates \textit{embedding quality}, not transcription or generation.

\textbf{Dataset Selection:}
We curate datasets according to four guiding principles: (1) \textit{domain diversity} across action recognition, social media understanding, emotion recognition, music, scene understanding, and instructional content; (2) \textit{task diversity} across the six families described in \S\ref{sec:tasks-eval}; (3) \textit{modality coverage} across video-only, video+text, and video+audio task variants; and (4) \textit{provenance and accessibility}, prioritizing datasets with established usage, clear licensing, and public availability.

\textbf{Task Selection:}
Evaluating models across the full 184-task pool (MVEB+) is prohibitively expensive for most groups. Following MMTEB, MIEB, and MAEB, which demonstrated that principled filtering maintains high rank correlation with exhaustive evaluation, we construct the curated MVEB leaderboard using five selection criteria applied to MVEB+: (1) \textit{Validity}: for directional tasks we prioritize the more semantically valid direction (e.g., text-to-video over video-to-text for retrieval); (2) \textit{Unique coverage}: tasks providing exclusive coverage of a domain or capability are retained regardless of other factors (e.g., the only emotion-clustering task or the only video+audio retrieval direction); (3) \textit{Linguistic breadth}: among comparable tasks, we retain those covering more languages; (4) \textit{Redundancy removal}: we compute pairwise correlation matrices across model rankings and remove tasks with Spearman $\rho > 0.85$ to a retained task, keeping the task with broader coverage or lower runtime; (5) \textit{Runtime efficiency}: among otherwise equivalent tasks, we select those with lower computational cost on video, where frame decoding and per-frame encoding dominate runtime.

Applying these filters produces the 23-task MVEB benchmark. Across representative models MVEB runs roughly $7$--$10\times$ faster than MVEB+ (\autoref{tab:benchmark-runtime}) while maintaining strong correlation in model scores (Pearson $r$=0.996) and ranking (Spearman $\rho$=0.944), so curation preserves relative model performance at a fraction of the runtime. 

\begin{table}[!ht]
    \centering
    \scriptsize
    \setlength{\tabcolsep}{4pt}
    \renewcommand{\arraystretch}{1.05}
    \begin{tabular}{lcccc}
    \toprule
    \textbf{Model} & \textbf{Params} & \textbf{MVEB+} & \textbf{MVEB} & \textbf{Speedup} \\
    \midrule
    LCO-Embedding-Omni-7B & 7B    & 302.7 & 31.9 & $9.5\times$ \\
    Qwen2.5-Omni-7B       & 7B    & 267.3 & 29.4 & $9.1\times$ \\
    pe-av-small           & 847M  & 159.7 & 16.7 & $9.6\times$ \\
    ebind-audio-vision    & 764M  & 122.7 & 16.8 & $7.3\times$ \\
    \bottomrule
    \end{tabular}
    \caption{GPU runtime (hours) on MVEB vs.\ MVEB+, measured on a single NVIDIA H100.}
    \label{tab:benchmark-runtime}
\end{table}

Alongside MVEB we expose two leaderboards drawn from the same task pool so that modality-restricted models can also be evaluated: \textbf{MVEB(text, video)} (19 tasks) for text-video models without an audio encoder, and \textbf{MVEB(video)} (9 tasks) for video-only encoders. \autoref{tab:benchmark-variants} summarises the four artifacts. The full dataset list is in \autoref{sec:overview}.

\begin{table}[!th]
    \centering
    \scriptsize
    \setlength{\tabcolsep}{4pt}
    \renewcommand{\arraystretch}{1.05}
    \begin{tabular}{lcll}
    \toprule
    \textbf{Artifact} & \textbf{Tasks} & \textbf{Modalities} & \textbf{Where} \\
    \midrule
    MVEB+             & 184 & a, t, v   & appdx, Tab.~\ref{tab:mveb-extended-results} \\
    MVEB              & 23  & a, t, v   & body, Tab.~\ref{tab:mveb-main-results} \\
    MVEB(text, video) & 19  & t, v      & appdx, Tab.~\ref{tab:mveb-text-video-results} \\
    MVEB(video)       & 9   & v         & appdx, Tab.~\ref{tab:mveb-video-results} \\
    \bottomrule
    \end{tabular}
    \caption{\textbf{MVEB+} is the 184-task collection. \textbf{MVEB} is the 23-task curated benchmark we release. \textbf{MVEB(text, video)} and \textbf{MVEB(video)} are leaderboards drawn from the same pool that let modality-restricted models also be evaluated. Modalities: a=audio, t=text, v=video.}
    \label{tab:benchmark-variants}
\end{table}

\textbf{Data Contamination: } We cross-reference each model's declared training data against MVEB tasks; per-model overlap is in \autoref{appdx:contamination}.

\textbf{Benchmark Ranking: }
Following MMTEB~\cite{enevoldsen2025mmteb}, we compute model ranks using a Borda count~\cite{colombo2022what} by treating each task as a preference voter over models. Since Borda is not a continuous measure, we report both the Borda rank and the mean in the leaderboard.

\textbf{Reproducibility and Benchmark Accessibility: }
Building on prior reproducibility efforts~\citep{enevoldsen2025mmteb, enevoldsen2024scandinavianembeddingbenchmarkscomprehensive, chung2025maintainingmteblongterm}, we extend MTEB with per-subset versioning, named experiment scoping, and richer model metadata; see Appendix~\ref{appendix:reproducibility}.

\subsection{Models}
\label{sec:methodology-models}

We evaluate 33 publicly-available checkpoints spanning the major paradigms for video embeddings, ranging from $\sim$200M to 10.7B parameters and covering the modality combinations MVEB exercises. The paradigms below correspond to the \texttt{Type} column in the leaderboard; full per-model details are in \autoref{appdx:models}.

\textbf{Self-supervised Video Encoders} learn from video alone, without paired text or audio. We evaluate the V-JEPA-2 family~\citep{assran2025vjepa2}; these models contribute only to MVEB(video).

\textbf{Video-Text Contrastive Encoders} learn a joint video-text space through contrastive pretraining over web-scale video-caption pairs. We evaluate the X-CLIP family~\citep{ni2022expanding}.

\textbf{Audio-Visual Contrastive Encoders} jointly align video, audio, and text in a shared embedding space, enabling cross-modal retrieval that involves the audio track of the video. We evaluate the Perception Encoder audio-visual family~\citep{vyas2025pushingfrontieraudiovisualperception}.

\textbf{Multimodal Binding Models} learn a single embedding space across modalities by aligning each to a shared anchor, in the style of ImageBind. We evaluate the eBind family~\citep{broadbent2025ebindpracticalapproachspace}.

\textbf{MLLM-Based Embedding Models} take a multimodal large language model backbone and adapt it into an embedding model through a dedicated contrastive or retrieval-objective training stage. This is the largest family in our roster and reflects the recent shift toward MLLM-backed embeddings observed in MIEB and MMEB-V2~\citep{xiao2025mieb}. We evaluate LCO-Embedding-Omni~\citep{xiaoscaling}, e5-omni~\citep{chen2026e5omni}, Tevatron OmniEmbed~\citep{zhuang2025tevatron}, BidirLM-Omni-Embedding~\citep{boizard2026bidirlmtextomnimodalbidirectional}, UME-R1~\citep{lan2025ume}, OmniEmbed-Nemotron~\citep{xu2025omni}, Qwen3-VL-Embedding~\citep{qwen3vlembedding}, VLM2Vec-V2.0~\citep{meng2025vlm2vecv2}, and Jina-Embeddings-v5-omni~\citep{akram2026jinaembeddingsv5texttasktargetedembeddingdistillation}.

\textbf{Generative MLLMs used as embedders} are multimodal large language models trained for generation rather than embedding; we obtain a representation by pooling the final-layer hidden states using the model's default pooling strategy. Including them lets us measure how far a generation-only training objective gets on embedding tasks, without any embedding-specific adaptation. We evaluate Qwen2.5-Omni~\citep{xu2025qwen25omnitechnicalreport}.

\subsection{Evaluation Protocol}
\label{sec:eval-protocol}

\paragraph{Video sampling.}
Video encoders split into two groups. Variable-length models (the omni MLLM-embedding family, the Qwen3-VL embedders, and the variable-length Perception Encoder checkpoints) sample at \texttt{fps=2} with a hard cap of \texttt{max\_frames=64} to bound peak GPU memory on long clips. Fixed-length models (X-CLIP, eBind, V-JEPA-2) sample the exact frame count their training pipeline expects; full per-model values are in \autoref{tab:sampling-config}. We honor each model's declared configuration on every task rather than forcing a benchmark-wide setting because the latter would push half the roster out of distribution. The effect of \textit{deliberately} varying the budget at test time is studied separately in \autoref{analyses_test_time_scaling_frames}.

\paragraph{Audio sampling.} Each model also declares its required \texttt{target\_sampling\_rate}, mono-conversion, and an optional duration cap (per-model values in \autoref{tab:sampling-config}). We apply a per-model cap rather than a benchmark-wide one because feature extractors differ: caps range from 30 s (PE-AV, the regime it was trained under) to no explicit cap (LCO-Embedding-Omni, which delegates truncation to its own processor). The shared audio collator resamples, mono-converts, and truncates accordingly.

\paragraph{Other protocol details.}
All models are evaluated zero-shot: we use each model's default embedding output, without fine-tuning or any layer-specific or pooling-specific tuning. The shared MTEB task evaluator consumes these embeddings and applies the per-task metric defined in \autoref{sec:tasks-eval}. Some models are instruction-tuned and expect a natural-language instruction prepended to each input (e.g., ``Represent this video for retrieval:''); others take raw inputs directly. We pass task-specific instructions only to models trained to use them, so each model sees the input format it was trained under.

\subsection{Modalities}
\label{sec:modalities}

For every dataset in MVEB whose source video contains an audio track, we extract that audio and run the task twice: once with video only, and once with both video and audio fed to the model. This gives us paired evaluations on the same underlying clips and lets us measure the contribution of the audio track to video understanding (per-task \texttt{video} vs.\ \texttt{audio + video} columns in the appendix tables; see \autoref{appdx:results}). We expose both variants rather than scoring models only on \texttt{va} because, as \S\ref{sec:audio-contribution} shows, on V-grounded datasets the audio channel is a confound rather than a fair test of embedding quality. Datasets whose source has no usable audio (e.g., HMDB51, SSv2, Diving48, Breakfast) cannot be paired and enter the benchmark in their video-only form only; they are still included in MVEB and MVEB+ where applicable, just without an \texttt{audio + video} counterpart. This dual-variant scheme applies to classification, zero-shot classification, clustering, pair classification, and QA; retrieval is handled separately via the eight directions listed in \S\ref{sec:tasks-eval}, since each direction's query and target already carry their own modality.

Each model runs only on the task variants compatible with its declared modalities; audio-bearing variants and audio retrieval directions are skipped for text-video-only models.



\subsection{Tasks and Evaluation}
\label{sec:tasks-eval}

We follow the task structure of MTEB, MIEB, and MAEB, adapting each task family to video inputs and adding cross-modal directions specific to the video setting.

\textbf{Classification} A logistic regression is trained on frozen video embeddings to predict labels~\citep{alain2018understandingintermediatelayersusing,radford2021learning}. We use few-shot linear probing~\citep{muennighoff2022mteb,cherti2023reproducible} with 8 examples per class, balancing evaluation quality with computational efficiency. Accuracy is the main metric.

\textbf{Zero-shot Classification} Video embeddings are matched against text embeddings of class-label prompts (e.g., ``a video of \{label\}'') without training a classifier. Prompt templates are hand-crafted per dataset. We measure accuracy following~\citet{radford2021learning}. This task is restricted to models that declare a joint video-text embedding space.

\textbf{Clustering} We use MiniBatchKMeans (with $k$ set to the number of true labels) over video embeddings and report V-measure~\citep{rosenberg-hirschberg-2007-v} as the main metric, evaluating whether embeddings group meaningfully according to semantic categories without any supervised signal.

\textbf{Retrieval} Retrieval evaluates ranking a corpus by relevance to a query under cosine similarity. We support eight retrieval directions across the modality combinations a video benchmark needs. Writing T for text, A for audio, V for video (combined codes like AT denote joint inputs), the directions are: T$\to$V, A$\to$V, AT$\to$V, V$\to$T, VA$\to$T, V$\to$A, VT$\to$A, and T$\to$VA. Multimodal queries or targets are handled by each model's own pathway: omni models use their native multimodal encoder; others fall back to late fusion of unimodal embeddings (mean or concatenation, as the model declares). nDCG@10 is the main metric.

\textbf{Pair Classification} Given two video inputs, predict whether they satisfy a binary criterion (e.g., same activity, same speaker). Similarity is computed between embeddings and the maximum average-precision over cosine similarity serves as the main metric.

\textbf{Video-Centric Question Answering (VCQA)} Given a video input (optionally with its audio track) and a textual question, select the most relevant answer from a small set of text options. This is implemented as retrieval over a query-specific candidate pool; accuracy is the main metric.


\section{Results}

\begin{table*}[!th]
    \centering
    \resizebox{\textwidth}{!}{\setlength{\tabcolsep}{4pt}{\footnotesize
    \begin{tabular}{lll|c|c|ccc|cccccc}
    \toprule
     & & & \textbf{Rank} ($\downarrow$) & \textbf{ZS \%} & \multicolumn{3}{c|}{\textbf{Average}} & \multicolumn{6}{c}{\textbf{Average per Category}} \\
    \cmidrule(r){4-4} \cmidrule(lr){5-5} \cmidrule(lr){6-8} \cmidrule(l){9-14}
    \textbf{Model} & \textbf{Type} & \textbf{Params} & MVEB & & Mean & \textbf{TV} & \textbf{V} & \textbf{Retr} & \textbf{QA} & \textbf{Cls} & \textbf{Clust} & \textbf{Pair} & \textbf{ZS} \\
    \midrule
    \textcolor{gray}{Number of tasks} & & & \textcolor{gray}{(23)} & & & \textcolor{gray}{(19)} & \textcolor{gray}{(9)} & \textcolor{gray}{(10)} & \textcolor{gray}{(1)} & \textcolor{gray}{(6)} & \textcolor{gray}{(2)} & \textcolor{gray}{(2)} & \textcolor{gray}{(2)} \\
    \midrule
    LCO-Embedding-Omni-7B & MLLM-based embedding & 8.9B & \textbf{1} & 100\% & \textbf{57.6} & \textbf{56.8} & \textbf{61.7} & 58.7 & \textbf{57.0} & 59.2 & \textbf{27.3} & 79.6 & 55.5 \\
    e5-omni-7B & MLLM-based embedding & 8.9B & 2 & 100\% & 55.0 & 54.1 & 55.7 & 59.4 & 43.8 & 54.7 & 21.5 & 77.3 & 50.2 \\
    ebind-full & Multimodal binding & 1.8B & 3$=$ & 83\% & 55.5 & 53.8 & 55.8 & \textbf{62.3} & 31.4 & 51.3 & 20.1 & 75.4 & \textbf{61.1} \\
    ebind-audio-vision & Multimodal binding & 764M & 3$=$ & 83\% & 55.5 & 53.8 & 55.8 & \textbf{62.3} & 31.4 & 51.3 & 20.1 & 75.4 & \textbf{61.1} \\
    LCO-Embedding-Omni-3B & MLLM-based embedding & 4.7B & 5 & 100\% & 54.6 & 54.8 & 61.6 & 54.1 & 56.4 & 56.1 & 27.0 & \textbf{80.7} & 52.7 \\
    OmniEmbed-v0.1 & MLLM-based embedding & 8.9B & 6$=$ & 100\% & 52.9 & 51.3 & 57.9 & 52.9 & 50.2 & 58.5 & 20.6 & 74.7 & 48.5 \\
    pe-av-large & Audio-visual contrastive & 2.2B & 6$=$ & 65\% & 54.3 & 52.4 & 55.2 & 62.0 & 29.2 & 54.8 & 21.7 & 75.2 & 37.9 \\
    pe-av-base & Audio-visual contrastive & 1.0B & 8 & 65\% & 53.1 & 49.7 & 53.8 & 59.6 & 30.6 & 54.8 & 22.6 & 75.7 & 34.6 \\
    BidirLM-Omni-2.5B-Embedding & MLLM-based embedding & 2.4B & 9 & 100\% & 51.2 & 52.0 & 58.0 & 45.5 & 55.8 & \textbf{61.2} & 17.1 & 78.7 & 53.9 \\
    pe-av-small & Audio-visual contrastive & 847M & 10 & 65\% & 52.2 & 50.2 & 53.3 & 57.6 & 27.8 & 54.8 & 22.9 & 75.3 & 35.4 \\
    e5-omni-3B & MLLM-based embedding & 4.7B & 11 & 100\% & 48.5 & 48.4 & 55.7 & 52.2 & 45.4 & 48.3 & 21.4 & 77.0 & 30.8 \\
    omni-embed-nemotron-3b & MLLM-based embedding & 4.7B & 12 & 100\% & 42.8 & 35.8 & 54.7 & 32.9 & 46.6 & 54.6 & 20.5 & 77.7 & 42.5 \\
    jina-embeddings-v5-omni-nano & MLLM-based embedding & 986M & 13 & NA & 20.8 & 7.7 & 24.9 & 10.3 & 24.4 & 28.0 & 12.6 & 59.1 & 20.0 \\
    jina-embeddings-v5-omni-small & MLLM-based embedding & 1.6B & 14 & NA & 19.4 & 8.2 & 24.7 & 6.2 & 21.2 & 28.5 & 13.7 & 58.6 & 23.8 \\
    Qwen2.5-Omni-7B & Generative MLLM & 10.7B & 15 & NA & 12.8 & 10.4 & 30.7 & 0.5 & 13.2 & 20.2 & 7.5 & 53.2 & 16.5 \\
    Qwen2.5-Omni-3B & Generative MLLM & 5.5B & 16 & NA & 11.4 & 7.8 & 30.1 & 0.6 & 14.4 & 19.3 & 5.9 & 53.7 & 3.8 \\
    \bottomrule
    \end{tabular}}}
    \caption{Top 16 Models on \textbf{MVEB} ranked by Borda count over 23 tasks. \textbf{Rank} marked with $=$ shares a Borda count with the row above. \textbf{ZS~\%}: disclosed-zero-shot percentage from \autoref{tab:contamination} (NA for models that do not disclose training data). \textbf{Mean} is the arithmetic mean over the model's 23 evaluated tasks; \textbf{TV} and \textbf{V} are the same model's mean on the MVEB(text, video) (19 tasks) and MVEB(video) (9 tasks) subsets respectively, lifted from \autoref{tab:mveb-text-video-results} and \autoref{tab:mveb-video-results}. Task categories: Retr (retrieval; per-direction breakdown in appendix), QA, Cls (classification), Clust (clustering), Pair (pair classification), ZS (zero-shot classification). Model types are defined in \S\ref{sec:methodology-models}. \textbf{Bold} = best per column \textit{within the 16 rows shown}; the global MVEB(text, video) and MVEB(video) leaders are Qwen3-VL-Embedding (see Appendix~\ref{appdx:results-scope-variants}).}
    \label{tab:mveb-main-results}
\end{table*}

\label{sec:key-findings}

\autoref{tab:mveb-main-results} reports the MVEB leaderboard for the 16 models that can run all 23 tasks; per-task scores, per-direction retrieval breakdowns, the MVEB+ aggregate over all 184 tasks, and the modality-restricted leaderboards are in Appendix~\ref{appdx:results}.

\begin{figure}[h]
    \centering
    \includegraphics[width=1.0\linewidth]{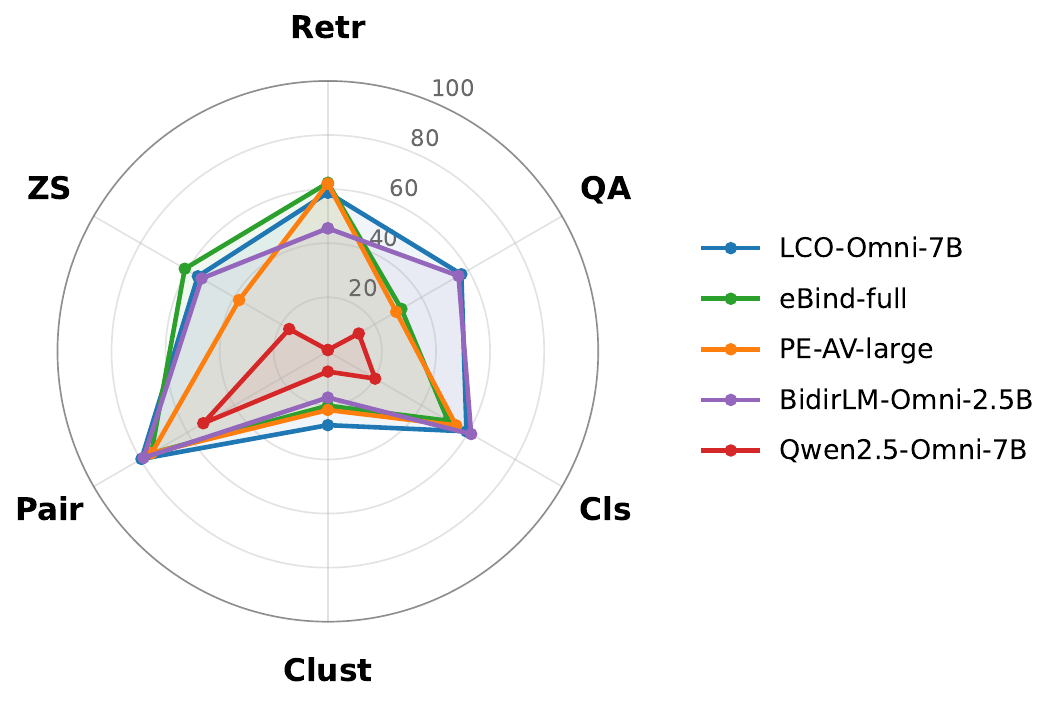}
    \caption{Per-category mean scores for five representative models spanning the four paradigms. LCO-Embedding-Omni-7B, eBind-full, PE-AV-large, and BidirLM-Omni-2.5B trace broadly similar but distinct shapes; Qwen2.5-Omni-7B (generative MLLM used as embedder) collapses inward across every category. No single model dominates all six.}
    \label{fig:model_performance_per_task_radar_chart}
\end{figure}

\paragraph{No single model dominates, but general-purpose contenders are emerging.} LCO-Embedding-Omni-7B ranks first by Borda count with a mean of 57.6 and is the strongest model on QA and clustering, while losing classification to the newer BidirLM-Omni-2.5B-Embedding (61.2), retrieval and zero-shot classification to eBind, and pair classification to LCO-Embedding-Omni-3B. The category leaders span two paradigms (MLLM-based embedding and multimodal binding). We read LCO-Embedding-Omni as the closest current candidate for a general-purpose video embedder: it leads or ties on two categories and stays within a few points of the leader on the rest. The per-category spread is visualized in \autoref{fig:model_performance_per_task_radar_chart}.

\paragraph{Text-video specialists lead MVEB(text, video).} The modality-restricted MVEB(text, video) leaderboard (\autoref{tab:mveb-text-video-results}) evaluates 25 models on the 19 text-video tasks of MVEB. Qwen3-VL-Embedding-8B and -2B~\cite{qwen3vlembedding}, vision-language embedders without an audio path, take ranks 1 and 2 at 60.9 and 58.1 mean, ahead of every omni audio+video+text model in our roster (LCO-Embedding-Omni-7B at 56.8); the Qwen3-VL family wins 5 of 6 task categories, with only QA staying with LCO-Embedding-Omni-3B (32.8). The reordering relative to the headline MVEB leaderboard shows that the text-video scope rewards a different family of models from the full audio+video+text scope, making MVEB(text, video) a useful evaluation surface in its own right rather than a fallback. Full ranking, per-category breakdown, and an extended discussion are in Appendix~\ref{appdx:text-video-leaderboard}.

\paragraph{Adapted MLLMs beat repurposed generative MLLMs by a wide margin.} The MLLM-based embedding family fills five of the top ten ranks on MVEB and dominates the top of the text-video scope (\autoref{tab:mveb-text-video-results}). Generative MLLMs used as embedders via hidden-state pooling collapse: Qwen2.5-Omni-7B and 3B score 12.8 and 11.4 mean on MVEB, an order of magnitude below their MLLM-embed counterparts at the same scale. Within a single backbone family the gap is even sharper: e5-omni-7B (an MLLM-embed adaptation of Qwen2.5-Omni-7B) scores 55.0 vs.\ 12.8 for the unadapted generative baseline. A contrastive embedding stage is therefore a near-prerequisite, not a refinement.

\begin{figure*}[t]
    \centering
    \includegraphics[width=\linewidth]{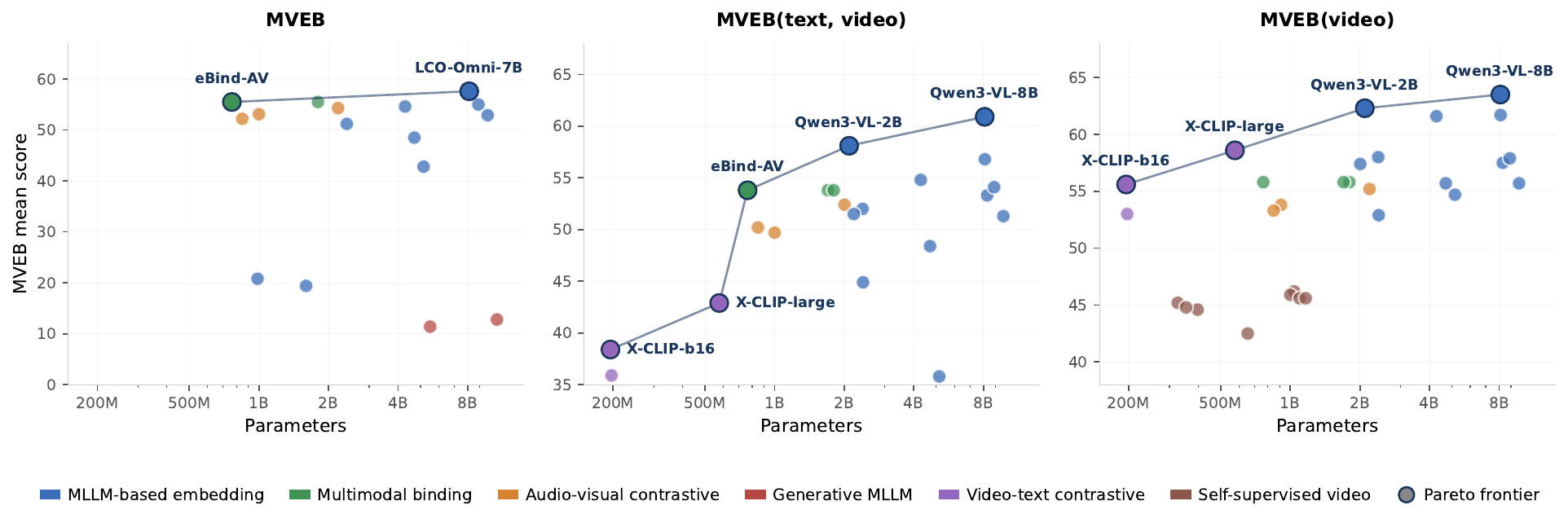}
    \caption{Mean score vs.\ parameter count for each of the three leaderboards: MVEB (16 models), MVEB(text, video) (25 models), and MVEB(video) (33 models). Pareto-frontier models are highlighted and labeled per panel; off-frontier dots are colored by paradigm but unlabeled. On MVEB the frontier runs from eBind-AV (764M, 55.5) to LCO-Omni-7B (8.9B, 57.6) (a 12$\times$ parameter gap for 2.1 points of score). Removing audio (MVEB(text, video) and MVEB(video)) replaces the frontier with Qwen3-VL-Embedding 2B/8B, consistent with the audio-coverage finding in \S\ref{sec:audio-contribution}.}
    \label{fig:param_vs_borda_rank}
\end{figure*}

\paragraph{Smaller specialized contrastive models punch above their weight.} eBind (1.8B / 764M) ties for third place by Borda and wins both retrieval (62.3) and zero-shot classification (61.1) outright, despite being four to ten times smaller than the MLLM-embed leaders (\autoref{fig:param_vs_borda_rank}). Perception Encoder audio-visual variants (847M--2.2B) similarly outscore the comparable-size MLLM-embed model OmniEmbed-Nemotron-3B (4.7B) on retrieval and clustering. We read this as evidence that training-data alignment matters more than backbone scale for cross-modal video tasks.

\paragraph{Clustering and zero-shot classification show the lowest absolute scores.} The strongest model reaches only 27.3 on clustering\footnote{V-measure is bounded by label granularity and dataset noise as well as embedding quality, and its scale is not directly comparable to accuracy-based categories, so we read clustering scores as a relative-comparison signal rather than a direct measure of difficulty.} (LCO-Embedding-Omni-7B) and no model exceeds 61.1 on zero-shot classification, against 74--81 on pair classification for top models. The pattern is consistent across model families.

\section{Analyses}

\subsection{Contribution of the Audio Track to Video Understanding}
\label{sec:audio-contribution}

We compare video-only (\texttt{v}) and video$+$audio (\texttt{va}) task
variants across 48 paired task groups spanning five task types and the
14 audio-capable models in our roster, measuring the audio delta
$\Delta = \text{score}_{\text{va}} - \text{score}_{\text{v}}$.\footnote{The
jina-embeddings-v5-omni-nano/small models are excluded from this analysis:
their video-only scores on the paired tasks sit at near-random levels
($<$0.01 on Kinetics splits), so the va$-$v delta is dominated by
base-line noise rather than by audio's contribution and would distort the
per-paradigm summary.}

\begin{table}[t]
\centering
\setlength{\tabcolsep}{4pt}
\renewcommand{\arraystretch}{1.05}
\begin{tabular}{lrrr}
\toprule
\textbf{Dataset type} & $\bar{\Delta}$ & $\sigma$ & $N$ \\
\midrule
AV-grounded & $+0.016$ & $0.064$ & $475$ \\
V-grounded  & $-0.046$ & $0.059$ & $195$ \\
\bottomrule
\end{tabular}
\caption{Mean audio delta ($\Delta = \text{score}_{\text{va}} -
         \text{score}_{\text{v}}$) across 48 paired task groups and
         14 models, split by annotation provenance
         (Table~\ref{tab:av-provenance}). $N$ counts model--dataset pairs.}
\label{tab:audio-intentionality}
\end{table}

\paragraph{Annotation provenance predicts audio's contribution.}
Table~\ref{tab:audio-intentionality} reports mean $\Delta$ split by the
annotation provenance defined in \autoref{tab:av-provenance}:
\textit{AV-grounded} datasets had labels produced from both audio and
visual content, \textit{V-grounded} datasets had labels produced from
visuals alone (even though the clips often carry audio). Audio helps
AV-grounded datasets by $+0.016$ on average and hurts V-grounded ones
by $-0.046$: a six-point gap consistent across task types and model
families. The V-grounded penalty is about three times the AV-grounded
benefit, so the result reads more as ``audio hurts when labels don't
reward it'' than as ``audio is a strong positive signal when they do.''

\paragraph{The effect varies more within paradigms than across them.}
Per-paradigm averages sit near zero ($\bar\Delta \in [-0.023, +0.010]$
across the four paradigms), with the within-paradigm range often
exceeding the between-paradigm spread; only multimodal binding
(eBind-full and eBind-audio-vision) consistently loses from audio
($-0.023$ each). MLLM-based embedders span almost the full range of
observed deltas ($-0.039$ at e5-omni-3B to $+0.023$ at
BidirLM-Omni-2.5B), so audio handling is implementation-specific
rather than paradigm-determined (per-model deltas in
Table~\ref{tab:audio_per_model}, Appendix~\ref{appdx:audio-contribution}).

Per-dataset and per-model breakdowns, including cross-modal retrieval alignment for datasets with paired v2a/a2v variants, are in Appendix~\ref{appdx:audio-contribution}.

\subsection{Test-Time Scaling of Temporal Context}
\label{analyses_test_time_scaling_frames}

We sweep the number of uniformly sampled video frames $N \in \{1, 8, 16, 32, 64\}$ at test time, holding audio sampling and all other inputs at each model's default. The analysis uses a subset of MVEB tasks with average clip duration above 15~s (selection criteria in Appendix~\ref{appdx:frame-scaling}).

\begin{figure}[h]
    \centering
    \includegraphics[width=1.0\linewidth]{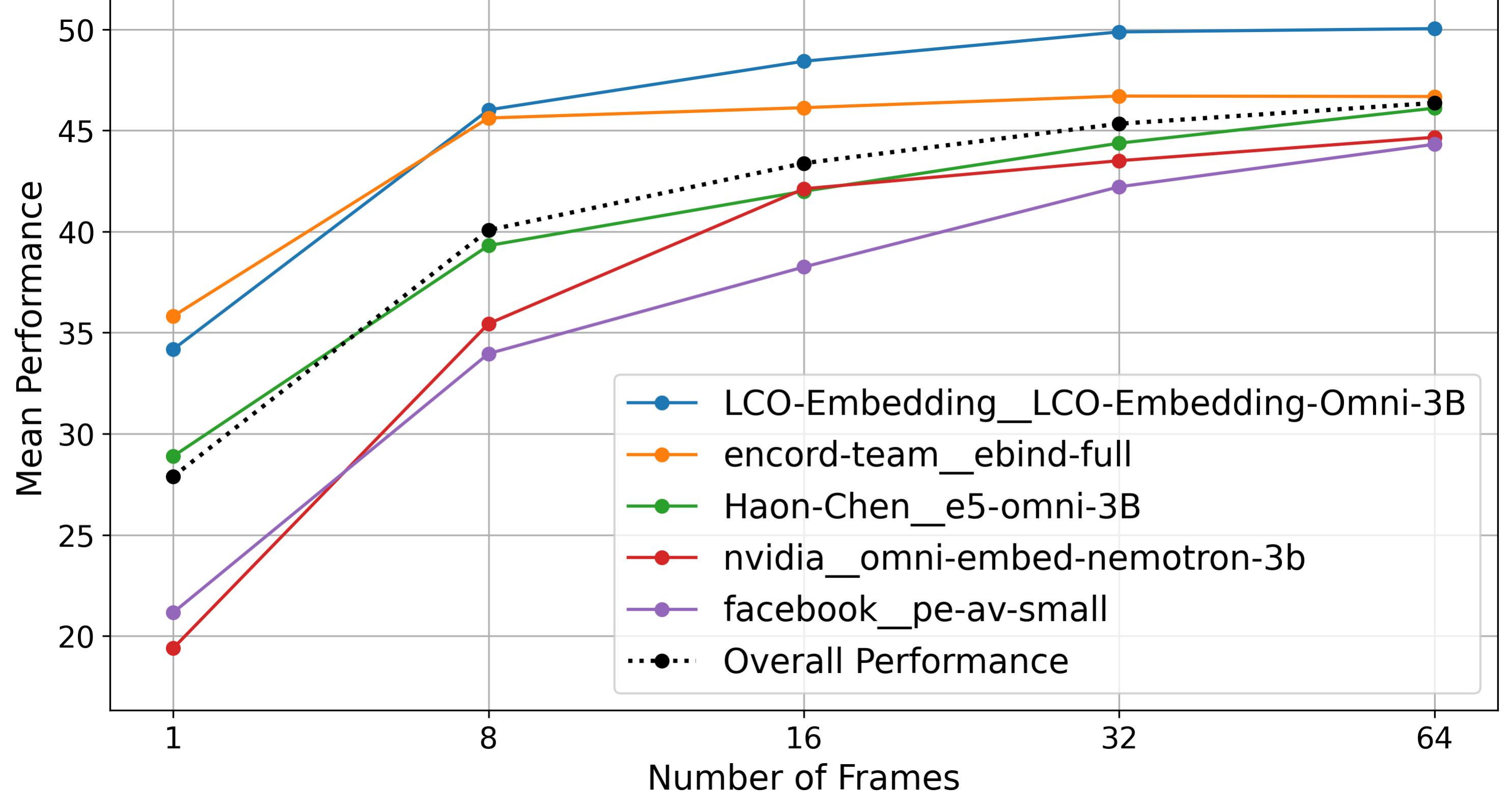}
    \caption{Per-model performance as a function of sampled frame count. The black dotted line is the mean over all models. Most of the gain comes from the first jump to multi-frame sampling; returns plateau beyond 32 frames.}
    \label{fig:model_performance_scaling_with_overall}
\end{figure}

Mean performance scales logarithmically with $N$ (\autoref{fig:model_performance_scaling_with_overall}). Moving from a single-frame spatial baseline to $N=8$ yields a 43.7\% relative improvement; doubling from 32 to 64 frames adds only 2.2\% absolute. For general-purpose evaluation, 32 frames is a reasonable ceiling.

The aggregate trend hides large per-task variance: some tasks (e.g.,~Breakfast classification, VATEX retrieval) scale steeply with $N$; others (e.g.,~OmniVideoBench QA, WorldSense-1min QA) barely move. Per-task and per-model curves are in Appendix~\ref{appdx:frame-scaling}.

\subsection{Cross-Modal Retrieval Direction Structure}
\label{sec:retrieval-direction-structure}

\begin{figure}[h]
    \centering
    \includegraphics[width=1.0\linewidth]{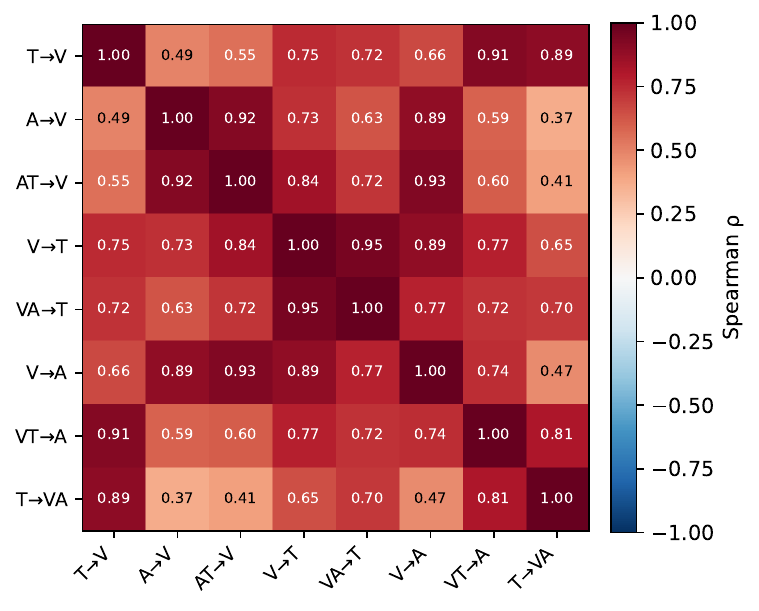}
    \caption{Pairwise Spearman correlation across the eight retrieval directions over 16 models. The most distinct pair is $T \to VA$ vs.\ $A \to V$ at $\rho = 0.38$.}
    \label{fig:retrieval-direction-correlation}
\end{figure}

Pairwise Spearman correlation across the eight retrieval directions (\autoref{fig:retrieval-direction-correlation}) shows the directions cluster into three capability groups rather than measuring eight independent things: text-target ($V \to T$, $VA \to T$ at $\rho = 0.96$), video-target with audio in the query ($A \to V$, $AT \to V$, $V \to A$; $\rho \geq 0.87$ among them), and text-as-query ($T \to V$, $T \to VA$, $VT \to A$; $\rho \geq 0.77$). The most decoupled pair is $T \to VA$ vs.\ $A \to V$ at $\rho = 0.38$, isolating audio-as-target as a distinct axis only because MVEB pairs audio$+$video as joint retrieval modalities. Full pairwise correlations and discussion are in Appendix~\ref{appdx:retrieval-correlation}.

\section{Conclusion}
MVEB provides a unified 23-task benchmark for video embedding models, drawn from a 184-task pool (MVEB+) and accompanied by modality-restricted leaderboards for text-video and video-only encoders. Across 33 models from the major video-embedding paradigms no single approach dominates. Our paired video-only vs.\ audio+video analysis shows that audio's contribution depends on dataset annotation provenance (AV-grounded vs.\ V-grounded), and the MVEB(text, video) leaderboard reorders the top of the field relative to MVEB. We release MVEB inside the MTEB ecosystem with task and model versioning, and a community contribution pipeline, so the benchmark can co-evolve with the field rather than freeze as a snapshot.

\section*{Limitations}

\paragraph{Model Coverage.} Our roster covers 33 publicly-available checkpoints across the major paradigms; the field moves quickly so the list is a snapshot, but new models are added on a rolling basis through the MTEB registry. Notable models we plan to add include ImageBind~\cite{girdhar2023imagebind}, LanguageBind~\cite{zhu2024languagebind}, InternVideo~\cite{wang2022internvideo}, VideoMAE~\cite{tong2022videomae}, and VJEPA-2.1~\cite{murlabadia2026vjepa21unlockingdense}; these have open weights but no clean inference API or versioned checkpoints yet.

\paragraph{Frame and Audio Budgets.} Each model uses its declared sampling configuration (\S\ref{sec:eval-protocol}), so absolute scores can partly reflect the frame and audio budgets each model was trained under; the frame-scaling analysis (\autoref{analyses_test_time_scaling_frames}) explores how scores move when this budget is varied at test time.

\paragraph{Dataset Coverage.} While MVEB+ covers 184 tasks across multiple domains, some areas remain thin: low-resource languages are underrepresented, long-form video content (lectures, documentaries) is largely absent due to runtime cost, and several fine-grained domains (sports beyond a few action-recognition datasets, scientific or medical video, sign language) have limited or no coverage. New datasets are added on a rolling basis through community contributions.

\paragraph{Training Data Disclosure and Leakage.} Many MVEB tasks draw from web-curated sources (Kinetics, UCF101, MSR-VTT, VATEX, Panda-70M) that are common in large-scale video-caption pretraining, so train-test contamination is a real risk. \autoref{appdx:contamination} cross-references each model's declared training data against MVEB task names and reports the resulting disclosed-zero-shot percentage per model. The audit is necessarily incomplete for the subset of our roster that does not disclose training data at dataset granularity; rather than assigning these models an optimistic 100\% zero-shot, we report their contamination columns as \texttt{NA} in \autoref{tab:contamination} and surface the same convention in the headline leaderboard. Their MVEB scores therefore cannot be certified as zero-shot, and per-dataset comparisons involving them should be interpreted with that disclosure gap in mind.

\paragraph{Annotation Quality.} A subset of MVEB source datasets inherits label conventions from their original releases that surface as task-level ambiguities, particularly in emotion-recognition tasks. Recurring issues include single-label assignment on utterances that carry multiple sentiments, labels that capture the surface emotion even when the utterance is sarcastic or non-literal, and a fraction of clips that appear plainly mislabeled. We do not silently include these splits: \autoref{appdx:annotation-quality} flags every affected dataset with concrete failure-mode examples so readers can gate per-dataset interpretation rather than treating scores uniformly across the benchmark. The right long-term fix is to replace affected splits with cleaner sources as the field releases them, and we plan to deprecate flagged splits in subsequent MVEB releases as replacements appear; the MVEB+ contribution pipeline accepts replacement datasets, re-annotation passes against existing splits, and human-model agreement audits in the style of HUME~\cite{assadi2025humemeasuringhumanmodelperformance}, originally developed for MTEB text tasks. We retain the as-released labels in the current release so MVEB scores remain comparable with prior published results.


\section*{Ethical Considerations}

MVEB aggregates publicly-released video datasets under each source's release license. For datasets where we host evaluation artifacts on HuggingFace (under the \texttt{mteb/} namespace), we redistribute decoded frame samples and 16~kHz mono audio extracted from the source clips, paired with the task wrapper (loader, evaluation code, and metric implementation); we do not re-encode or re-host the original full-resolution video files. For datasets where the upstream owner already provides a public mirror, MVEB ships only the task wrapper and links back to that release. The frame/audio packaging is a working-around-bandwidth choice rather than a content choice: the underlying clips, labels, and consent regime are the upstream releases unchanged. Several included datasets (e.g.,~Kinetics, UCF101, AVMeme, AVE-Dataset, MELD, RAVDESS, VGGSound) contain identifiable individuals; we evaluate models only as embedders on the labels and tasks those datasets already release, and do not introduce new identifying labels or biometric tasks.

The 33 models we evaluate include multimodal large language models capable of free-form generation; we use them only to produce frozen embeddings, never to generate text, audio, or video. We therefore do not amplify generative risks beyond what each model's release already permits.

The benchmark inherits the biases of the underlying datasets, including domain skew (most clips are English-language and from web-curated or US-centric sources) and demographic skew in datasets that depict people. We flag these explicitly in the Dataset Coverage limitation and encourage community contributions that broaden coverage to under-represented languages, regions, and domains.

\section*{Acknowledgement}
We are extremely thankful to Laude Institute for supporting and enabling this work.
Márton Kardos and Kenneth Enevoldsen are funded by the Danish Foundation Models project (4378-00001B) and the European Union, Horizon Europe (101178170). Kenneth Enevoldsen is additionally funded by the Danish National Research Foundation (DNRF193), the Aage and Johanne Louis-Hansens Foundation (25-1-17733), and the Augustinus Foundation (2025-0299).
We thank Nguyen Tai and Guangyu Song for contributing dataset metadata for several MVEB+ tasks.

\bibliography{refs}


\appendix

\section{Tasks overview}
\label{sec:overview}


\begin{table*}[ht]
\centering
\resizebox{\linewidth}{!}{
\begin{tabular}{lccccrrrcc}
\toprule
\textbf{Dataset} & \textbf{Citation} & \textbf{MVEB} & \textbf{TV} & \textbf{V} & \textbf{N.samples} & \textbf{Total Duration(s)} & \textbf{N.Langs} & \textbf{Domains} & \textbf{Main metric} \\
\midrule
\multicolumn{10}{l}{\textbf{Video Classification}} \\
AVEDatasetVideoClassification & \cite{tian2018audio} &  &  &  & 402 & 1h (4,026s) & 1 & Web, AudioScene & accuracy \\
AVMemeVideoClassification & \cite{jiang2026avmeme} &  & \checkmark & \checkmark & 900 & 3h (11,010s) & 16 & Web, Entertainment, Music & accuracy \\
BreakfastClassification & \cite{kuehne2014language} & \checkmark & \checkmark & \checkmark & 433 & 17h (62,710s) & 1 & Scene & accuracy \\
Diving48Classification.V2 & \cite{Li_2018_ECCV} &  &  &  & 1,970 & 2h (8,621s) & 1 & Sport & accuracy \\
HMDB51Classification & \cite{6126543} &  &  & \checkmark & 1,530 & 1h (4,833s) & 1 & Scene & accuracy \\
HumanAnimalCartoonV & \cite{dong2023simmmdg} &  &  &  & 644 & 1h (5,118s) & 1 & Web, Scene & accuracy \\
Kinetics400V & \cite{kay2017kineticshumanactionvideo} &  &  &  & 3,995 & 10h (38,195s) & 1 & Web, Scene & accuracy \\
Kinetics600V & \cite{carreira2018short} &  &  & \checkmark & 9,576 & 1d (92,401s) & 1 & Web, Scene & accuracy \\
Kinetics700V & \cite{smaira2020short} &  & \checkmark &  & 11,190 & 1d (108,857s) & 1 & Web, Scene & accuracy \\
MELDVideoClassification & \cite{poria2019meld} &  &  & \checkmark & 2,048 & 1h (7,134s) & 1 & Entertainment & accuracy \\
MusicAVQACLSVideoClassification & \cite{li2022learning} &  &  &  & 1,706 & 1d (101,285s) & 1 & Music & accuracy \\
RAVDESSVClassification & \cite{10.1371/journal.pone.0196391} &  &  &  & 1,440 & 1h (5,312s) & 1 & Spoken & accuracy \\
SomethingSomethingV2Classification & \cite{goyal2017something} &  &  &  & 5,444 & 6h (22,173s) & 1 & Scene & accuracy \\
UCF101VideoClassification & \cite{Soomro2012UCF101} &  &  &  & 1,944 & 3h (13,304s) & 1 & Web, Scene & accuracy \\
VGGSoundV & \cite{chen2020vggsound} &  & \checkmark &  & 9,888 & 1d (98,459s) & 1 & Web & accuracy \\
WorldSenseVideoClassification & \cite{hong2025worldsense} &  &  & \checkmark & 1,047 & 13h (50,231s) & 1 & Scene, AudioScene, Music, Entertainment & accuracy \\
\midrule
\multicolumn{10}{l}{\textbf{Video Clustering}} \\
AVEDatasetVideoClustering & \cite{Tian_2018_ECCV} &  &  &  & 402 & 1h (4,026s) & 1 & Spoken, Scene, Music & v\_measure \\
HMDB51Clustering & \cite{kuehne2011hmdb} &  &  &  & 1,530 & 1h (4,833s) & 1 & Scene & v\_measure \\
MELDEmotionVideoClustering & \cite{poria2019meld} &  &  &  & 1,354 & 1h (4,980s) & 1 & Entertainment & v\_measure \\
MELDSpeakerVideoClustering & \cite{poria2019meld} &  &  &  & 2,610 & 2h (8,844s) & 1 & Entertainment & v\_measure \\
MusicAVQACLSVideoClustering & \cite{li2022learning} &  &  &  & 1,706 & 1d (101,285s) & 1 & Music & v\_measure \\
RAVDESSVideoClustering & \cite{10.1371/journal.pone.0196391} &  & \checkmark &  & 1,440 & 1h (5,312s) & 1 & Spoken & v\_measure \\
UCF101VideoClustering & \cite{Soomro2012UCF101} &  &  &  & 1,944 & 3h (13,304s) & 1 & Web, Scene & v\_measure \\
WorldSense1MinDomainVideoClustering & \cite{hong2025worldsense} &  &  &  & 568 & 7h (27,808s) & 1 & Scene, Web, Entertainment & v\_measure \\
\midrule
\multicolumn{10}{l}{\textbf{Video Pair Classification}} \\
AVEDatasetVPairClassification & \cite{tian2018audio} &  &  &  & 804 & 4h (16,104s) & 1 & Web, AudioScene & max\_ap \\
AVSpeakerBenchPairClassification & \cite{nguyen2024seehearunderstand} &  &  &  & 2,048 & 15h (56,521s) & 1 & Spoken & max\_ap \\
HumanAnimalCartoonVPairClassification & \cite{dong2023simmmdg} &  & \checkmark & \checkmark & 1,288 & 5h (20,445s) & 1 & Web, Scene & max\_ap \\
MELDVPairClassification & \cite{poria2019meld} &  &  &  & 2,048 & 3h (14,093s) & 1 & Entertainment & max\_ap \\
MusicAVQAVPairClassification & \cite{li2022learning} &  &  & \checkmark & 2,048 & 2d (243,133s) & 1 & Music & max\_ap \\
RAVDESSAVVPairClassification & \cite{10.1371/journal.pone.0196391} &  &  & \checkmark & 2,048 & 4h (15,096s) & 1 & Spoken & max\_ap \\
VinogroundPairClassification & \cite{zhang2024vinoground} &  &  &  & 2,000 & 9h (33,330s) & 1 & Scene & max\_ap \\
\bottomrule
\end{tabular}
}
\caption{Video-only tasks in MVEB. All tasks use video modality exclusively.}
\label{tab:mveb-video-only-tasks}
\end{table*}

\begin{table*}[ht]
\centering
\resizebox{\linewidth}{!}{
\begin{tabular}{lcccrrrcc}
\toprule
\textbf{Dataset} & \textbf{Citation} & \textbf{MVEB} & \textbf{TV} & \textbf{N.samples} & \textbf{Total Duration(s)} & \textbf{N.Langs} & \textbf{Domains} & \textbf{Main metric} \\
\midrule
\multicolumn{9}{l}{\textbf{Video Zero-shot Classification}} \\
AVEDatasetVideoZeroShot & \cite{tian2018audio} &  &  & 402 & 1h (4,026s) & 1 & Web, AudioScene & accuracy \\
AVMemeVideoZeroShot & \cite{jiang2026avmeme} &  &  & 900 & 3h (11,010s) & 16 & Web, Entertainment, Music & accuracy \\
BreakfastZeroShot & \cite{kuehne2014language} &  &  & 433 & 17h (62,710s) & 1 & Scene & accuracy \\
HMDB51ZeroShot & \cite{6126543} & \checkmark &  & 1,530 & 1h (4,833s) & 1 & Scene & accuracy \\
HumanAnimalCartoonZeroShot & \cite{dong2023simmmdg} &  &  & 644 & 1h (5,118s) & 1 & Web, Scene & accuracy \\
Kinetics400ZeroShot & \cite{kay2017kineticshumanactionvideo} &  & \checkmark & 3,995 & 10h (38,195s) & 1 & Web, Scene & accuracy \\
Kinetics600VZeroShot & \cite{carreira2018short} &  &  & 9,576 & 1d (92,401s) & 1 & Web, Scene & accuracy \\
Kinetics700VZeroShot & \cite{smaira2020short} &  &  & 11,190 & 1d (108,857s) & 1 & Web, Scene & accuracy \\
MELDVideoZeroShot & \cite{poria2019meld} &  & \checkmark & 2,048 & 1h (7,134s) & 1 & Entertainment & accuracy \\
MusicAVQACLSVideoZeroShot & \cite{li2022learning} &  &  & 1,706 & 1d (101,285s) & 1 & Music & accuracy \\
RAVDESSVZeroShot & \cite{10.1371/journal.pone.0196391} &  &  & 1,440 & 1h (5,312s) & 1 & Spoken & accuracy \\
UCF101VideoZeroShotClassification & \cite{Soomro2012UCF101} &  & \checkmark & 1,944 & 3h (13,304s) & 1 & Web, Scene & accuracy \\
VGGSoundVideoZeroshot & \cite{chen2020vggsound} &  &  & 9,888 & 1d (98,459s) & 1 & Web & accuracy \\
WorldSenseVideoZeroShot & \cite{hong2025worldsense} &  & \checkmark & 1,047 & 13h (50,231s) & 1 & Scene, AudioScene, Music, Entertainment & accuracy \\
\midrule
\multicolumn{9}{l}{\textbf{Video Pair Classification}} \\
VideoConPairClassification & \cite{bansal2023videocon} &  &  & 1,138 & 11h (42,348s) & 1 & Scene & max\_ap \\
\midrule
\multicolumn{9}{l}{\textbf{Video-Centric QA}} \\
AVMemeExamVideoCentricQA & \cite{jiang2026avmeme} &  &  & 4,612 & 3h (11,010s) & 1 & Web & accuracy \\
AVQAVideoCentricQA & \cite{yang2022avqa} &  &  & 4,605 & 2h (9,184s) & 1 & Web & accuracy \\
AVSpeakerBenchVideoCentricQA & \cite{nguyen2024seehearunderstand} &  &  & 16,060 & 12h (46,068s) & 1 & Web & accuracy \\
DailyOmniVideoCentricQA & \cite{zhou2025dailyomni} &  &  & 5,980 & 14h (51,726s) & 1 & Web & accuracy \\
EgoSchemaVideoCentricQA & \cite{mangalam2023egoschema} & \checkmark &  & 3,000 & 1d (90,000s) & 1 & Web & accuracy \\
NExTQAVideoCentricQA & \cite{xiao2021next} &  &  & 5,958 & 10h (39,190s) & 1 & Web & accuracy \\
OmniVideoBenchVideoCentricQA & \cite{li2025omnivideobench} &  & \checkmark & 2,545 & 18h (66,152s) & 1 & Web & accuracy \\
PerceptionTestVideoCentricQA & \cite{patraucean2024perception} &  &  & 3,752 & 5h (19,843s) & 1 & Web & accuracy \\
VideoMMEShortVideoCentricQA & \cite{fu2024video} &  &  & 4,500 & 20h (72,599s) & 1 & Web & accuracy \\
WorldQAVideoCentricQA & \cite{zhang2024worldqa} &  &  & 16,372 & 3d (313,887s) & 1 & Web & accuracy \\
WorldSense1MinVideoCentricQA & \cite{hong2025worldsense} &  &  & 5,165 & 13h (50,231s) & 1 & Web & accuracy \\
\midrule
\multicolumn{9}{l}{\textbf{Video Retrieval}} \\
AVMemeExamT2VRetrieval & \cite{jiang2026avmeme} &  & \checkmark & 1,800 & 3h (11,010s) & 1 & Web, Social & ndcg\_at\_10 \\
AVMemeExamV2TRetrieval & \cite{jiang2026avmeme} &  &  & 1,800 & 3h (11,010s) & 1 & Web, Social & ndcg\_at\_10 \\
ActivityNetCaptionsT2VRetrieval & \cite{krishna2017dense} & \checkmark & \checkmark & 9,768 & 6d (579,288s) & 1 & Web, Spoken & ndcg\_at\_10 \\
ActivityNetCaptionsV2TRetrieval & \cite{krishna2017dense} &  &  & 9,768 & 6d (579,288s) & 1 & Web, Spoken & ndcg\_at\_10 \\
AudioCapsAVT2VRetrieval & \cite{kim2019audiocaps} &  & \checkmark & 1,330 & 1h (6,541s) & 1 & Encyclopaedic, Web & ndcg\_at\_10 \\
AudioCapsAVV2TRetrieval & \cite{kim2019audiocaps} &  &  & 1,330 & 1h (6,541s) & 1 & Encyclopaedic, Web & ndcg\_at\_10 \\
DiDeMoT2VRetrieval & \cite{hendricks2017localizing} &  &  & 1,998 & 13h (50,032s) & 1 & Web, Spoken & ndcg\_at\_10 \\
DiDeMoV2TRetrieval & \cite{hendricks2017localizing} &  & \checkmark & 1,998 & 13h (50,032s) & 1 & Web, Spoken & ndcg\_at\_10 \\
MSRVTTT2V & \cite{xu2016msrvtt} &  &  & 1,758 & 3h (13,369s) & 1 & Unknown & ndcg\_at\_10 \\
MSRVTTV2T & \cite{xu2016msrvtt} &  &  & 1,758 & 3h (13,369s) & 1 & Unknown & ndcg\_at\_10 \\
MSVDT2VRetrieval & \cite{microsoft2011msvd} & \checkmark &  & 1,320 & 1h (6,642s) & 1 & Web, Spoken & ndcg\_at\_10 \\
MSVDV2TRetrieval & \cite{microsoft2011msvd} &  & \checkmark & 1,320 & 1h (6,642s) & 1 & Web, Spoken & ndcg\_at\_10 \\
Panda70MT2VRetrieval & \cite{chen2024panda} &  & \checkmark & 6,790 & 10h (37,488s) & 1 & Web, Spoken & ndcg\_at\_10 \\
Panda70MV2TRetrieval & \cite{chen2024panda} &  &  & 6,790 & 10h (37,488s) & 1 & Web, Spoken & ndcg\_at\_10 \\
Shot2Story20KT2VRetrieval & \cite{han2023shot2story} &  &  & 8,046 & 18h (67,101s) & 1 & Web, Spoken & ndcg\_at\_10 \\
Shot2Story20KV2TRetrieval & \cite{han2023shot2story} &  &  & 8,046 & 18h (67,101s) & 1 & Web, Spoken & ndcg\_at\_10 \\
TUNABenchT2VRetrieval & \cite{ye2025tuna} &  &  & 2,000 & 3h (14,248s) & 1 & Web, Spoken & ndcg\_at\_10 \\
TUNABenchV2TRetrieval & \cite{ye2025tuna} &  &  & 2,000 & 3h (14,248s) & 1 & Web, Spoken & ndcg\_at\_10 \\
VALOR32KT2VRetrieval & \cite{chen2023valor} &  & \checkmark & 6,982 & 9h (34,636s) & 1 & Web, Spoken & ndcg\_at\_10 \\
VALOR32KV2TRetrieval & \cite{chen2023valor} &  &  & 6,982 & 9h (34,636s) & 1 & Web, Spoken & ndcg\_at\_10 \\
VATEXT2VRetrieval & \cite{wang2019vatex} &  & \checkmark & 2,000 & 1d (157,615s) & 1 & Web, Spoken & ndcg\_at\_10 \\
VATEXV2TRetrieval & \cite{wang2019vatex} &  &  & 2,000 & 1d (157,615s) & 1 & Web, Spoken & ndcg\_at\_10 \\
VGGSoundAVT2VRetrieval & \cite{chen2020vggsound} &  &  & 1,392 & 1h (6,952s) & 1 & Web, Spoken & ndcg\_at\_10 \\
VGGSoundAVV2TRetrieval & \cite{chen2020vggsound} &  &  & 1,392 & 1h (6,952s) & 1 & Web, Spoken & ndcg\_at\_10 \\
YouCook2T2VRetrieval & \cite{zhou2018towards} &  &  & 6,208 & 17h (61,878s) & 1 & Web, Spoken & ndcg\_at\_10 \\
YouCook2V2TRetrieval & \cite{zhou2018towards} &  &  & 6,208 & 17h (61,878s) & 1 & Web, Spoken & ndcg\_at\_10 \\
\bottomrule
\end{tabular}
}
\caption{Video-text multimodal tasks in MVEB. Tasks use both video and text modalities.}
\label{tab:mveb-video-text-tasks}
\end{table*}

\begin{table*}[ht]
\centering
\resizebox{\linewidth}{!}{
\begin{tabular}{lccrrrcc}
\toprule
\textbf{Dataset} & \textbf{Citation} & \textbf{MVEB} & \textbf{N.samples} & \textbf{Total Duration(s)} & \textbf{N.Langs} & \textbf{Domains} & \textbf{Main metric} \\
\midrule
\multicolumn{8}{l}{\textbf{Video Classification}} \\
AVEDatasetClassification & \cite{tian2018audio} & \checkmark & 402 & 1h (4,026s) & 1 & Web, AudioScene & accuracy \\
AVMemeAudioVideoClassification & \cite{jiang2026avmeme} & \checkmark & 900 & 3h (11,010s) & 16 & Web, Entertainment, Music & accuracy \\
HumanAnimalCartoonVA & \cite{dong2023simmmdg} &  & 644 & 1h (5,118s) & 1 & Web, Scene & accuracy \\
Kinetics400VA & \cite{kay2017kineticshumanactionvideo} &  & 3,995 & 10h (38,195s) & 1 & Web, Scene & accuracy \\
Kinetics600VA & \cite{carreira2018short} &  & 9,576 & 1d (92,401s) & 1 & Web, Scene & accuracy \\
Kinetics700VA & \cite{smaira2020short} & \checkmark & 11,190 & 1d (108,857s) & 1 & Web, Scene & accuracy \\
MELDAudioVideoClassification & \cite{poria2019meld} &  & 2,048 & 1h (7,134s) & 1 & Entertainment & accuracy \\
MusicAVQACLSAudioVideoClassification & \cite{li2022learning} &  & 1,706 & 1d (101,285s) & 1 & Music & accuracy \\
RAVDESSAVClassification & \cite{10.1371/journal.pone.0196391} & \checkmark & 1,440 & 1h (5,312s) & 1 & Spoken & accuracy \\
UCF101VideoAudioClassification & \cite{Soomro2012UCF101} & \checkmark & 1,944 & 3h (13,304s) & 1 & Web, Scene & accuracy \\
VGGSoundVA & \cite{chen2020vggsound} &  & 9,888 & 1d (98,459s) & 1 & Web & accuracy \\
WorldSenseAudioVideoClassification & \cite{hong2025worldsense} &  & 1,047 & 13h (50,231s) & 1 & Scene, AudioScene, Music, Entertainment & accuracy \\
\midrule
\multicolumn{8}{l}{\textbf{Video Clustering}} \\
AVEDatasetAudioVideoClustering & \cite{Tian_2018_ECCV} &  & 402 & 1h (4,026s) & 1 & Spoken, Scene, Music & v\_measure \\
MELDEmotionAudioVideoClustering & \cite{poria2019meld} & \checkmark & 1,354 & 1h (4,980s) & 1 & Entertainment & v\_measure \\
MELDSpeakerAudioVideoClustering & \cite{poria2019meld} &  & 2,610 & 2h (8,844s) & 1 & Entertainment & v\_measure \\
MusicAVQACLSAudioVideoClustering & \cite{li2022learning} & \checkmark & 1,706 & 1d (101,285s) & 1 & Music & v\_measure \\
RAVDESSAVClustering & \cite{10.1371/journal.pone.0196391} &  & 1,440 & 1h (5,312s) & 1 & Spoken & v\_measure \\
UCF101AudioVideoClustering & \cite{Soomro2012UCF101} &  & 1,944 & 3h (13,304s) & 1 & Web, Scene & v\_measure \\
WorldSense1MinDomainAudioVideoClustering & \cite{hong2025worldsense} &  & 568 & 7h (27,808s) & 1 & Scene, Web, Entertainment & v\_measure \\
\midrule
\multicolumn{8}{l}{\textbf{Video Zero-shot Classification}} \\
AVEDatasetZeroShot & \cite{tian2018audio} &  & 402 & 1h (4,026s) & 1 & Web, AudioScene & accuracy \\
AVMemeAudioVideoZeroShot & \cite{jiang2026avmeme} &  & 900 & 3h (11,010s) & 16 & Web, Entertainment, Music & accuracy \\
HumanAnimalCartoonVAZeroShot & \cite{dong2023simmmdg} &  & 644 & 1h (5,118s) & 1 & Web, Scene & accuracy \\
Kinetics400VAZeroShot & \cite{kay2017kineticshumanactionvideo} &  & 3,995 & 10h (38,195s) & 1 & Web, Scene & accuracy \\
Kinetics600VAZeroShot & \cite{carreira2018short} &  & 9,576 & 1d (92,401s) & 1 & Web, Scene & accuracy \\
Kinetics700VAZeroShot & \cite{smaira2020short} &  & 11,190 & 1d (108,857s) & 1 & Web, Scene & accuracy \\
MELDAudioVideoZeroShot & \cite{poria2019meld} &  & 2,048 & 1h (7,134s) & 1 & Entertainment & accuracy \\
MusicAVQACLSAudioVideoZeroShot & \cite{li2022learning} &  & 1,706 & 1d (101,285s) & 1 & Music & accuracy \\
RAVDESSAVZeroShot & \cite{10.1371/journal.pone.0196391} &  & 1,440 & 1h (5,312s) & 1 & Spoken & accuracy \\
UCF101VideoAudioZeroShotClassification & \cite{Soomro2012UCF101} &  & 1,944 & 3h (13,304s) & 1 & Web, Scene & accuracy \\
VGGSoundVideoAudioZeroshot & \cite{chen2020vggsound} &  & 9,888 & 1d (98,459s) & 1 & Web & accuracy \\
WorldSenseAudioVideoZeroShot & \cite{hong2025worldsense} & \checkmark & 1,047 & 13h (50,231s) & 1 & Scene, AudioScene, Music, Entertainment & accuracy \\
\midrule
\multicolumn{8}{l}{\textbf{Video Pair Classification}} \\
AVEDatasetVAPairClassification & \cite{tian2018audio} &  & 804 & 4h (16,104s) & 1 & Web, AudioScene & max\_ap \\
HumanAnimalCartoonVAPairClassification & \cite{dong2023simmmdg} & \checkmark & 1,288 & 5h (20,445s) & 1 & Web, Scene & max\_ap \\
MELDVAPairClassification & \cite{poria2019meld} &  & 2,048 & 3h (14,093s) & 1 & Entertainment & max\_ap \\
MusicAVQAVAPairClassification & \cite{li2022learning} & \checkmark & 2,048 & 2d (243,133s) & 1 & Music & max\_ap \\
RAVDESSAVVAPairClassification & \cite{10.1371/journal.pone.0196391} &  & 2,048 & 4h (15,096s) & 1 & Spoken & max\_ap \\
\midrule
\multicolumn{8}{l}{\textbf{Video-Centric QA}} \\
AVMemeExamVideoAudioCentricQA & \cite{jiang2026avmeme} &  & 4,612 & 3h (11,010s) & 1 & Web & accuracy \\
AVQAVideoAudioCentricQA & \cite{yang2022avqa} &  & 4,605 & 2h (9,184s) & 1 & Web & accuracy \\
AVSpeakerBenchVideoAudioCentricQA & \cite{nguyen2024seehearunderstand} &  & 16,060 & 12h (46,068s) & 1 & Web & accuracy \\
DailyOmniVideoAudioCentricQA & \cite{zhou2025dailyomni} &  & 5,980 & 14h (51,726s) & 1 & Web & accuracy \\
OmniVideoBenchVideoAudioCentricQA & \cite{li2025omnivideobench} &  & 2,545 & 18h (66,152s) & 1 & Web & accuracy \\
PerceptionTestVideoAudioCentricQA & \cite{patraucean2024perception} &  & 3,752 & 5h (19,843s) & 1 & Web & accuracy \\
VideoMMEShortVideoAudioCentricQA & \cite{fu2024video} &  & 4,500 & 20h (72,599s) & 1 & Web & accuracy \\
WorldQAVideoAudioCentricQA & \cite{zhang2024worldqa} &  & 16,372 & 3d (313,887s) & 1 & Web & accuracy \\
WorldSense1MinVideoAudioCentricQA & \cite{hong2025worldsense} &  & 5,165 & 13h (50,231s) & 1 & Web & accuracy \\
\bottomrule
\end{tabular}
}
\caption{Video-audio(-text) multimodal tasks in MVEB (non-retrieval). Tasks use video with audio and optionally text.}
\label{tab:mveb-video-audio-text-tasks}
\end{table*}

\begin{table*}[ht]
\centering
\resizebox{\linewidth}{!}{
\begin{tabular}{lccrrrcc}
\toprule
\textbf{Dataset} & \textbf{Citation} & \textbf{MVEB} & \textbf{N.samples} & \textbf{Total Duration(s)} & \textbf{N.Langs} & \textbf{Domains} & \textbf{Main metric} \\
\midrule
\multicolumn{8}{l}{\textbf{Video Retrieval}} \\
AVMemeExamA2VRetrieval & \cite{jiang2026avmeme} &  & 1,800 & 3h (11,010s) & 1 & Web, Social & ndcg\_at\_10 \\
AVMemeExamAT2VRetrieval & \cite{jiang2026avmeme} & \checkmark & 1,800 & 3h (11,010s) & 1 & Web, Social & ndcg\_at\_10 \\
AVMemeExamT2VARetrieval & \cite{jiang2026avmeme} &  & 1,800 & 3h (11,010s) & 1 & Web, Social & ndcg\_at\_10 \\
AVMemeExamV2ARetrieval & \cite{jiang2026avmeme} &  & 1,800 & 3h (11,010s) & 1 & Web, Social & ndcg\_at\_10 \\
AVMemeExamVA2TRetrieval & \cite{jiang2026avmeme} &  & 1,800 & 3h (11,010s) & 1 & Web, Social & ndcg\_at\_10 \\
AVMemeExamVT2ARetrieval & \cite{jiang2026avmeme} &  & 1,800 & 3h (11,010s) & 1 & Web, Social & ndcg\_at\_10 \\
AudioCapsAVA2VRetrieval & \cite{kim2019audiocaps} &  & 1,330 & 1h (6,541s) & 1 & Encyclopaedic, Web & ndcg\_at\_10 \\
AudioCapsAVAT2VRetrieval & \cite{kim2019audiocaps} &  & 1,330 & 1h (6,541s) & 1 & Encyclopaedic, Web & ndcg\_at\_10 \\
AudioCapsAVT2VARetrieval & \cite{kim2019audiocaps} &  & 1,330 & 1h (6,541s) & 1 & Encyclopaedic, Web & ndcg\_at\_10 \\
AudioCapsAVV2ARetrieval & \cite{kim2019audiocaps} &  & 1,330 & 1h (6,541s) & 1 & Encyclopaedic, Web & ndcg\_at\_10 \\
AudioCapsAVVA2TRetrieval & \cite{kim2019audiocaps} & \checkmark & 1,330 & 1h (6,541s) & 1 & Encyclopaedic, Web & ndcg\_at\_10 \\
AudioCapsAVVT2ARetrieval & \cite{kim2019audiocaps} & \checkmark & 1,330 & 1h (6,541s) & 1 & Encyclopaedic, Web & ndcg\_at\_10 \\
DiDeMoA2VRetrieval & \cite{hendricks2017localizing} &  & 1,998 & 13h (50,032s) & 1 & Web, Spoken & ndcg\_at\_10 \\
DiDeMoAT2VRetrieval & \cite{hendricks2017localizing} &  & 1,998 & 13h (50,032s) & 1 & Web, Spoken & ndcg\_at\_10 \\
DiDeMoT2VARetrieval & \cite{hendricks2017localizing} &  & 1,998 & 13h (50,032s) & 1 & Web, Spoken & ndcg\_at\_10 \\
DiDeMoV2ARetrieval & \cite{hendricks2017localizing} &  & 1,998 & 13h (50,032s) & 1 & Web, Spoken & ndcg\_at\_10 \\
DiDeMoVA2TRetrieval & \cite{hendricks2017localizing} &  & 1,998 & 13h (50,032s) & 1 & Web, Spoken & ndcg\_at\_10 \\
DiDeMoVT2ARetrieval & \cite{hendricks2017localizing} &  & 1,998 & 13h (50,032s) & 1 & Web, Spoken & ndcg\_at\_10 \\
MSRVTTA2V & \cite{xu2016msrvtt} &  & 1,758 & 3h (13,369s) & 1 & Unknown & ndcg\_at\_10 \\
MSRVTTAT2V & \cite{xu2016msrvtt} &  & 1,758 & 3h (13,369s) & 1 & Unknown & ndcg\_at\_10 \\
MSRVTTT2VA & \cite{xu2016msrvtt} &  & 1,758 & 3h (13,369s) & 1 & Unknown & ndcg\_at\_10 \\
MSRVTTV2A & \cite{xu2016msrvtt} &  & 1,758 & 3h (13,369s) & 1 & Unknown & ndcg\_at\_10 \\
MSRVTTVA2T & \cite{xu2016msrvtt} &  & 1,758 & 3h (13,369s) & 1 & Unknown & ndcg\_at\_10 \\
MSRVTTVT2A & \cite{xu2016msrvtt} &  & 1,758 & 3h (13,369s) & 1 & Unknown & ndcg\_at\_10 \\
Panda70MT2VARetrieval & \cite{chen2024panda} &  & 6,790 & 10h (37,488s) & 1 & Web, Spoken & ndcg\_at\_10 \\
Panda70MVA2TRetrieval & \cite{chen2024panda} &  & 6,790 & 10h (37,488s) & 1 & Web, Spoken & ndcg\_at\_10 \\
Shot2Story20KA2VRetrieval & \cite{han2023shot2story} &  & 8,046 & 18h (67,101s) & 1 & Web, Spoken & ndcg\_at\_10 \\
Shot2Story20KAT2VRetrieval & \cite{han2023shot2story} &  & 8,046 & 18h (67,101s) & 1 & Web, Spoken & ndcg\_at\_10 \\
Shot2Story20KT2VARetrieval & \cite{han2023shot2story} &  & 8,046 & 18h (67,101s) & 1 & Web, Spoken & ndcg\_at\_10 \\
Shot2Story20KV2ARetrieval & \cite{han2023shot2story} &  & 8,046 & 18h (67,101s) & 1 & Web, Spoken & ndcg\_at\_10 \\
Shot2Story20KVA2TRetrieval & \cite{han2023shot2story} &  & 8,046 & 18h (67,101s) & 1 & Web, Spoken & ndcg\_at\_10 \\
Shot2Story20KVT2ARetrieval & \cite{han2023shot2story} &  & 8,046 & 18h (67,101s) & 1 & Web, Spoken & ndcg\_at\_10 \\
VALOR32KA2VRetrieval & \cite{chen2023valor} &  & 6,982 & 9h (34,636s) & 1 & Web, Spoken & ndcg\_at\_10 \\
VALOR32KAT2VRetrieval & \cite{chen2023valor} &  & 6,982 & 9h (34,636s) & 1 & Web, Spoken & ndcg\_at\_10 \\
VALOR32KT2VARetrieval & \cite{chen2023valor} & \checkmark & 6,982 & 9h (34,636s) & 1 & Web, Spoken & ndcg\_at\_10 \\
VALOR32KV2ARetrieval & \cite{chen2023valor} &  & 6,982 & 9h (34,636s) & 1 & Web, Spoken & ndcg\_at\_10 \\
VALOR32KVA2TRetrieval & \cite{chen2023valor} &  & 6,982 & 9h (34,636s) & 1 & Web, Spoken & ndcg\_at\_10 \\
VALOR32KVT2ARetrieval & \cite{chen2023valor} &  & 6,982 & 9h (34,636s) & 1 & Web, Spoken & ndcg\_at\_10 \\
VATEXA2VRetrieval & \cite{wang2019vatex} &  & 2,000 & 1d (157,615s) & 1 & Web, Spoken & ndcg\_at\_10 \\
VATEXAT2VRetrieval & \cite{wang2019vatex} &  & 2,000 & 1d (157,615s) & 1 & Web, Spoken & ndcg\_at\_10 \\
VATEXT2VARetrieval & \cite{wang2019vatex} &  & 2,000 & 1d (157,615s) & 1 & Web, Spoken & ndcg\_at\_10 \\
VATEXV2ARetrieval & \cite{wang2019vatex} & \checkmark & 2,000 & 1d (157,615s) & 1 & Web, Spoken & ndcg\_at\_10 \\
VATEXVA2TRetrieval & \cite{wang2019vatex} & \checkmark & 2,000 & 1d (157,615s) & 1 & Web, Spoken & ndcg\_at\_10 \\
VATEXVT2ARetrieval & \cite{wang2019vatex} &  & 2,000 & 1d (157,615s) & 1 & Web, Spoken & ndcg\_at\_10 \\
VGGSoundAVA2VRetrieval & \cite{chen2020vggsound} & \checkmark & 1,392 & 1h (6,952s) & 1 & Web, Spoken & ndcg\_at\_10 \\
VGGSoundAVAT2VRetrieval & \cite{chen2020vggsound} &  & 1,392 & 1h (6,952s) & 1 & Web, Spoken & ndcg\_at\_10 \\
VGGSoundAVT2VARetrieval & \cite{chen2020vggsound} &  & 1,392 & 1h (6,952s) & 1 & Web, Spoken & ndcg\_at\_10 \\
VGGSoundAVV2ARetrieval & \cite{chen2020vggsound} &  & 1,392 & 1h (6,952s) & 1 & Web, Spoken & ndcg\_at\_10 \\
VGGSoundAVVA2TRetrieval & \cite{chen2020vggsound} &  & 1,392 & 1h (6,952s) & 1 & Web, Spoken & ndcg\_at\_10 \\
VGGSoundAVVT2ARetrieval & \cite{chen2020vggsound} &  & 1,392 & 1h (6,952s) & 1 & Web, Spoken & ndcg\_at\_10 \\
YouCook2A2VRetrieval & \cite{zhou2018towards} &  & 6,208 & 17h (61,878s) & 1 & Web, Spoken & ndcg\_at\_10 \\
YouCook2AT2VRetrieval & \cite{zhou2018towards} &  & 6,208 & 17h (61,878s) & 1 & Web, Spoken & ndcg\_at\_10 \\
YouCook2T2VARetrieval & \cite{zhou2018towards} & \checkmark & 6,208 & 17h (61,878s) & 1 & Web, Spoken & ndcg\_at\_10 \\
YouCook2V2ARetrieval & \cite{zhou2018towards} &  & 6,208 & 17h (61,878s) & 1 & Web, Spoken & ndcg\_at\_10 \\
YouCook2VA2TRetrieval & \cite{zhou2018towards} &  & 6,208 & 17h (61,878s) & 1 & Web, Spoken & ndcg\_at\_10 \\
YouCook2VT2ARetrieval & \cite{zhou2018towards} &  & 6,208 & 17h (61,878s) & 1 & Web, Spoken & ndcg\_at\_10 \\
\bottomrule
\end{tabular}
}
\caption{Video-audio(-text) retrieval tasks in MVEB. Tasks use video with audio and optionally text for retrieval.}
\label{tab:mveb-video-audio-text-retrieval-tasks}
\end{table*}

This appendix provides detailed information on all tasks within MVEB, including size, language, metrics, and other relevant details in \autoref{tab:mveb-video-only-tasks}, \autoref{tab:mveb-video-text-tasks}, \autoref{tab:mveb-video-audio-text-tasks}, and \autoref{tab:mveb-video-audio-text-retrieval-tasks}.

\paragraph{Locating each benchmark scope in the tables.} Three nested benchmarks live in the same task pool: MVEB (23 tasks), MVEB(text, video) (19 tasks), and MVEB(video) (9 tasks). Membership is indicated by checkmark columns:
\begin{itemize}
    \item \textbf{MVEB} tasks are checkmarked in the \textbf{MVEB} column of all four sub-tables (\autoref{tab:mveb-video-only-tasks}, \autoref{tab:mveb-video-text-tasks}, \autoref{tab:mveb-video-audio-text-tasks}, \autoref{tab:mveb-video-audio-text-retrieval-tasks}).
    \item \textbf{MVEB(text, video)} tasks are checkmarked in the \textbf{TV} column of \autoref{tab:mveb-video-only-tasks} (video-only tasks) and \autoref{tab:mveb-video-text-tasks} (video-text tasks). The two audio-bearing sub-tables (\autoref{tab:mveb-video-audio-text-tasks}, \autoref{tab:mveb-video-audio-text-retrieval-tasks}) have no TV column because audio-bearing tasks are excluded from MVEB(text, video) by construction.
    \item \textbf{MVEB(video)} tasks are checkmarked in the \textbf{V} column of \autoref{tab:mveb-video-only-tasks} only; the other three sub-tables involve text or audio and are excluded from MVEB(video) by construction.
\end{itemize}

\subsection{Audio--video annotation provenance}
\label{sec:av-annotation}

When a source video carries an audio track, MVEB pairs it into both a video-only (\texttt{v}) and a video$+$audio (\texttt{va}) task variant; \S\ref{sec:audio-contribution} measures the per-task audio delta $\Delta = \text{score}_{\text{va}} - \text{score}_{\text{v}}$ from those paired evaluations. To interpret that $\Delta$, we classify each source dataset by what modalities its annotators worked from. \textbf{AV-grounded} datasets had labels produced from both audio and visual content (audio-visual event labels, dialogue-emotion labels, music QA, multimodal exam-style questions, audio-visual captions). \textbf{V-grounded} datasets had labels produced from visuals alone, even though the source clips may carry audio (action-recognition splits, frame-conditioned captioning). In both cases the source clips still have audio that we feed to the model in the \texttt{va} variant; the provenance distinction is about the labels, not the clips. \autoref{tab:av-provenance} lists the families in each group, and \S\ref{sec:audio-contribution} reports that the provenance predicts whether audio helps a model on a dataset: it helps on AV-grounded and hurts on V-grounded.

\begin{table*}[t]
    \centering
    \footnotesize
    \setlength{\tabcolsep}{6pt}
    \renewcommand{\arraystretch}{1.15}
    \begin{tabular}{l p{0.78\textwidth}}
    \toprule
    \textbf{Provenance} & \textbf{Dataset families} \\
    \midrule
    AV-grounded         & AudioCaps\_AV, AVEDataset, AVMeme, AVMemeExam, AVQA, AVSpeakerBench, DailyOmni, MELD, MusicAVQA, MusicAVQACLS, OmniVideoBench, PerceptionTest, RAVDESS, Shot2Story20K, VALOR-32K, VGGSoundAV, VideoMMEShort, WorldQA, WorldSense, WorldSense1Min, YouCook2\\
    \midrule
    V-grounded          & ActivityNetCaptions, Breakfast, DiDeMo, Diving48, EgoSchema, HMDB51, HumanAnimalCartoon, Kinetics-400, Kinetics-600, Kinetics-700, MSR-VTT, MSVD, NExT-QA, Panda70M, SomethingSomethingV2, TUNA-Bench, UCF101, VATEX, VideoCon, Vinoground\\
    \bottomrule
    \end{tabular}
    \caption{Annotation provenance of MVEB source datasets. AV-grounded families had labels produced from both audio and visual content; V-grounded families had labels produced from visuals alone, even though the source clips often carry audio. Both groups are paired into video-only and video$+$audio task variants in MVEB; the distinction matters for interpreting the per-task audio delta in \S\ref{sec:audio-contribution}.}
    \label{tab:av-provenance}
\end{table*}

\section{Dataset Construction}
\label{sec:dataset-descriptions}

This appendix provides a brief description of each dataset contributed to MVEB by this work, including the source data, the sampling and filtering applied for our benchmark, and the upstream release used to obtain the raw clips. Datasets are grouped by primary task type; a single dataset may be used in multiple task formats (e.g.\ classification, zero-shot classification, and clustering all draw on the same underlying clips). The processed datasets described here are publicly released to support reproducible evaluation.

\subsection{Video Classification}

\textbf{AVE-Dataset} \citep{tian2018audio} is an audio-visual event localization dataset of approximately 4K YouTube clips spanning 28 audio-visual event categories, with each clip approximately 10 seconds long. Used the official train and test splits across 28 audio-visual event classes ($\sim$3{,}312 train / $\sim$402 test). We extracted 16~kHz mono audio via ffmpeg from each clip, dropping clips where extraction failed. The raw video data are retrieved from the AVE official release.

\textbf{AVMeme Exam} \citep{jiang2026avmeme} is a multilingual, multicultural multi-modal benchmark of internet memes whose questions probe cultural and contextual reasoning. Used the \texttt{full}/\texttt{test} split ($\sim$900 examples). We extracted 16~kHz mono audio via ffmpeg from each clip, dropping clips where extraction failed. The raw video data are retrieved from the upstream HF mirror.

\textbf{Breakfast} \citep{kuehne2014language} is a cooking activity dataset of approximately 1.9K crowdsourced videos covering 10 breakfast preparation activities, filmed from multiple camera viewpoints. Used only camera-01 clips, capped at 50 per class across 10 meal classes ($\sim$433 test). The raw video data are retrieved from the Breakfast Actions official release.

\textbf{Human-Animal-Cartoon (HAC)} \citep{dong2023simmmdg} is a domain-generalization dataset spanning human, animal, and cartoon clips of 7 shared action categories. Concatenated the three HAC test-only domain CSVs, capped at 32 per class across 7 action classes ($\sim$644 examples). We extracted 16~kHz mono audio via ffmpeg from each clip, dropping clips where extraction failed. The raw video data are retrieved from the HAC official release.

\textbf{HMDB51} \citep{6126543} is a video dataset for human motion recognition containing approximately 6K clips from movies and online sources, organized into 51 action categories such as catch, drink, kick, and walk. We used the official train/test split~1 across all 51 action classes, yielding $\sim$3{,}570 train and $\sim$1{,}530 test clips. The raw video data are retrieved from the Brown serre-lab official release.

\textbf{Kinetics-400} \citep{kay2017kineticshumanactionvideo} is a large-scale action recognition dataset of approximately 300K YouTube clips spanning 400 human action classes such as cooking, sports, and dance. Each clip is approximately 10 seconds long. We sampled 25 random shard files from each of the official train and test path lists and capped at 10 examples per class per split, yielding $\sim$3{,}970 train and $\sim$4{,}000 test clips. We extracted 16~kHz mono audio via ffmpeg from each clip, dropping clips where extraction failed. The raw video data are retrieved from the CVD Foundation Kinetics downloader.

\textbf{Kinetics-600} \citep{carreira2018short} extends Kinetics-400 to 600 action classes with an expanded YouTube video pool. We sampled the official test split capped at approximately 16 examples per class, yielding $\sim$9{,}576 clips. We extracted 16~kHz mono audio via ffmpeg from each clip, dropping clips where extraction failed. The raw video data are retrieved from the CVD Foundation Kinetics downloader.

\textbf{Kinetics-700-2020} \citep{smaira2020short} is the 2020 release of Kinetics-700 with 700 action classes and refreshed annotations. We sampled the official test split capped at approximately 16 examples per class, yielding $\sim$11{,}190 clips. We extracted 16~kHz mono audio via ffmpeg from each clip, dropping clips where extraction failed. The raw video data are retrieved from the CVD Foundation Kinetics downloader.

\textbf{MELD} \citep{poria2019meld} is a multi-party conversation dataset drawn from the TV show \textit{Friends}, with utterances annotated with emotion and sentiment labels. Sampled the test split ($\sim$2{,}610 examples). We extracted 16~kHz mono audio via ffmpeg from each clip, dropping clips where extraction failed. The raw video data are retrieved from the MELD official release.

\textbf{MUSIC-AVQA} \citep{li2022learning} is an audio-visual question-answering dataset over musical instrument performances. Filtered the test split to rows whose answer is one of the 22 instrument-vocabulary labels ($\sim$1{,}706 examples). We extracted 16~kHz mono audio via ffmpeg from each clip, dropping clips where extraction failed. The raw video data are retrieved from the MUSIC-AVQA official release.

\textbf{RAVDESS} \citep{10.1371/journal.pone.0196391} is a dataset of emotional acted speech and song recordings by 24 professional actors; we use the 1{,}440-clip speech audio-visual subset covering 8 emotion categories. Used the metadata train split (the only split RAVDESS exposes) remapped to its audio-visual ``01-'' filenames ($\sim$1{,}440 examples). We extracted 16~kHz mono audio via ffmpeg from each clip, dropping clips where extraction failed. The raw video data are retrieved from the RAVDESS Zenodo release.

\textbf{Something-Something v2} \citep{goyal2017something} is a dataset of approximately 220K crowdsourced clips showing fine-grained physical interactions between humans and objects, with 174 action classes. Used the official validation as test, capped at 32 per class across 174 classes ($\sim$5{,}444 examples). The raw video data are retrieved from the Qualcomm SSv2 release.

\textbf{UCF101-51VA} \citep{Soomro2012UCF101} is an action recognition dataset of approximately 13K YouTube videos across 101 action categories. We restricted to a 51-class audio-bearing subset (the ``51VA'' partition) and used the official train/test split~1, yielding $\sim$4{,}890 train and $\sim$1{,}944 test clips. We extracted 16~kHz mono audio via ffmpeg from each clip, dropping clips where extraction failed. The raw video data are retrieved from the CRCV UCF official release.

\textbf{VGGSound} \citep{chen2020vggsound} is a large-scale audio-visual dataset of approximately 200K YouTube clips with sound-event class labels. Filtered the test split, capped at 32 per class across 309 classes ($\sim$9{,}888 examples). We extracted 16~kHz mono audio via ffmpeg from each clip, dropping clips where extraction failed. The raw video data are retrieved from the VGGSound official release.

\textbf{WorldSense} \citep{hong2025worldsense} is a multi-modal video understanding benchmark spanning multiple domains and durations, with a short-clip subset used here. Filtered the test split to short ($<$1\,min) clips ($\sim$1{,}047 examples). We extracted 16~kHz mono audio via ffmpeg from each clip, dropping clips where extraction failed. The raw video data are retrieved from the WorldSense official release.

\subsection{Video-Centric QA}

\textbf{AVMeme Exam} \citep{jiang2026avmeme} is a multilingual, multicultural multi-modal benchmark of internet memes whose questions probe cultural and contextual reasoning. Same data as the classification variant ($\sim$900 examples), evaluated as multiple-choice retrieval over the candidate set. We extracted 16~kHz mono audio via ffmpeg from each clip, dropping clips where extraction failed. The raw video data are retrieved from the upstream HF mirror.

\textbf{AVQA} \citep{yang2022avqa} is an audio-visual multiple-choice question answering benchmark. Used the test QA split keyed by \texttt{(video, question)} since one clip can carry multiple questions ($\sim$921 examples). We extracted 16~kHz mono audio via ffmpeg from each clip, dropping clips where extraction failed. The raw video data are retrieved from the upstream AVQA HF mirror.

\textbf{Daily-Omni} \citep{zhou2025dailyomni} is a multi-modal QA benchmark over everyday-life videos. Used the upstream evaluation split keyed by \texttt{(video, question)} ($\sim$1{,}200 examples). We extracted 16~kHz mono audio via ffmpeg from each clip, dropping clips where extraction failed. The raw video data are retrieved from the Daily-Omni HF mirror.

\textbf{EgoSchema} \citep{mangalam2023egoschema} is an egocentric (first-person) long-form video question-answering benchmark. Used the \texttt{Subset} config test split (500 examples). The raw video data are retrieved from the EgoSchema official release.

\textbf{NExT-QA} \citep{xiao2021next} is a video question answering benchmark focused on causal and temporal reasoning over untrimmed daily-activity videos, supporting both multiple-choice and open-ended QA. We use the multiple-choice portion: we sampled the test split, yielding $\sim$993 examples. We extracted 16~kHz mono audio via ffmpeg from each clip, dropping clips where extraction failed. The raw video data are retrieved from the upstream HF mirror.

\textbf{OmniVideoBench} \citep{li2025omnivideobench} is a video QA benchmark spanning a range of clip durations and content types. Filtered the test split to clips under 5 minutes ($\sim$509 examples). We extracted 16~kHz mono audio via ffmpeg from each clip, dropping clips where extraction failed. The raw video data are retrieved from the OmniVideoBench HF mirror.

\textbf{Perception Test} \citep{patraucean2024perception} is a multi-modal benchmark probing perception and reasoning over diverse video clips. Used the \texttt{mc\_question\_val} validation split ($\sim$938 examples). We extracted 16~kHz mono audio via ffmpeg from each clip, dropping clips where extraction failed. The raw video data are retrieved from the Perception Test HF mirror.

\textbf{Video-MME (short)} \citep{fu2024video} is a comprehensive video-MLLM benchmark with multi-domain clips at three duration buckets. Filtered the test split to \texttt{duration == "short"} clips, deterministically shuffled (seed=42), capped at 2{,}000 ($\sim$900 examples). We extracted 16~kHz mono audio via ffmpeg from each clip, dropping clips where extraction failed. The raw video data are retrieved from the Video-MME HF mirror.

\textbf{WorldQA} \citep{zhang2024worldqa} is a long-horizon video QA benchmark drawn from real-world activities. Used the \texttt{MC} config test split keyed by \texttt{(video, question\_idx)} ($\sim$3{,}284 examples). We extracted 16~kHz mono audio via ffmpeg from each clip, dropping clips where extraction failed. The raw video data are retrieved from the WorldQA HF mirror.

\textbf{WorldSense (VCQA)} \citep{hong2025worldsense} is a multi-modal video understanding benchmark spanning multiple domains and durations. Same data as the WorldSense classification variant, evaluated as multiple-choice retrieval ($\sim$1{,}047 examples). We extracted 16~kHz mono audio via ffmpeg from each clip, dropping clips where extraction failed. The raw video data are retrieved from the WorldSense official release.

\subsection{Video Retrieval}

\textbf{ActivityNet Captions} \citep{krishna2017dense} is a dense video captioning dataset of approximately 20K untrimmed videos with temporally-localized natural-language captions. Used the upstream \texttt{val2} split as the test split ($\sim$4{,}884 examples). The raw video data are retrieved from the ActivityNet Captions official release.

\textbf{AudioCaps-AV} \citep{kim2019audiocaps} is a dataset of approximately 46K 10-second YouTube clips with human-written audio captions, derived from AudioSet. Sampled the test split ($\sim$665 examples). We extracted 16~kHz mono audio via ffmpeg from each clip, dropping clips where extraction failed. The raw video data are retrieved from the upstream HF AudioCaps mirror joined with a Kaggle video archive.

\textbf{DiDeMo} \citep{hendricks2017localizing} is a video moment localization dataset with natural-language descriptions of short temporally-grounded events. Used the test split ($\sim$999 examples). We extracted 16~kHz mono audio via ffmpeg from each clip, dropping clips where extraction failed. The raw video data are retrieved from the DiDeMo official release.

\textbf{MSR-VTT} \citep{xu2016msrvtt} is a video description dataset containing 10K video clips and 200K human-written captions covering 20 categories. We used the standard 1{,}000-clip retrieval test split (\texttt{msrvtt\_ret\_test1k}), yielding 879 final clips after filtering. We extracted 16~kHz mono audio via ffmpeg from each clip, dropping clips where extraction failed. The raw video data are retrieved from a Kaggle mirror of the official MSR-VTT test partition.

\textbf{MSVD} \citep{microsoft2011msvd} is a dataset of approximately 2K short YouTube clips with multiple paraphrased English captions per clip. Used the test split with the first caption per clip ($\sim$660 examples). The raw video data are retrieved from the UT Austin YouTubeClips release.

\textbf{PANDA-70M} \citep{chen2024panda} is a large-scale collection of YouTube clips paired with dense automatic captions. Sampled the test split ($\sim$3{,}395 examples). We extracted 16~kHz mono audio via ffmpeg from each clip, dropping clips where extraction failed. The raw video data are retrieved from the PANDA-70M official release.

\textbf{Shot2Story-20K} \citep{han2023shot2story} is a dataset of approximately 20K multi-shot videos paired with shot-level captions, narration captions, and multi-shot video summaries. Used the test split ($\sim$4{,}023 examples). We extracted 16~kHz mono audio via ffmpeg from each clip, dropping clips where extraction failed. The raw video data are retrieved from the Shot2Story-20K official release.

\textbf{TUNA-Bench} \citep{ye2025tuna} is a dense fine-grained temporal video captioning benchmark. Used the \texttt{TUNA-1K} config test split ($\sim$1{,}000 examples). The raw video data are retrieved from the TUNA-Bench HF mirror.

\textbf{VALOR-32K} \citep{chen2023valor} is a dataset of 32K paired audio-visual-language samples for cross-modal alignment. Used the official \texttt{desc\_test.json} test split ($\sim$3{,}491 examples). We extracted 16~kHz mono audio via ffmpeg from each clip, dropping clips where extraction failed. The raw video data are retrieved from the VALOR-32K official release.

\textbf{VATEX} \citep{wang2019vatex} is a multilingual video captioning dataset of approximately 41K video clips drawn from Kinetics, paired with 10 captions each in English and Chinese. Used the \texttt{vatex\_test} config with the first English caption per clip, capped at 1{,}000 ($\sim$1{,}000 examples). We extracted 16~kHz mono audio via ffmpeg from each clip, dropping clips where extraction failed. The raw video data are retrieved from the VATEX official release.

\textbf{VGGSound-AV Retrieval} \citep{chen2020vggsound} is a text-annotated extension of VGGSound that pairs each clip with separate visual and audio captions, repurposed here for audio-visual retrieval evaluation. Used the VGGSound-T2AV test split ($\sim$696 examples). We extracted 16~kHz mono audio via ffmpeg from each clip, dropping clips where extraction failed. The raw video data are retrieved from the VGGSound official release.

\textbf{YouCook2} \citep{zhou2018towards} is a dataset of approximately 2K untrimmed instructional cooking videos with temporally-localized step-level annotations. Used the official validation annotations cut into recipe-step segments (3{,}104 examples). We extracted 16~kHz mono audio via ffmpeg from each clip, dropping clips where extraction failed. The raw video data are retrieved from the YouCook2 official release.
\section{Overview of Models}
\label{appdx:models}

All models used in the evaluations are listed in \autoref{tab:all-models}.

\begin{table*}[ht]
\centering
\begin{tabular*}{\textwidth}{l@{\extracolsep{\fill}}rp{2cm}}
\toprule
\textbf{Model} & \textbf{Parameters (M)} & \textbf{Modalities} \\
\midrule
BidirLM/BidirLM-Omni-2.5B-Embedding & 2445.0 & a, i, t, v \\
Haon-Chen/e5-omni-3B \cite{chen2026e5omni} & 4703.5 & a, i, t, v \\
Haon-Chen/e5-omni-7B \cite{chen2026e5omni} & 8931.8 & a, i, t, v \\
LCO-Embedding/LCO-Embedding-Omni-3B \cite{xiaoscaling} & 4703.5 & a, i, t, v \\
LCO-Embedding/LCO-Embedding-Omni-7B \cite{xiaoscaling} & 8931.8 & a, i, t, v \\
Qwen/Qwen2.5-Omni-3B \cite{xu2025qwen25omnitechnicalreport} & 5537.1 & a, i, t, v \\
Qwen/Qwen2.5-Omni-7B \cite{xu2025qwen25omnitechnicalreport} & 10732.2 & a, i, t, v \\
Qwen/Qwen3-VL-Embedding-2B & 2127.5 & i, t, v \\
Qwen/Qwen3-VL-Embedding-8B & 8144.8 & i, t, v \\
Qwen/Qwen3-Omni-30B-A3B-Captioner$^{\dagger}$ \cite{xu2025qwen3omnitechnicalreport} & 31719.2 & a, i, t, v \\
Qwen/Qwen3-Omni-30B-A3B-Instruct$^{\dagger}$ \cite{xu2025qwen3omnitechnicalreport} & 35259.8 & a, i, t, v \\
Qwen/Qwen3-Omni-30B-A3B-Thinking$^{\dagger}$ \cite{xu2025qwen3omnitechnicalreport} & 31719.2 & a, i, t, v \\
Tevatron/OmniEmbed-v0.1 \cite{zhuang2025tevatron} & 8931.8 & a, i, t, v \\
encord-team/ebind-audio-vision \cite{broadbent2025ebindpracticalapproachspace} & 764.2 & a, i, t, v \\
encord-team/ebind-full \cite{broadbent2025ebindpracticalapproachspace} & 1790.0 & a, i, t, v \\
encord-team/ebind-points-vision \cite{broadbent2025ebindpracticalapproachspace} & 1694.2 & i, t, v \\
facebook/pe-av-base \cite{vyas2025pushingfrontieraudiovisualperception} & 1033.7 & a, t, v \\
facebook/pe-av-base-16-frame$^{\dagger}$ \cite{vyas2025pushingfrontieraudiovisualperception} & 1033.7 & a, t, v \\
facebook/pe-av-large \cite{vyas2025pushingfrontieraudiovisualperception} & 2234.7 & a, t, v \\
facebook/pe-av-large-16-frame$^{\dagger}$ \cite{vyas2025pushingfrontieraudiovisualperception} & 2234.7 & a, t, v \\
facebook/pe-av-small \cite{vyas2025pushingfrontieraudiovisualperception} & 847.0 & a, t, v \\
facebook/pe-av-small-16-frame$^{\dagger}$ \cite{vyas2025pushingfrontieraudiovisualperception} & 847.0 & a, t, v \\
facebook/vjepa2-vitg-fpc32-384-diving48 & 1127.9 & v \\
facebook/vjepa2-vitg-fpc64-256 & 1034.6 & v \\
facebook/vjepa2-vitg-fpc64-384 & 1034.6 & v \\
facebook/vjepa2-vitg-fpc64-384-ssv2 & 1128.0 & v \\
facebook/vjepa2-vith-fpc64-256 & 653.9 & v \\
facebook/vjepa2-vitl-fpc16-256-ssv2 & 375.5 & v \\
facebook/vjepa2-vitl-fpc32-256-diving48 & 375.4 & v \\
facebook/vjepa2-vitl-fpc64-256 & 326.0 & v \\
jinaai/jina-embeddings-v5-omni-nano & 986.0 & a, i, t, v \\
jinaai/jina-embeddings-v5-omni-small & 1626.3 & a, i, t, v \\
microsoft/xclip-base-patch16 \cite{ni2022expanding} & 194.9 & t, v \\
microsoft/xclip-base-patch32 \cite{ni2022expanding} & 196.6 & t, v \\
microsoft/xclip-large-patch14 \cite{ni2022expanding} & 575.7 & t, v \\
VLM2Vec/VLM2Vec-V2.0 \cite{meng2025vlm2vecv2} & 2209.0 & i, t, v \\
nvidia/omni-embed-nemotron-3b \cite{xu2025omni} & 4703.5 & a, i, t, v \\
nvidia/omnivinci$^{\dagger}$ \cite{ye2025omnivinci} & 8742.9 & a, i, t, v \\
zhibinlan/UME-R1-2B \cite{lan2025ume} & 2209.0 & i, t, v \\
zhibinlan/UME-R1-7B \cite{lan2025ume} & 8291.4 & i, t, v \\
\bottomrule
\end{tabular*}
\caption{Models registered in MVEB. Model sizes are in millions of parameters. $^{\dagger}$Registered in the MTEB model registry but not yet evaluated on MVEB.}
\label{tab:all-models}
\end{table*}      
\section{Per-Task Results}
\label{appdx:results}

This appendix reports raw per-task scores for every model on every MVEB
task. Each cell is the \texttt{main\_score} as defined by the upstream
\texttt{mteb} task (nDCG@10 for retrieval, accuracy for classification and
zero-shot, $v$-measure for clustering, max-AP for pair classification, accuracy
for question answering), scaled to 0--100. \textbf{Bold} = best per column within a table. A \texttt{--} cell means
the task variant is incompatible with the model's input modalities (for
example, an \texttt{audio + video} variant on a model without an audio
encoder such as X-CLIP or V-JEPA-2, or any retrieval direction with audio
in the query/target for a text-video model). Models with no scores at all
on a given task family are hidden from that family's table; the full
roster is in \autoref{tab:all-models}. Tables are
auto-generated from \texttt{results/results/<model>/<rev>/*.json}; do not
edit by hand.

\subsection{Modality-restricted leaderboards}
\label{appdx:results-scope-variants}

Two modality-restricted leaderboards accompany MVEB for models that cannot accept the full input surface:
\texttt{MVEB(text, video)} for models without an audio encoder and
\texttt{MVEB(video)} for video-only encoders. Each table lists only the models compatible with that input surface, ranked among themselves; these are not parallel benchmarks to MVEB but accommodations for restricted modality coverage.

\begin{table*}[!th]
    \centering
    \resizebox{\textwidth}{!}{\setlength{\tabcolsep}{4pt}{\footnotesize
    \begin{tabular}{lll|c|cc|cccccc}
    \toprule
     & & & \textbf{Rank} ($\downarrow$) & \multicolumn{2}{c|}{\textbf{Average}} & \multicolumn{6}{c}{\textbf{Average per Category}} \\
    \cmidrule(r){4-4} \cmidrule(lr){5-6} \cmidrule(l){7-12}
    \textbf{Model} & \textbf{Type} & \textbf{Params} & MVEB(text, video) & Mean & Macro & \textbf{Retr} & \textbf{QA} & \textbf{Cls} & \textbf{Clust} & \textbf{Pair} & \textbf{ZS} \\
    \midrule
    \textcolor{gray}{Number of tasks} & & & & & & \textcolor{gray}{(8)} & \textcolor{gray}{(1)} & \textcolor{gray}{(4)} & \textcolor{gray}{(1)} & \textcolor{gray}{(1)} & \textcolor{gray}{(4)} \\
    \midrule
    Qwen3-VL-Embedding-8B & MLLM-based embedding & 8.1B & \textbf{1} & \textbf{60.9} & \textbf{53.7} & \textbf{69.2} & 27.3 & \textbf{57.5} & 22.2 & 86.7 & \textbf{59.2} \\
    Qwen3-VL-Embedding-2B & MLLM-based embedding & 2.1B & 2 & 58.1 & 52.6 & 65.7 & 26.7 & 54.8 & \textbf{24.9} & \textbf{88.3} & 55.0 \\
    LCO-Embedding-Omni-7B & MLLM-based embedding & 8.9B & 3 & 56.8 & 51.8 & 62.3 & 30.8 & 54.3 & 20.8 & 85.9 & 56.4 \\
    LCO-Embedding-Omni-3B & MLLM-based embedding & 4.7B & 4 & 54.8 & 50.8 & 59.1 & \textbf{32.8} & 54.0 & 17.6 & 87.7 & 53.7 \\
    ebind-points-vision & Multimodal binding & 1.7B & 5 & 53.8 & 44.8 & 62.9 & 22.2 & 46.2 & 1.4 & 78.2 & 57.9 \\
    UME-R1-7B & MLLM-based embedding & 8.3B & 6 & 53.3 & 46.7 & 62.5 & 26.7 & 48.1 & 8.8 & 84.0 & 50.0 \\
    ebind-audio-vision & Multimodal binding & 764M & 7 & 53.8 & 44.8 & 62.9 & 22.2 & 46.2 & 1.4 & 78.2 & 57.9 \\
    ebind-full & Multimodal binding & 1.8B & 7 & 53.8 & 44.8 & 62.9 & 22.2 & 46.2 & 1.4 & 78.2 & 57.9 \\
    e5-omni-7B & MLLM-based embedding & 8.9B & 9 & 54.1 & 46.5 & 63.6 & 26.5 & 45.3 & 3.5 & 84.3 & 55.8 \\
    pe-av-large & Audio-visual contrastive & 2.2B & 10 & 52.4 & 46.1 & 60.3 & 26.3 & 45.3 & 11.5 & 79.2 & 53.6 \\
    BidirLM-Omni-2.5B-Embedding & MLLM-based embedding & 2.4B & 11 & 52.0 & 48.7 & 56.3 & 26.9 & 48.5 & 21.6 & 85.7 & 53.4 \\
    UME-R1-2B & MLLM-based embedding & 2.2B & 12 & 51.5 & 46.4 & 57.6 & 25.1 & 48.0 & 14.9 & 82.1 & 50.9 \\
    OmniEmbed-v0.1 & MLLM-based embedding & 8.9B & 13 & 51.3 & 45.8 & 57.6 & 25.9 & 49.0 & 10.6 & 81.8 & 50.1 \\
    pe-av-base & Audio-visual contrastive & 1.0B & 14 & 49.7 & 43.7 & 57.0 & 25.1 & 45.4 & 10.7 & 75.2 & 49.0 \\
    pe-av-small & Audio-visual contrastive & 847M & 15 & 50.2 & 44.1 & 56.1 & 23.6 & 45.0 & 9.0 & 76.5 & 54.2 \\
    e5-omni-3B & MLLM-based embedding & 4.7B & 16 & 48.4 & 43.5 & 57.9 & 28.3 & 45.1 & 8.0 & 82.1 & 39.5 \\
    xclip-large-patch14 & Video-text contrastive & 576M & 17 & 42.9 & 41.5 & 37.0 & 24.2 & 48.2 & 8.2 & 77.3 & 54.1 \\
    VLM2Vec-V2.0 & MLLM-based embedding & 2.2B & 18 & 44.9 & 41.0 & 50.0 & 29.1 & 40.6 & 0.2 & 81.3 & 45.0 \\
    omni-embed-nemotron-3b & MLLM-based embedding & 4.7B & 19 & 35.8 & 38.4 & 29.3 & 25.1 & 43.7 & 10.9 & 82.9 & 38.3 \\
    xclip-base-patch16 & Video-text contrastive & 195M & 20 & 38.4 & 38.8 & 30.6 & 22.4 & 41.9 & 10.6 & 74.9 & 52.3 \\
    xclip-base-patch32 & Video-text contrastive & 197M & 21 & 35.9 & 38.0 & 26.9 & 26.5 & 37.2 & 11.1 & 74.7 & 51.6 \\
    Qwen2.5-Omni-7B & Generative MLLM & 10.7B & 22 & 10.4 & 18.9 & 0.4 & 25.1 & 11.8 & 7.4 & 53.1 & 15.5 \\
    jina-embeddings-v5-omni-nano & MLLM-based embedding & 986M & 23 & 7.7 & 15.8 & 0.4 & 26.3 & 8.9 & 1.0 & 51.5 & 6.9 \\
    Qwen2.5-Omni-3B & Generative MLLM & 5.5B & 24 & 7.8 & 17.0 & 0.4 & 24.4 & 10.6 & 7.9 & 54.5 & 4.1 \\
    jina-embeddings-v5-omni-small & MLLM-based embedding & 1.6B & 25 & 8.2 & 16.2 & 0.4 & 24.4 & 8.4 & 1.2 & 52.4 & 10.2 \\
    \bottomrule
    \end{tabular}}}
    \caption{Models on \textbf{MVEB(text, video)} ranked by Borda count over 19 tasks. \textbf{Mean} is the arithmetic mean over the model's evaluated tasks; \textbf{Macro} is the macro-average across task categories (each category weighted equally regardless of task count). Task categories: Retr (retrieval; per-direction breakdown in appendix), QA, Cls (classification), Clust (clustering), Pair (pair classification), ZS (zero-shot classification). Model types are defined in \S\ref{sec:methodology-models}. \textbf{Bold} = best per column.}
    \label{tab:mveb-text-video-results}
\end{table*}

\begin{table*}[!th]
    \centering
    \resizebox{\textwidth}{!}{\setlength{\tabcolsep}{4pt}{\footnotesize
    \begin{tabular}{lll|c|cc|cc}
    \toprule
     & & & \textbf{Rank} ($\downarrow$) & \multicolumn{2}{c|}{\textbf{Average}} & \multicolumn{2}{c}{\textbf{Average per Category}} \\
    \cmidrule(r){4-4} \cmidrule(lr){5-6} \cmidrule(l){7-8}
    \textbf{Model} & \textbf{Type} & \textbf{Params} & MVEB(video) & Mean & Macro & \textbf{Cls} & \textbf{Pair} \\
    \midrule
    \textcolor{gray}{Number of tasks} & & & & & & \textcolor{gray}{(6)} & \textcolor{gray}{(3)} \\
    \midrule
    LCO-Embedding-Omni-3B & MLLM-based embedding & 4.7B & \textbf{1} & 61.6 & 65.1 & 54.4 & \textbf{75.9} \\
    LCO-Embedding-Omni-7B & MLLM-based embedding & 8.9B & 2 & 61.7 & 65.0 & 55.1 & 74.9 \\
    Qwen3-VL-Embedding-2B & MLLM-based embedding & 2.1B & 2 & 62.3 & 65.6 & 55.6 & 75.7 \\
    Qwen3-VL-Embedding-8B & MLLM-based embedding & 8.1B & 4 & \textbf{63.5} & \textbf{66.3} & \textbf{58.0} & 74.6 \\
    xclip-large-patch14 & Video-text contrastive & 576M & 5 & 58.6 & 62.3 & 51.0 & 73.7 \\
    UME-R1-2B & MLLM-based embedding & 2.2B & 6 & 57.4 & 61.2 & 49.9 & 72.4 \\
    UME-R1-7B & MLLM-based embedding & 8.3B & 7 & 57.5 & 61.2 & 50.0 & 72.4 \\
    BidirLM-Omni-2.5B-Embedding & MLLM-based embedding & 2.4B & 7 & 58.0 & 62.0 & 50.1 & 73.8 \\
    ebind-full & Multimodal binding & 1.8B & 9 & 55.8 & 59.4 & 48.6 & 70.3 \\
    ebind-audio-vision & Multimodal binding & 764M & 9 & 55.8 & 59.4 & 48.6 & 70.3 \\
    ebind-points-vision & Multimodal binding & 1.7B & 11 & 55.8 & 59.4 & 48.6 & 70.3 \\
    OmniEmbed-v0.1 & MLLM-based embedding & 8.9B & 12 & 57.9 & 61.3 & 51.2 & 71.4 \\
    xclip-base-patch16 & Video-text contrastive & 195M & 13 & 55.6 & 59.8 & 47.3 & 72.2 \\
    e5-omni-3B & MLLM-based embedding & 4.7B & 14 & 55.7 & 59.3 & 48.3 & 70.4 \\
    e5-omni-7B & MLLM-based embedding & 8.9B & 15 & 55.7 & 59.4 & 48.2 & 70.7 \\
    pe-av-large & Audio-visual contrastive & 2.2B & 16 & 55.2 & 59.2 & 47.2 & 71.2 \\
    omni-embed-nemotron-3b & MLLM-based embedding & 4.7B & 16 & 54.7 & 58.8 & 46.3 & 71.3 \\
    VLM2Vec-V2.0 & MLLM-based embedding & 2.2B & 18 & 52.9 & 56.8 & 45.0 & 68.5 \\
    pe-av-base & Audio-visual contrastive & 1.0B & 19 & 53.8 & 58.2 & 45.0 & 71.3 \\
    xclip-base-patch32 & Video-text contrastive & 197M & 20 & 53.0 & 57.6 & 43.8 & 71.5 \\
    pe-av-small & Audio-visual contrastive & 847M & 21 & 53.3 & 57.6 & 44.8 & 70.4 \\
    vjepa2-vitg-fpc32-384-diving48 & Self-supervised video & 1.1B & 22 & 46.2 & 50.3 & 38.0 & 62.5 \\
    vjepa2-vitl-fpc16-256-ssv2 & Self-supervised video & 375M & 23 & 44.6 & 49.1 & 35.7 & 62.5 \\
    vjepa2-vitl-fpc64-256 & Self-supervised video & 326M & 24 & 45.2 & 49.8 & 36.0 & 63.5 \\
    vjepa2-vitg-fpc64-256 & Self-supervised video & 1.0B & 25 & 45.9 & 49.8 & 38.0 & 61.7 \\
    vjepa2-vitg-fpc64-384 & Self-supervised video & 1.0B & 26 & 45.6 & 49.5 & 37.9 & 61.0 \\
    vjepa2-vitg-fpc64-384-ssv2 & Self-supervised video & 1.1B & 26 & 45.6 & 49.5 & 37.9 & 61.0 \\
    vjepa2-vitl-fpc32-256-diving48 & Self-supervised video & 375M & 28 & 44.8 & 49.4 & 35.5 & 63.2 \\
    vjepa2-vith-fpc64-256 & Self-supervised video & 654M & 29 & 42.5 & 46.8 & 34.0 & 59.5 \\
    Qwen2.5-Omni-3B & Generative MLLM & 5.5B & 30 & 30.1 & 36.4 & 17.5 & 55.3 \\
    Qwen2.5-Omni-7B & Generative MLLM & 10.7B & 31 & 30.7 & 36.6 & 18.9 & 54.3 \\
    jina-embeddings-v5-omni-nano & MLLM-based embedding & 986M & 32 & 24.9 & 31.3 & 12.0 & 50.6 \\
    jina-embeddings-v5-omni-small & MLLM-based embedding & 1.6B & 33 & 24.7 & 31.2 & 11.6 & 50.8 \\
    \bottomrule
    \end{tabular}}}
    \caption{Models on \textbf{MVEB(video)} ranked by Borda count over 9 tasks. \textbf{Mean} is the arithmetic mean over the model's evaluated tasks; \textbf{Macro} is the macro-average across task categories (each category weighted equally regardless of task count). Task categories: Cls (classification), Pair (pair classification). Model types are defined in \S\ref{sec:methodology-models}. \textbf{Bold} = best per column.}
    \label{tab:mveb-video-results}
\end{table*}

\subsection{MVEB+ aggregate}
\label{appdx:results-extended}

Table~\ref{tab:mveb-extended-results} reports the aggregate over the full MVEB+ pool (all 184 tasks). MVEB is a correlation-preserving subset of MVEB+ (Pearson $r$=0.996, Spearman $\rho$=0.944 between the two; \autoref{sec:benchmark-construction}). This aggregate is included as a transparency artifact, showing that the 23-task MVEB does not distort relative model performance versus running every task in the pool; it is not framed as a separate leaderboard.

\begin{table*}[!th]
    \centering
    \resizebox{\textwidth}{!}{\setlength{\tabcolsep}{4pt}{\footnotesize
    \begin{tabular}{lll|c|cc|cccccc}
    \toprule
     & & & \textbf{Rank} ($\downarrow$) & \multicolumn{2}{c|}{\textbf{Average}} & \multicolumn{6}{c}{\textbf{Average per Category}} \\
    \cmidrule(r){4-4} \cmidrule(lr){5-6} \cmidrule(l){7-12}
    \textbf{Model} & \textbf{Type} & \textbf{Params} & MVEB+ & Mean & Macro & \textbf{Retr} & \textbf{QA} & \textbf{Cls} & \textbf{Clust} & \textbf{Pair} & \textbf{ZS} \\
    \midrule
    \textcolor{gray}{Number of tasks} & & & & & & \textcolor{gray}{(82)} & \textcolor{gray}{(20)} & \textcolor{gray}{(28)} & \textcolor{gray}{(15)} & \textcolor{gray}{(13)} & \textcolor{gray}{(26)} \\
    \midrule
    LCO-Embedding-Omni-7B & MLLM-based embedding & 8.9B & \textbf{1} & \textbf{58.7} & \textbf{58.1} & \textbf{61.0} & \textbf{51.8} & \textbf{57.4} & 46.5 & 73.9 & \textbf{58.3} \\
    LCO-Embedding-Omni-3B & MLLM-based embedding & 4.7B & 2 & 56.3 & 56.6 & 57.3 & 49.8 & 55.9 & \textbf{46.7} & \textbf{74.3} & 55.6 \\
    e5-omni-7B & MLLM-based embedding & 8.9B & 3 & 56.2 & 54.7 & 60.8 & 46.2 & 52.2 & 43.2 & 72.4 & 53.4 \\
    ebind-audio-vision & Multimodal binding & 764M & 4 & 54.1 & 53.3 & 57.3 & 36.4 & 50.8 & 45.3 & 71.9 & 58.1 \\
    ebind-full & Multimodal binding & 1.8B & 5 & 54.1 & 53.3 & 57.2 & 36.4 & 50.8 & 45.3 & 71.9 & 58.1 \\
    BidirLM-Omni-2.5B-Embedding & MLLM-based embedding & 2.4B & 6 & 51.6 & 53.3 & 49.4 & 40.4 & 56.9 & 45.2 & 73.9 & 54.1 \\
    pe-av-large & Audio-visual contrastive & 2.2B & 7 & 52.7 & 51.1 & 57.7 & 30.7 & 50.7 & 42.8 & 72.0 & 52.6 \\
    pe-av-base & Audio-visual contrastive & 1.0B & 8 & 51.3 & 50.4 & 55.0 & 30.5 & 50.1 & 41.9 & 72.1 & 52.7 \\
    OmniEmbed-v0.1 & MLLM-based embedding & 8.9B & 9 & 49.7 & 50.4 & 50.6 & 37.5 & 53.3 & 43.9 & 72.3 & 44.8 \\
    pe-av-small & Audio-visual contrastive & 847M & 10 & 50.8 & 50.2 & 53.7 & 30.0 & 50.2 & 41.5 & 72.0 & 53.5 \\
    e5-omni-3B & MLLM-based embedding & 4.7B & 11 & 49.2 & 50.1 & 50.6 & 42.7 & 48.6 & 43.6 & 72.0 & 43.2 \\
    omni-embed-nemotron-3b & MLLM-based embedding & 4.7B & 12 & 43.7 & 47.4 & 39.7 & 38.2 & 51.8 & 43.9 & 72.5 & 38.5 \\
    jina-embeddings-v5-omni-nano & MLLM-based embedding & 986M & 13 & 16.6 & 24.4 & 7.2 & 29.7 & 16.9 & 20.7 & 57.3 & 14.7 \\
    jina-embeddings-v5-omni-small & MLLM-based embedding & 1.6B & 14 & 15.8 & 24.4 & 5.0 & 28.3 & 17.2 & 22.3 & 57.9 & 15.5 \\
    Qwen2.5-Omni-7B & Generative MLLM & 10.7B & 15 & 11.6 & 20.7 & 0.4 & 23.9 & 14.2 & 20.9 & 56.5 & 8.4 \\
    Qwen2.5-Omni-3B & Generative MLLM & 5.5B & 16 & 11.3 & 20.5 & 0.5 & 25.2 & 14.2 & 20.6 & 58.0 & 4.8 \\
    \bottomrule
    \end{tabular}}}
    \caption{Models on \textbf{MVEB+} ranked by Borda count over 184 tasks. \textbf{Mean} is the arithmetic mean over the model's evaluated tasks; \textbf{Macro} is the macro-average across task categories (each category weighted equally regardless of task count). Task categories: Retr (retrieval; per-direction breakdown in appendix), QA, Cls (classification), Clust (clustering), Pair (pair classification), ZS (zero-shot classification). Model types are defined in \S\ref{sec:methodology-models}. \textbf{Bold} = best per column.}
    \label{tab:mveb-extended-results}
\end{table*}

\subsection{Retrieval}
\label{appdx:results-retrieval}

One table per query/target direction (\autoref{tab:results-retrieval-t2v}, \autoref{tab:results-retrieval-a2v}, \autoref{tab:results-retrieval-at2v}, \autoref{tab:results-retrieval-v2t}, \autoref{tab:results-retrieval-va2t}, \autoref{tab:results-retrieval-v2a}, \autoref{tab:results-retrieval-vt2a}, \autoref{tab:results-retrieval-t2va}). Two patterns are consistent across the eight directions. (1)~On purely text$\leftrightarrow$video directions (T$\to$V, V$\to$T), Qwen3-VL-Embedding-8B leads at 75.8 and 72.6 nDCG@10 respectively, ahead of eBind, the Perception Encoder audio-visual family, and the omni MLLM-embed models; the smaller Qwen3-VL-Embedding-2B comes in second on T$\to$V. (2)~On audio-conditioned directions (A$\to$V, AT$\to$V, V$\to$A, VT$\to$A, VA$\to$T, T$\to$VA), Qwen3-VL is unable to participate (no audio path) and the audio-capable families share the lead: LCO-Embedding-Omni-7B wins V$\to$A (44.5), VT$\to$A (53.8), VA$\to$T (64.5), and T$\to$VA (64.7); pe-av-large wins A$\to$V (52.8); eBind wins AT$\to$V (74.8). Generative MLLMs (Qwen2.5-Omni) score near zero on every retrieval direction, including text$\to$video; embedding-specific training is essential for retrieval.

\begin{table*}[ht]
\centering
\small
\setlength{\tabcolsep}{4pt}
\renewcommand{\arraystretch}{1.15}
\resizebox{\linewidth}{!}{%
\begin{tabular}{lcccccccccccccc}
\toprule
\textbf{Model} & \makecell[c]{AVMemeExam \\ T$\to$V} & \makecell[c]{ActivityNet \\ T$\to$V} & \makecell[c]{AudioCapsAV \\ T$\to$V} & \makecell[c]{DiDeMo \\ T$\to$V} & \makecell[c]{MSRVTT \\ T$\to$V} & \makecell[c]{MSVD \\ T$\to$V} & \makecell[c]{Panda70M \\ T$\to$V} & \makecell[c]{Shot2Story20K \\ T$\to$V} & \makecell[c]{TUNABench \\ T$\to$V} & \makecell[c]{VALOR32K \\ T$\to$V} & \makecell[c]{VATEX \\ T$\to$V} & \makecell[c]{VGGSoundAV \\ T$\to$V} & \makecell[c]{YouCook2 \\ T$\to$V} & \textbf{Avg.} \\
\midrule
BidirLM/BidirLM-Omni-2.5B-Embedding & 33.70 & 65.02 & 28.39 & 56.73 & 59.22 & 82.43 & -- & 95.44 & 96.39 & 56.24 & 75.86 & 97.72 & 26.13 & 64.44 \\
Haon-Chen/e5-omni-3B & 39.20 & 71.48 & 24.56 & 63.24 & 66.66 & 81.74 & 78.12 & 97.88 & 95.20 & 60.77 & 79.47 & 97.65 & 32.10 & 68.31 \\
Haon-Chen/e5-omni-7B & 42.32 & 74.17 & 21.13 & 65.55 & 68.05 & 82.62 & 83.33 & 99.07 & 98.58 & 64.88 & 81.06 & 98.75 & 37.95 & 70.57 \\
LCO-Embedding/LCO-Embedding-Omni-3B & 44.16 & 63.28 & 26.93 & 61.61 & 62.06 & 79.00 & 70.15 & 95.51 & 95.55 & 55.23 & 76.45 & 94.74 & 34.37 & 66.08 \\
LCO-Embedding/LCO-Embedding-Omni-7B & 48.26 & 70.65 & 27.77 & 65.87 & 64.35 & 79.82 & 70.58 & 96.64 & 97.02 & 59.40 & 80.16 & 96.48 & 41.31 & 69.10 \\
Qwen/Qwen2.5-Omni-3B & 0.45 & 0.08 & 0.65 & 0.41 & 0.49 & 1.00 & 0.18 & 0.13 & 0.64 & 0.12 & 0.40 & 0.60 & 0.18 & 0.41 \\
Qwen/Qwen2.5-Omni-7B & 0.50 & 0.10 & 0.77 & 0.33 & 0.63 & 1.11 & 0.21 & 0.11 & 0.55 & 0.15 & 0.36 & 0.74 & 0.17 & 0.44 \\
Qwen/Qwen3-VL-Embedding-2B & 43.83 & 73.85 & 25.79 & 69.92 & 70.94 & 86.87 & 81.66 & 98.84 & 99.06 & 67.47 & 82.53 & 99.12 & 46.98 & 72.84 \\
Qwen/Qwen3-VL-Embedding-8B & \textbf{50.00} & \textbf{78.82} & 27.60 & \textbf{74.81} & \textbf{74.01} & \textbf{87.56} & 83.38 & \textbf{99.45} & \textbf{99.45} & 71.20 & \textbf{85.30} & \textbf{99.40} & \textbf{53.87} & \textbf{75.76} \\
Tevatron/OmniEmbed-v0.1 & 40.59 & 71.46 & 18.03 & 61.53 & 68.47 & 80.75 & 75.85 & 98.61 & 96.57 & 58.21 & 77.81 & 96.00 & 40.16 & 68.00 \\
encord-team/ebind-audio-vision & 42.28 & 61.18 & 25.84 & 58.31 & 66.60 & 85.46 & \textbf{84.18} & 59.49 & 67.50 & 61.91 & 83.05 & 94.56 & 39.15 & 63.81 \\
encord-team/ebind-full & 42.28 & 61.18 & 25.84 & 58.31 & 66.60 & 85.46 & \textbf{84.18} & 59.49 & 67.50 & 61.91 & 83.05 & 94.56 & 39.15 & 63.81 \\
encord-team/ebind-points-vision & 42.28 & 61.19 & 25.84 & 58.35 & 66.55 & 85.46 & \textbf{84.18} & 59.50 & 67.50 & 61.92 & 83.12 & 94.56 & 39.15 & 63.81 \\
facebook/pe-av-base & 21.95 & 69.41 & 30.86 & 52.14 & 56.90 & 75.63 & 61.18 & 94.78 & 95.99 & 68.99 & 80.54 & 97.42 & 16.47 & 63.25 \\
facebook/pe-av-large & 26.32 & 69.48 & \textbf{35.63} & 55.05 & 59.79 & 78.56 & 64.12 & 95.98 & 96.94 & \textbf{72.00} & 81.69 & 98.33 & 18.96 & 65.60 \\
facebook/pe-av-small & 20.39 & 68.60 & 27.25 & 51.76 & 55.81 & 74.60 & 62.64 & 94.88 & 96.71 & 68.23 & 79.04 & 97.91 & 15.17 & 62.54 \\
jinaai/jina-embeddings-v5-omni-nano & 0.40 & 0.09 & 0.86 & 0.61 & 0.52 & 0.55 & 0.21 & 0.18 & 0.52 & 0.19 & 0.45 & 0.70 & 0.09 & 0.41 \\
jinaai/jina-embeddings-v5-omni-small & 1.03 & 0.05 & 0.53 & 0.47 & 0.33 & 0.55 & 0.15 & 0.05 & 0.52 & 0.23 & 0.53 & 0.73 & 0.11 & 0.41 \\
microsoft/xclip-base-patch16 & 17.72 & 24.02 & 12.56 & 24.48 & 26.91 & 47.93 & 34.30 & 26.80 & 37.18 & 17.17 & 45.21 & 65.07 & 8.01 & 29.80 \\
microsoft/xclip-base-patch32 & 11.69 & 21.22 & 11.02 & 19.60 & 19.33 & 39.93 & 27.96 & 25.81 & 33.21 & 13.42 & 39.51 & 64.14 & 6.63 & 25.65 \\
microsoft/xclip-large-patch14 & 26.58 & 30.47 & 17.15 & 32.49 & 40.39 & 59.19 & 44.41 & 40.52 & 46.57 & 22.36 & 56.88 & 72.60 & 11.23 & 38.53 \\
VLM2Vec/VLM2Vec-V2.0 & 33.40 & 52.66 & 19.94 & 51.46 & 50.23 & 69.76 & 67.92 & 95.02 & 96.17 & 42.19 & 69.19 & 95.37 & 22.37 & 58.90 \\
nvidia/omni-embed-nemotron-3b & 14.63 & 48.61 & 12.41 & 39.71 & 31.99 & 29.09 & 27.30 & 89.64 & 89.40 & 33.28 & 27.27 & 95.09 & 1.15 & 41.51 \\
zhibinlan/UME-R1-2B & 38.37 & 61.63 & 25.10 & 58.63 & 59.83 & 78.01 & 70.42 & 94.51 & 94.42 & 52.48 & 74.05 & 95.94 & 34.60 & 64.46 \\
zhibinlan/UME-R1-7B & 45.20 & 69.53 & 26.30 & 62.34 & 63.10 & 81.69 & 75.45 & 97.88 & 97.47 & 58.01 & 79.18 & 98.58 & 39.16 & 68.76 \\
\bottomrule
\end{tabular}%
}
\caption{\textbf{Retrieval: Text $\to$ Video} (13 tasks, nDCG@10).}
\label{tab:results-retrieval-t2v}
\end{table*}

\begin{table*}[ht]
\centering
\small
\setlength{\tabcolsep}{4pt}
\renewcommand{\arraystretch}{1.15}
\resizebox{\linewidth}{!}{%
\begin{tabular}{lcccccccccc}
\toprule
\textbf{Model} & \makecell[c]{AVMemeExam \\ A$\to$V} & \makecell[c]{AudioCapsAV \\ A$\to$V} & \makecell[c]{DiDeMo \\ A$\to$V} & \makecell[c]{MSRVTT \\ A$\to$V} & \makecell[c]{Shot2Story20K \\ A$\to$V} & \makecell[c]{VALOR32K \\ A$\to$V} & \makecell[c]{VATEX \\ A$\to$V} & \makecell[c]{VGGSoundAV \\ A$\to$V} & \makecell[c]{YouCook2 \\ A$\to$V} & \textbf{Avg.} \\
\midrule
BidirLM/BidirLM-Omni-2.5B-Embedding & 25.31 & 13.96 & 10.23 & 34.86 & 60.39 & 8.73 & 22.88 & 14.17 & 17.34 & 23.10 \\
Haon-Chen/e5-omni-3B & 29.87 & 22.94 & 16.09 & 36.19 & 69.70 & 12.45 & 35.55 & 24.91 & 16.79 & 29.39 \\
Haon-Chen/e5-omni-7B & 31.95 & 19.53 & 20.16 & 45.52 & 86.12 & 12.71 & 43.02 & 27.16 & 31.65 & 35.31 \\
LCO-Embedding/LCO-Embedding-Omni-3B & 40.60 & 31.30 & 27.86 & 45.89 & 82.83 & 20.87 & 46.71 & 34.82 & 31.81 & 40.30 \\
LCO-Embedding/LCO-Embedding-Omni-7B & \textbf{43.60} & 35.86 & 29.36 & 49.47 & \textbf{89.08} & 23.21 & 49.05 & 40.25 & \textbf{41.94} & 44.65 \\
Qwen/Qwen2.5-Omni-3B & 1.10 & 0.91 & 0.61 & 0.57 & 0.14 & 0.13 & 0.70 & 0.92 & 0.15 & 0.58 \\
Qwen/Qwen2.5-Omni-7B & 0.69 & 0.71 & 0.41 & 0.30 & 0.26 & 0.14 & 0.53 & 0.43 & 0.12 & 0.40 \\
Tevatron/OmniEmbed-v0.1 & 21.09 & 13.68 & 13.53 & 25.26 & 68.54 & 6.40 & 35.13 & 15.69 & 14.97 & 23.81 \\
encord-team/ebind-audio-vision & 36.51 & 78.13 & 39.74 & 41.19 & 17.48 & 47.71 & 45.32 & 67.98 & 7.89 & 42.44 \\
encord-team/ebind-full & 36.51 & 78.13 & 39.74 & 41.19 & 17.48 & 47.71 & 45.32 & 67.98 & 7.89 & 42.44 \\
facebook/pe-av-base & 23.86 & 69.35 & 59.99 & 43.36 & 21.85 & 48.24 & 62.20 & 74.58 & 10.55 & 46.00 \\
facebook/pe-av-large & 24.52 & \textbf{79.01} & \textbf{65.81} & \textbf{49.92} & 27.80 & \textbf{58.44} & \textbf{68.77} & \textbf{84.27} & 16.28 & \textbf{52.76} \\
facebook/pe-av-small & 16.84 & 65.60 & 55.74 & 39.83 & 20.52 & 42.09 & 61.33 & 69.13 & 9.29 & 42.26 \\
jinaai/jina-embeddings-v5-omni-nano & 0.50 & 1.21 & 0.53 & 0.63 & 0.05 & 0.17 & 0.47 & 0.64 & 0.21 & 0.49 \\
jinaai/jina-embeddings-v5-omni-small & 0.60 & 1.02 & 0.53 & 0.25 & 0.11 & 0.11 & 0.40 & 0.53 & 0.04 & 0.40 \\
nvidia/omni-embed-nemotron-3b & 29.77 & 24.88 & 20.38 & 37.10 & 66.64 & 16.09 & 35.24 & 30.03 & 20.39 & 31.17 \\
\bottomrule
\end{tabular}%
}
\caption{\textbf{Retrieval: Audio $\to$ Video} (9 tasks, nDCG@10).}
\label{tab:results-retrieval-a2v}
\end{table*}

\begin{table*}[ht]
\centering
\small
\setlength{\tabcolsep}{4pt}
\renewcommand{\arraystretch}{1.15}
\resizebox{\linewidth}{!}{%
\begin{tabular}{lcccccccccc}
\toprule
\textbf{Model} & \makecell[c]{AVMemeExam \\ AT$\to$V} & \makecell[c]{AudioCapsAV \\ AT$\to$V} & \makecell[c]{DiDeMo \\ AT$\to$V} & \makecell[c]{MSRVTT \\ AT$\to$V} & \makecell[c]{Shot2Story20K \\ AT$\to$V} & \makecell[c]{VALOR32K \\ AT$\to$V} & \makecell[c]{VATEX \\ AT$\to$V} & \makecell[c]{VGGSoundAV \\ AT$\to$V} & \makecell[c]{YouCook2 \\ AT$\to$V} & \textbf{Avg.} \\
\midrule
BidirLM/BidirLM-Omni-2.5B-Embedding & 39.93 & 30.50 & 45.65 & 58.57 & 90.50 & 48.28 & 72.95 & 95.04 & 29.38 & 56.76 \\
Haon-Chen/e5-omni-3B & 47.70 & 29.51 & 55.31 & 65.26 & 98.44 & 50.20 & 75.73 & 95.68 & 32.33 & 61.13 \\
Haon-Chen/e5-omni-7B & 42.52 & 24.46 & 59.54 & 73.12 & \textbf{99.39} & 48.97 & 77.71 & 95.98 & 43.58 & 62.81 \\
LCO-Embedding/LCO-Embedding-Omni-3B & 53.32 & 35.24 & 68.96 & 77.39 & 97.81 & 61.89 & 83.36 & 94.04 & 41.46 & 68.16 \\
LCO-Embedding/LCO-Embedding-Omni-7B & 55.95 & 38.31 & 72.52 & \textbf{79.19} & 98.68 & 65.97 & 88.22 & 94.45 & 52.93 & 71.80 \\
Qwen/Qwen2.5-Omni-3B & 0.55 & 0.82 & 0.42 & 0.50 & 0.15 & 0.12 & 0.60 & 0.67 & 0.16 & 0.45 \\
Qwen/Qwen2.5-Omni-7B & 0.77 & 0.65 & 0.59 & 0.31 & 0.18 & 0.15 & 0.42 & 1.04 & 0.23 & 0.48 \\
Tevatron/OmniEmbed-v0.1 & 31.23 & 16.08 & 34.36 & 47.97 & 98.79 & 21.70 & 58.48 & 76.04 & 22.10 & 45.19 \\
encord-team/ebind-audio-vision & \textbf{58.98} & \textbf{69.89} & 74.88 & 76.81 & 68.44 & 79.00 & 92.20 & 96.56 & \textbf{56.83} & \textbf{74.84} \\
encord-team/ebind-full & \textbf{58.98} & \textbf{69.89} & 74.88 & 76.78 & 68.44 & 79.00 & 92.20 & 96.56 & \textbf{56.83} & 74.84 \\
facebook/pe-av-base & 28.77 & 57.58 & 72.58 & 69.01 & 94.22 & 79.62 & 90.36 & 98.29 & 26.18 & 68.51 \\
facebook/pe-av-large & 31.50 & 68.54 & \textbf{76.07} & 73.43 & 95.71 & \textbf{84.69} & \textbf{93.08} & \textbf{98.74} & 29.68 & 72.38 \\
facebook/pe-av-small & 25.85 & 50.60 & 71.46 & 67.08 & 94.32 & 77.39 & 89.51 & 98.10 & 23.02 & 66.37 \\
jinaai/jina-embeddings-v5-omni-nano & 0.48 & 1.21 & 0.54 & 0.83 & 0.09 & 0.18 & 0.47 & 0.59 & 0.18 & 0.51 \\
jinaai/jina-embeddings-v5-omni-small & 0.75 & 1.06 & 0.56 & 0.25 & 0.09 & 0.14 & 0.47 & 0.59 & 0.03 & 0.44 \\
nvidia/omni-embed-nemotron-3b & 33.34 & 24.53 & 21.35 & 38.31 & 85.84 & 17.56 & 36.00 & 63.41 & 20.86 & 37.91 \\
\bottomrule
\end{tabular}%
}
\caption{\textbf{Retrieval: Audio+Text $\to$ Video} (9 tasks, nDCG@10).}
\label{tab:results-retrieval-at2v}
\end{table*}

\begin{table*}[ht]
\centering
\small
\setlength{\tabcolsep}{4pt}
\renewcommand{\arraystretch}{1.15}
\resizebox{\linewidth}{!}{%
\begin{tabular}{lcccccccccccccc}
\toprule
\textbf{Model} & \makecell[c]{AVMemeExam \\ V$\to$T} & \makecell[c]{ActivityNet \\ V$\to$T} & \makecell[c]{AudioCapsAV \\ V$\to$T} & \makecell[c]{DiDeMo \\ V$\to$T} & \makecell[c]{MSRVTT \\ V$\to$T} & \makecell[c]{MSVD \\ V$\to$T} & \makecell[c]{Panda70M \\ V$\to$T} & \makecell[c]{Shot2Story20K \\ V$\to$T} & \makecell[c]{TUNABench \\ V$\to$T} & \makecell[c]{VALOR32K \\ V$\to$T} & \makecell[c]{VATEX \\ V$\to$T} & \makecell[c]{VGGSoundAV \\ V$\to$T} & \makecell[c]{YouCook2 \\ V$\to$T} & \textbf{Avg.} \\
\midrule
BidirLM/BidirLM-Omni-2.5B-Embedding & 28.36 & 60.26 & 30.32 & 54.14 & 56.35 & 81.05 & -- & 93.15 & 94.42 & 57.63 & 74.72 & 97.14 & 20.20 & 62.31 \\
Haon-Chen/e5-omni-3B & 23.94 & 48.05 & 20.28 & 36.12 & 49.28 & 73.29 & 68.87 & 79.29 & 80.78 & 48.56 & 66.85 & 88.88 & 18.86 & 54.08 \\
Haon-Chen/e5-omni-7B & 37.60 & 67.86 & 27.88 & 64.77 & 63.06 & 77.03 & 79.10 & 98.55 & 98.27 & 59.71 & 79.37 & \textbf{98.87} & 31.11 & 67.94 \\
LCO-Embedding/LCO-Embedding-Omni-3B & 37.77 & 58.95 & 29.85 & 57.14 & 58.07 & 79.46 & 69.62 & 93.22 & 95.12 & 53.12 & 76.71 & 94.46 & 35.51 & 64.54 \\
LCO-Embedding/LCO-Embedding-Omni-7B & 40.43 & 65.67 & 31.10 & 60.62 & 61.15 & 81.32 & 71.75 & 95.52 & 96.33 & 58.08 & 80.07 & 96.86 & 39.20 & 67.55 \\
Qwen/Qwen2.5-Omni-3B & 0.53 & 0.12 & 0.74 & 0.44 & 0.53 & 0.54 & 0.16 & 0.14 & 0.37 & 0.16 & 0.31 & 0.79 & 0.12 & 0.38 \\
Qwen/Qwen2.5-Omni-7B & 0.57 & 0.11 & 0.62 & 0.46 & 0.50 & 0.60 & 0.11 & 0.16 & 0.54 & 0.13 & 0.42 & 0.64 & 0.13 & 0.38 \\
Qwen/Qwen3-VL-Embedding-2B & 40.46 & 57.32 & 29.16 & 68.72 & 58.01 & 81.95 & 78.74 & 96.89 & 98.36 & 63.76 & 80.51 & 98.11 & 43.94 & 68.92 \\
Qwen/Qwen3-VL-Embedding-8B & \textbf{45.29} & \textbf{69.84} & 29.29 & \textbf{72.87} & 62.80 & 84.16 & 80.70 & \textbf{98.75} & \textbf{98.70} & \textbf{66.42} & \textbf{83.43} & 98.80 & \textbf{52.45} & \textbf{72.58} \\
Tevatron/OmniEmbed-v0.1 & 18.32 & 56.52 & 26.82 & 48.50 & 54.06 & 70.23 & 58.70 & 96.75 & 93.91 & 43.32 & 64.05 & 94.62 & 18.44 & 57.25 \\
encord-team/ebind-audio-vision & 44.85 & 59.33 & 32.03 & 58.92 & 67.41 & 86.07 & \textbf{85.63} & 60.06 & 67.09 & 65.10 & 82.92 & 93.61 & 41.60 & 64.97 \\
encord-team/ebind-full & 44.85 & 59.33 & 32.03 & 58.92 & 67.41 & 86.07 & \textbf{85.63} & 60.06 & 67.09 & 65.12 & 82.92 & 93.61 & 41.60 & 64.97 \\
encord-team/ebind-points-vision & 44.85 & 59.31 & 32.03 & 58.92 & \textbf{67.47} & \textbf{86.07} & \textbf{85.63} & 60.06 & 67.09 & 65.10 & 82.92 & 93.61 & 41.61 & 64.97 \\
facebook/pe-av-base & 21.83 & 54.84 & 31.19 & 50.32 & 56.36 & 72.38 & 57.19 & 93.35 & 93.97 & 56.89 & 78.27 & 94.84 & 18.70 & 60.01 \\
facebook/pe-av-large & 26.22 & 56.31 & \textbf{34.96} & 54.17 & 60.03 & 79.34 & 65.65 & 95.47 & 96.23 & 64.99 & 80.08 & 96.25 & 22.54 & 64.02 \\
facebook/pe-av-small & 22.91 & 53.86 & 29.77 & 49.14 & 56.44 & 73.33 & 62.02 & 92.89 & 94.91 & 57.25 & 77.91 & 95.12 & 17.80 & 60.26 \\
jinaai/jina-embeddings-v5-omni-nano & 0.50 & 0.10 & 0.57 & 0.61 & 0.50 & 0.72 & 0.15 & 0.14 & 0.44 & 0.14 & 0.48 & 0.53 & 0.15 & 0.39 \\
jinaai/jina-embeddings-v5-omni-small & 0.55 & 0.07 & 0.74 & 0.40 & 0.50 & 0.56 & 0.12 & 0.12 & 0.40 & 0.17 & 0.62 & 0.65 & 0.17 & 0.39 \\
microsoft/xclip-base-patch16 & 23.48 & 28.85 & 18.02 & 33.69 & 39.72 & 60.27 & 42.60 & 39.35 & 48.92 & 25.86 & 56.92 & 70.64 & 6.52 & 38.06 \\
microsoft/xclip-base-patch32 & 22.23 & 28.98 & 19.27 & 34.89 & 38.83 & 55.59 & 41.12 & 38.96 & 45.76 & 25.09 & 55.56 & 70.58 & 6.94 & 37.21 \\
microsoft/xclip-large-patch14 & 27.74 & 29.22 & 18.27 & 34.57 & 42.24 & 63.53 & 46.62 & 46.06 & 52.66 & 24.59 & 59.22 & 74.57 & 9.89 & 40.71 \\
VLM2Vec/VLM2Vec-V2.0 & 31.92 & 57.77 & 21.56 & 44.49 & 48.52 & 69.86 & 72.58 & 95.62 & 96.36 & 50.44 & 65.55 & 96.59 & 24.91 & 59.71 \\
nvidia/omni-embed-nemotron-3b & 25.72 & 39.81 & 18.47 & 36.44 & 28.29 & 34.45 & 49.87 & 90.13 & 93.13 & 31.85 & 46.80 & 94.20 & 10.91 & 46.16 \\
zhibinlan/UME-R1-2B & 34.51 & 59.88 & 29.95 & 59.35 & 57.25 & 79.51 & 71.39 & 92.62 & 93.15 & 56.19 & 74.47 & 96.20 & 37.04 & 64.73 \\
zhibinlan/UME-R1-7B & 44.68 & 69.13 & 31.67 & 63.81 & 62.39 & 82.34 & 77.51 & 97.17 & 96.89 & 62.39 & 80.23 & 98.57 & 42.66 & 69.96 \\
\bottomrule
\end{tabular}%
}
\caption{\textbf{Retrieval: Video $\to$ Text} (13 tasks, nDCG@10).}
\label{tab:results-retrieval-v2t}
\end{table*}

\begin{table*}[ht]
\centering
\small
\setlength{\tabcolsep}{4pt}
\renewcommand{\arraystretch}{1.15}
\resizebox{\linewidth}{!}{%
\begin{tabular}{lccccccccccc}
\toprule
\textbf{Model} & \makecell[c]{AVMemeExam \\ VA$\to$T} & \makecell[c]{AudioCapsAV \\ VA$\to$T} & \makecell[c]{DiDeMo \\ VA$\to$T} & \makecell[c]{MSRVTT \\ VA$\to$T} & \makecell[c]{Panda70M \\ VA$\to$T} & \makecell[c]{Shot2Story20K \\ VA$\to$T} & \makecell[c]{VALOR32K \\ VA$\to$T} & \makecell[c]{VATEX \\ VA$\to$T} & \makecell[c]{VGGSoundAV \\ VA$\to$T} & \makecell[c]{YouCook2 \\ VA$\to$T} & \textbf{Avg.} \\
\midrule
BidirLM/BidirLM-Omni-2.5B-Embedding & 47.49 & 35.38 & 53.01 & 61.97 & -- & 96.82 & 59.75 & 76.61 & 96.92 & 31.20 & 62.13 \\
Haon-Chen/e5-omni-3B & 30.29 & 39.80 & 19.26 & 32.49 & 46.20 & 78.98 & 30.97 & 50.96 & 55.79 & 24.89 & 40.96 \\
Haon-Chen/e5-omni-7B & 51.23 & 47.48 & \textbf{58.85} & \textbf{63.32} & 77.32 & \textbf{98.95} & 60.87 & \textbf{79.15} & \textbf{98.73} & 50.74 & \textbf{68.66} \\
LCO-Embedding/LCO-Embedding-Omni-3B & \textbf{59.26} & 46.52 & 48.17 & 51.84 & 48.58 & 93.78 & 53.69 & 72.21 & 92.14 & 59.21 & 62.54 \\
LCO-Embedding/LCO-Embedding-Omni-7B & 58.13 & 49.29 & 47.32 & 54.74 & 50.77 & 96.01 & 55.81 & 75.49 & 94.66 & \textbf{63.00} & 64.52 \\
Qwen/Qwen2.5-Omni-3B & 0.78 & 0.59 & 0.49 & 0.61 & 0.15 & 0.15 & 0.16 & 0.41 & 0.48 & 0.12 & 0.39 \\
Qwen/Qwen2.5-Omni-7B & 0.63 & 0.60 & 0.44 & 0.38 & 0.12 & 0.18 & 0.11 & 0.46 & 0.77 & 0.15 & 0.39 \\
Tevatron/OmniEmbed-v0.1 & 21.86 & 52.79 & 37.58 & 48.13 & 42.66 & 98.60 & 29.40 & 59.52 & 89.99 & 22.92 & 50.35 \\
encord-team/ebind-audio-vision & 42.51 & 45.59 & 51.59 & 62.30 & \textbf{79.45} & 52.46 & 64.62 & 78.48 & 90.96 & 35.35 & 60.33 \\
encord-team/ebind-full & 42.51 & 45.59 & 51.59 & 62.30 & \textbf{79.45} & 52.46 & 64.62 & 78.48 & 90.96 & 35.35 & 60.33 \\
facebook/pe-av-base & 22.00 & 55.23 & 44.52 & 55.16 & 50.35 & 90.79 & \textbf{66.51} & 76.41 & 95.99 & 16.43 & 57.34 \\
facebook/pe-av-large & 22.27 & \textbf{59.58} & 43.12 & 53.13 & 44.47 & 87.53 & 65.87 & 74.89 & 93.90 & 17.25 & 56.20 \\
facebook/pe-av-small & 23.24 & 55.30 & 42.43 & 56.61 & 54.48 & 91.00 & 63.82 & 76.94 & 95.44 & 16.34 & 57.56 \\
jinaai/jina-embeddings-v5-omni-nano & 22.99 & 32.19 & 5.71 & 16.00 & 5.64 & 46.51 & 8.00 & 9.20 & 29.00 & 9.86 & 18.51 \\
jinaai/jina-embeddings-v5-omni-small & 21.61 & 28.30 & 2.92 & 13.87 & 5.78 & 45.60 & 5.57 & 9.46 & 25.32 & 4.90 & 16.33 \\
nvidia/omni-embed-nemotron-3b & 39.41 & 23.41 & 30.52 & 33.32 & 47.90 & 94.54 & 34.88 & 46.98 & 94.30 & 21.71 & 46.70 \\
\bottomrule
\end{tabular}%
}
\caption{\textbf{Retrieval: Video+Audio $\to$ Text} (10 tasks, nDCG@10).}
\label{tab:results-retrieval-va2t}
\end{table*}

\begin{table*}[ht]
\centering
\small
\setlength{\tabcolsep}{4pt}
\renewcommand{\arraystretch}{1.15}
\resizebox{\linewidth}{!}{%
\begin{tabular}{lcccccccccc}
\toprule
\textbf{Model} & \makecell[c]{AVMemeExam \\ V$\to$A} & \makecell[c]{AudioCapsAV \\ V$\to$A} & \makecell[c]{DiDeMo \\ V$\to$A} & \makecell[c]{MSRVTT \\ V$\to$A} & \makecell[c]{Shot2Story20K \\ V$\to$A} & \makecell[c]{VALOR32K \\ V$\to$A} & \makecell[c]{VATEX \\ V$\to$A} & \makecell[c]{VGGSoundAV \\ V$\to$A} & \makecell[c]{YouCook2 \\ V$\to$A} & \textbf{Avg.} \\
\midrule
BidirLM/BidirLM-Omni-2.5B-Embedding & 23.81 & 17.10 & 12.72 & 30.76 & 53.04 & 10.31 & 23.11 & 15.57 & 15.25 & 22.41 \\
Haon-Chen/e5-omni-3B & 28.58 & 26.08 & 22.06 & 28.64 & 56.15 & 14.64 & 34.72 & 29.69 & 13.40 & 28.22 \\
Haon-Chen/e5-omni-7B & 39.77 & 35.63 & 30.79 & \textbf{50.78} & \textbf{83.98} & 24.31 & 46.54 & 38.95 & 30.56 & 42.37 \\
LCO-Embedding/LCO-Embedding-Omni-3B & 40.58 & 35.05 & 28.08 & 41.72 & 75.14 & 23.29 & 43.80 & 38.02 & 29.31 & 39.44 \\
LCO-Embedding/LCO-Embedding-Omni-7B & \textbf{45.32} & 39.20 & 30.17 & 48.41 & 83.13 & 27.33 & 47.98 & 43.42 & \textbf{35.59} & \textbf{44.51} \\
Qwen/Qwen2.5-Omni-3B & 0.51 & 0.74 & 0.63 & 0.63 & 0.15 & 0.17 & 0.50 & 0.63 & 0.21 & 0.46 \\
Qwen/Qwen2.5-Omni-7B & 0.65 & 0.81 & 0.46 & 0.76 & 0.11 & 0.12 & 0.53 & 0.66 & 0.21 & 0.48 \\
Tevatron/OmniEmbed-v0.1 & 29.58 & 32.90 & 26.98 & 40.81 & 57.37 & 20.27 & 38.47 & 35.12 & 12.06 & 32.62 \\
encord-team/ebind-audio-vision & 34.05 & \textbf{81.40} & 40.92 & 43.34 & 20.41 & \textbf{50.98} & 46.67 & 70.13 & 7.80 & 43.97 \\
encord-team/ebind-full & 34.05 & \textbf{81.40} & 40.92 & 43.34 & 20.41 & \textbf{50.98} & 46.67 & 70.13 & 7.80 & 43.97 \\
facebook/pe-av-base & 10.94 & 65.58 & 57.69 & 32.61 & 19.56 & 38.02 & 43.13 & 65.90 & 4.91 & 37.59 \\
facebook/pe-av-large & 12.17 & 71.60 & \textbf{61.90} & 37.98 & 24.64 & 41.94 & \textbf{49.97} & \textbf{77.07} & 7.02 & 42.70 \\
facebook/pe-av-small & 11.73 & 59.61 & 51.32 & 34.14 & 19.33 & 31.95 & 35.70 & 61.23 & 4.59 & 34.40 \\
jinaai/jina-embeddings-v5-omni-nano & 0.57 & 0.78 & 0.33 & 0.50 & 0.14 & 0.16 & 0.44 & 0.90 & 0.12 & 0.44 \\
jinaai/jina-embeddings-v5-omni-small & 0.51 & 0.68 & 0.53 & 0.40 & 0.15 & 0.17 & 0.56 & 0.74 & 0.13 & 0.43 \\
nvidia/omni-embed-nemotron-3b & 36.67 & 26.24 & 20.46 & 31.29 & 52.15 & 16.10 & 33.93 & 32.21 & 17.50 & 29.62 \\
\bottomrule
\end{tabular}%
}
\caption{\textbf{Retrieval: Video $\to$ Audio} (9 tasks, nDCG@10).}
\label{tab:results-retrieval-v2a}
\end{table*}

\begin{table*}[ht]
\centering
\small
\setlength{\tabcolsep}{4pt}
\renewcommand{\arraystretch}{1.15}
\resizebox{\linewidth}{!}{%
\begin{tabular}{lcccccccccc}
\toprule
\textbf{Model} & \makecell[c]{AVMemeExam \\ VT$\to$A} & \makecell[c]{AudioCapsAV \\ VT$\to$A} & \makecell[c]{DiDeMo \\ VT$\to$A} & \makecell[c]{MSRVTT \\ VT$\to$A} & \makecell[c]{Shot2Story20K \\ VT$\to$A} & \makecell[c]{VALOR32K \\ VT$\to$A} & \makecell[c]{VATEX \\ VT$\to$A} & \makecell[c]{VGGSoundAV \\ VT$\to$A} & \makecell[c]{YouCook2 \\ VT$\to$A} & \textbf{Avg.} \\
\midrule
BidirLM/BidirLM-Omni-2.5B-Embedding & 44.79 & 21.77 & 12.64 & 36.19 & 78.07 & 11.70 & 25.44 & 21.59 & 25.31 & 30.83 \\
Haon-Chen/e5-omni-3B & 46.78 & 63.82 & 20.81 & 36.10 & 80.20 & 25.13 & 40.76 & 55.24 & 25.87 & 43.86 \\
Haon-Chen/e5-omni-7B & 59.34 & 66.39 & 30.33 & \textbf{52.99} & \textbf{93.33} & 33.40 & \textbf{49.83} & 63.13 & 44.64 & \textbf{54.82} \\
LCO-Embedding/LCO-Embedding-Omni-3B & 62.08 & 48.87 & 21.33 & 41.72 & 90.58 & 24.68 & 42.93 & 56.41 & 42.61 & 47.91 \\
LCO-Embedding/LCO-Embedding-Omni-7B & \textbf{67.05} & 55.86 & 25.89 & 48.54 & 93.11 & 30.79 & 48.88 & 61.81 & \textbf{51.81} & 53.75 \\
Qwen/Qwen2.5-Omni-3B & 1.11 & 1.13 & 0.77 & 0.59 & 0.21 & 0.13 & 1.19 & 0.34 & 0.51 & 0.66 \\
Qwen/Qwen2.5-Omni-7B & 1.40 & 0.71 & 0.58 & 0.50 & 0.18 & 0.17 & 0.76 & 1.04 & 0.33 & 0.63 \\
Tevatron/OmniEmbed-v0.1 & 44.61 & 61.49 & 26.53 & 42.95 & 77.89 & 27.32 & 42.33 & 62.65 & 16.36 & 44.68 \\
encord-team/ebind-audio-vision & 34.87 & \textbf{86.51} & 35.78 & 41.65 & 17.78 & 50.34 & 40.15 & 63.30 & 5.29 & 41.74 \\
encord-team/ebind-full & 34.87 & \textbf{86.51} & 35.78 & 41.66 & 17.78 & 50.34 & 40.15 & 63.30 & 5.29 & 41.74 \\
facebook/pe-av-base & 16.53 & 80.49 & 39.34 & 25.89 & 22.85 & 48.12 & 36.93 & 81.94 & 2.08 & 39.35 \\
facebook/pe-av-large & 18.54 & 83.12 & \textbf{49.84} & 27.96 & 26.35 & \textbf{52.42} & 41.92 & \textbf{87.55} & 2.75 & 43.38 \\
facebook/pe-av-small & 16.87 & 78.72 & 37.94 & 24.49 & 21.67 & 44.04 & 29.11 & 81.58 & 1.76 & 37.35 \\
jinaai/jina-embeddings-v5-omni-nano & 26.04 & 41.42 & 5.66 & 16.85 & 48.32 & 9.57 & 11.90 & 34.62 & 8.93 & 22.59 \\
jinaai/jina-embeddings-v5-omni-small & 3.75 & 7.79 & 1.17 & 3.39 & 7.34 & 1.59 & 3.72 & 6.73 & 0.59 & 4.01 \\
nvidia/omni-embed-nemotron-3b & 53.19 & 38.01 & 20.57 & 35.97 & 81.36 & 20.12 & 37.68 & 45.35 & 35.98 & 40.91 \\
\bottomrule
\end{tabular}%
}
\caption{\textbf{Retrieval: Video+Text $\to$ Audio} (9 tasks, nDCG@10).}
\label{tab:results-retrieval-vt2a}
\end{table*}

\begin{table*}[ht]
\centering
\small
\setlength{\tabcolsep}{4pt}
\renewcommand{\arraystretch}{1.15}
\resizebox{\linewidth}{!}{%
\begin{tabular}{lccccccccccc}
\toprule
\textbf{Model} & \makecell[c]{AVMemeExam \\ T$\to$VA} & \makecell[c]{AudioCapsAV \\ T$\to$VA} & \makecell[c]{DiDeMo \\ T$\to$VA} & \makecell[c]{MSRVTT \\ T$\to$VA} & \makecell[c]{Panda70M \\ T$\to$VA} & \makecell[c]{Shot2Story20K \\ T$\to$VA} & \makecell[c]{VALOR32K \\ T$\to$VA} & \makecell[c]{VATEX \\ T$\to$VA} & \makecell[c]{VGGSoundAV \\ T$\to$VA} & \makecell[c]{YouCook2 \\ T$\to$VA} & \textbf{Avg.} \\
\midrule
BidirLM/BidirLM-Omni-2.5B-Embedding & 53.22 & 30.86 & 57.43 & 64.16 & -- & 98.16 & 59.48 & 77.93 & 96.87 & 36.89 & 63.89 \\
Haon-Chen/e5-omni-3B & 56.74 & 59.25 & 54.28 & 62.53 & 72.40 & 98.69 & 64.32 & 77.20 & 95.93 & 42.52 & 68.39 \\
Haon-Chen/e5-omni-7B & 64.51 & 52.92 & \textbf{63.30} & \textbf{71.76} & \textbf{79.49} & \textbf{99.47} & 72.08 & \textbf{83.61} & \textbf{99.06} & \textbf{56.31} & \textbf{74.25} \\
LCO-Embedding/LCO-Embedding-Omni-3B & 64.13 & 39.85 & 50.16 & 58.32 & 46.59 & 94.70 & 55.34 & 69.47 & 92.66 & 43.65 & 61.49 \\
LCO-Embedding/LCO-Embedding-Omni-7B & \textbf{66.97} & 43.67 & 51.19 & 61.55 & 48.62 & 95.38 & 59.09 & 73.85 & 94.21 & 52.71 & 64.72 \\
Qwen/Qwen2.5-Omni-3B & 1.21 & 0.54 & 0.46 & 0.66 & 0.12 & 0.18 & 0.19 & 0.47 & 0.62 & 0.15 & 0.46 \\
Qwen/Qwen2.5-Omni-7B & 0.54 & 0.54 & 0.46 & 0.33 & 0.16 & 0.15 & 0.17 & 0.31 & 0.78 & 0.14 & 0.36 \\
Tevatron/OmniEmbed-v0.1 & 60.41 & 57.33 & 56.05 & 66.82 & 72.41 & 99.39 & 65.96 & 75.20 & 97.23 & 51.41 & 70.22 \\
encord-team/ebind-audio-vision & 44.34 & 44.12 & 50.97 & 62.76 & 76.03 & 53.00 & 63.25 & 76.08 & 90.76 & 28.94 & 59.03 \\
encord-team/ebind-full & 44.34 & 44.12 & 50.97 & 62.66 & 76.03 & 53.00 & 63.25 & 76.08 & 90.76 & 28.94 & 59.02 \\
facebook/pe-av-base & 25.52 & 60.92 & 45.50 & 58.41 & 58.01 & 93.36 & \textbf{76.72} & 78.35 & 97.17 & 15.40 & 60.94 \\
facebook/pe-av-large & 22.50 & \textbf{61.62} & 42.25 & 54.16 & 51.13 & 92.02 & 75.55 & 74.80 & 96.12 & 13.04 & 58.32 \\
facebook/pe-av-small & 26.04 & 59.89 & 45.55 & 55.35 & 62.04 & 93.52 & 76.39 & 78.08 & 96.68 & 15.18 & 60.87 \\
jinaai/jina-embeddings-v5-omni-nano & 23.26 & 32.76 & 5.36 & 15.76 & 6.34 & 42.46 & 8.87 & 10.34 & 25.67 & 8.70 & 17.95 \\
jinaai/jina-embeddings-v5-omni-small & 22.89 & 31.31 & 5.29 & 17.10 & 7.81 & 51.28 & 8.14 & 12.01 & 28.90 & 6.12 & 19.08 \\
nvidia/omni-embed-nemotron-3b & 29.05 & 23.25 & 31.13 & 31.81 & 24.38 & 94.80 & 39.95 & 17.04 & 94.47 & 5.47 & 39.13 \\
\bottomrule
\end{tabular}%
}
\caption{\textbf{Retrieval: Text $\to$ Video+Audio} (10 tasks, nDCG@10).}
\label{tab:results-retrieval-t2va}
\end{table*}

\subsection{Classification}
\label{appdx:results-classification}

Per-task scores are in \autoref{tab:results-classification-a} and \autoref{tab:results-classification-b}. LCO-Embedding-Omni-7B leads on the classification family average (57.4), with BidirLM-Omni-2.5B (56.9) and Qwen3-VL-Embedding-8B (56.8) close behind; BidirLM wins many individual audio$+$video columns despite ranking ninth on the headline MVEB leaderboard, and Qwen3-VL-Embedding-8B is the strongest text-video classifier (top on HumAnimCartoon-video, Breakfast, HMDB51). X-CLIP-large-patch14 remains the strongest video-only-input model on the Kinetics family and UCF101 (e.g.\ 83.3 on Kinetics-400, 76.6 on Kinetics-600, 96.1 on UCF101), reflecting Kinetics-supervised pretraining transferring cleanly to evaluation on the same dataset family (which we audit as in-distribution in Appendix~\ref{appdx:contamination}). For audio$+$video variants the MLLM-embed family is uniformly stronger because the X-CLIP, V-JEPA-2, and Qwen3-VL models cannot consume the audio channel; the audio$+$video columns are blank for those families. Audio rarely degrades performance for omni models: on Kinetics-700, MELD, and RAVDESS the audio$+$video score is within a few points of the video-only score for most models.

\begin{table*}[ht]
\centering
\small
\setlength{\tabcolsep}{4pt}
\renewcommand{\arraystretch}{1.15}
\resizebox{\linewidth}{!}{%
\begin{tabular}{lccccccccccccccc}
\toprule
\textbf{Model} & \makecell[c]{AVEDataset \\ audio + video} & \makecell[c]{AVEDataset \\ video} & \makecell[c]{AVMeme \\ audio + video} & \makecell[c]{AVMeme \\ video} & \makecell[c]{Breakfast \\ video} & \makecell[c]{Diving48 \\ video} & \makecell[c]{HMDB51 \\ video} & \makecell[c]{HumAnimCartoon \\ video} & \makecell[c]{HumAnimCartoon \\ audio + video} & \makecell[c]{Kinetics400 \\ video} & \makecell[c]{Kinetics400 \\ audio + video} & \makecell[c]{Kinetics600 \\ video} & \makecell[c]{Kinetics600 \\ audio + video} & \makecell[c]{Kinetics700 \\ video} & \textbf{Avg.} \\
\midrule
BidirLM/BidirLM-Omni-2.5B-Embedding & \textbf{61.09} & \textbf{60.05} & 54.33 & 46.33 & 43.14 & 5.91 & 63.22 & 93.33 & \textbf{92.24} & 66.57 & \textbf{67.56} & 68.23 & \textbf{69.60} & 57.02 & 56.89 \\
Haon-Chen/e5-omni-3B & 56.84 & 56.29 & 62.22 & \textbf{52.56} & 42.71 & 3.35 & 56.41 & 86.81 & 79.66 & 56.12 & 46.54 & 56.37 & 48.62 & 45.30 & 48.60 \\
Haon-Chen/e5-omni-7B & 58.66 & 57.11 & 59.67 & 48.67 & 40.63 & 3.72 & 59.03 & 87.43 & 86.19 & 59.90 & 58.43 & 60.35 & 59.66 & 49.33 & 52.23 \\
LCO-Embedding/LCO-Embedding-Omni-3B & 58.43 & 58.76 & 59.00 & 49.56 & 58.65 & 3.97 & 64.73 & 92.40 & 90.53 & 68.73 & 63.25 & 69.14 & 63.83 & 59.30 & 55.90 \\
LCO-Embedding/LCO-Embedding-Omni-7B & 60.00 & 59.38 & \textbf{62.44} & 49.22 & 59.33 & 4.56 & 66.26 & 93.33 & 89.91 & 70.01 & 65.03 & 70.75 & 67.26 & 60.36 & \textbf{57.43} \\
Qwen/Qwen2.5-Omni-3B & 22.39 & 14.05 & 43.89 & 28.78 & 12.69 & 2.43 & 13.02 & 26.25 & 19.10 & 0.25 & 0.25 & 0.93 & 0.65 & 0.44 & 14.19 \\
Qwen/Qwen2.5-Omni-7B & 23.61 & 9.85 & 37.78 & 31.56 & 15.01 & 2.15 & 10.49 & 22.22 & 24.54 & 0.25 & 0.25 & 0.38 & 0.81 & 0.20 & 14.17 \\
Qwen/Qwen3-VL-Embedding-2B & -- & 58.83 & -- & 49.22 & 62.78 & 4.61 & 67.09 & 93.32 & -- & 69.38 & -- & 70.04 & -- & 60.14 & 54.98 \\
Qwen/Qwen3-VL-Embedding-8B & -- & 58.71 & -- & 49.89 & \textbf{70.90} & 4.02 & \textbf{70.75} & \textbf{94.41} & -- & 71.40 & -- & 71.63 & -- & 60.87 & 56.81 \\
Tevatron/OmniEmbed-v0.1 & 56.84 & 56.22 & 61.89 & 51.33 & 50.34 & 4.06 & 62.85 & 88.36 & 82.61 & 61.74 & 56.15 & 62.73 & 58.46 & 51.41 & 53.33 \\
encord-team/ebind-audio-vision & 59.20 & 58.53 & 60.67 & 44.78 & 38.09 & 5.07 & 61.92 & 88.36 & 79.97 & 64.27 & 43.59 & 63.83 & 43.20 & 54.44 & 50.83 \\
encord-team/ebind-full & 59.20 & 58.53 & 60.67 & 44.78 & 38.09 & 5.07 & 61.92 & 88.36 & 79.97 & 64.27 & 43.59 & 63.83 & 43.20 & 54.44 & 50.83 \\
encord-team/ebind-points-vision & -- & 58.53 & -- & 44.78 & 38.09 & 5.07 & 61.92 & 88.36 & -- & 64.27 & -- & 63.83 & -- & 54.44 & 50.05 \\
facebook/pe-av-base & 59.75 & 57.59 & 57.56 & 36.44 & 49.43 & 2.95 & 43.19 & 76.39 & 80.60 & 59.60 & 53.70 & 61.03 & 55.15 & 51.02 & 50.13 \\
facebook/pe-av-large & 59.15 & 57.59 & 57.11 & 42.22 & 50.56 & 4.36 & 52.39 & 83.85 & 83.69 & 62.72 & 53.20 & 63.47 & 53.91 & 42.85 & 50.73 \\
facebook/pe-av-small & 58.78 & 57.11 & 56.00 & 39.44 & 46.19 & 4.12 & 42.93 & 80.44 & 84.01 & 60.02 & 53.08 & 60.21 & 54.17 & 50.70 & 50.21 \\
facebook/vjepa2-vitg-fpc32-384-diving48 & -- & 42.96 & -- & 41.78 & 30.47 & 5.86 & 51.98 & 69.10 & -- & 33.32 & -- & 32.93 & -- & 22.98 & 38.13 \\
facebook/vjepa2-vitg-fpc64-256 & -- & 42.14 & -- & 39.11 & 38.08 & 6.74 & 49.70 & 67.08 & -- & 32.11 & -- & 31.95 & -- & 21.92 & 37.70 \\
facebook/vjepa2-vitg-fpc64-384 & -- & 42.29 & -- & 41.33 & 35.99 & 5.86 & 49.65 & 66.92 & -- & 31.87 & -- & 31.50 & -- & 22.03 & 37.46 \\
facebook/vjepa2-vitg-fpc64-384-ssv2 & -- & 42.29 & -- & 41.33 & 35.99 & 5.86 & 49.65 & 66.92 & -- & 31.87 & -- & 31.50 & -- & 22.03 & 37.46 \\
facebook/vjepa2-vith-fpc64-256 & -- & 39.35 & -- & 40.56 & 28.17 & 6.49 & 42.16 & 68.02 & -- & 25.55 & -- & 26.25 & -- & 17.86 & 33.29 \\
facebook/vjepa2-vitl-fpc16-256-ssv2 & -- & 45.32 & -- & 40.33 & 24.00 & 6.48 & 47.26 & 69.89 & -- & 32.46 & -- & 33.38 & -- & 22.93 & 36.12 \\
facebook/vjepa2-vitl-fpc32-256-diving48 & -- & 44.68 & -- & 39.89 & 25.63 & 6.82 & 47.37 & 68.65 & -- & 32.32 & -- & 32.63 & -- & 22.52 & 35.98 \\
facebook/vjepa2-vitl-fpc64-256 & -- & 44.38 & -- & 39.44 & 30.95 & 6.50 & 45.61 & 70.04 & -- & 31.74 & -- & 32.06 & -- & 22.36 & 35.89 \\
jinaai/jina-embeddings-v5-omni-nano & 47.46 & 5.10 & 57.67 & 25.89 & 8.78 & 1.96 & 5.65 & 19.88 & 38.52 & 0.54 & 7.87 & 0.22 & 6.98 & 0.20 & 16.95 \\
jinaai/jina-embeddings-v5-omni-small & 48.66 & 5.17 & 57.00 & 24.67 & 8.54 & 1.75 & 5.15 & 22.20 & 37.11 & 0.25 & 7.92 & 0.30 & 6.89 & 0.21 & 17.20 \\
microsoft/xclip-base-patch16 & -- & 55.65 & -- & 41.00 & 30.02 & 6.88 & 66.76 & 85.56 & -- & 76.44 & -- & 69.80 & -- & 55.22 & 50.61 \\
microsoft/xclip-base-patch32 & -- & 54.38 & -- & 39.00 & 22.15 & 6.88 & 62.70 & 82.62 & -- & 70.89 & -- & 63.80 & -- & 48.81 & 47.44 \\
microsoft/xclip-large-patch14 & -- & 58.53 & -- & 42.56 & 40.64 & \textbf{7.37} & 67.82 & 85.87 & -- & \textbf{83.31} & -- & \textbf{76.59} & -- & \textbf{63.62} & 53.97 \\
VLM2Vec/VLM2Vec-V2.0 & -- & 53.91 & -- & 46.22 & 30.48 & 3.27 & 57.49 & 88.52 & -- & 56.28 & -- & 55.47 & -- & 44.40 & 45.98 \\
nvidia/omni-embed-nemotron-3b & 59.60 & 56.89 & 59.44 & 50.67 & 34.64 & 3.09 & 57.21 & 88.67 & 85.57 & 57.71 & 56.49 & 58.11 & 57.82 & 46.45 & 51.81 \\
zhibinlan/UME-R1-2B & -- & 56.87 & -- & 48.78 & 46.65 & 3.62 & 58.35 & 88.36 & -- & 61.87 & -- & 62.07 & -- & 51.69 & 50.26 \\
zhibinlan/UME-R1-7B & -- & 57.61 & -- & 50.22 & 43.41 & 3.73 & 59.29 & 92.55 & -- & 63.00 & -- & 63.75 & -- & 52.89 & 50.74 \\
\bottomrule
\end{tabular}%
}
\caption{\textbf{Classification} (28 tasks, accuracy), part 1 of 2.}
\label{tab:results-classification-a}
\end{table*}
\begin{table*}[ht]
\centering
\small
\setlength{\tabcolsep}{4pt}
\renewcommand{\arraystretch}{1.15}
\resizebox{\linewidth}{!}{%
\begin{tabular}{lccccccccccccccc}
\toprule
\textbf{Model} & \makecell[c]{Kinetics700 \\ audio + video} & \makecell[c]{MELD \\ audio + video} & \makecell[c]{MELD \\ video} & \makecell[c]{MusicAVQA \\ audio + video} & \makecell[c]{MusicAVQA \\ video} & \makecell[c]{RAVDESS \\ audio + video} & \makecell[c]{RAVDESS \\ video} & \makecell[c]{SSv2 \\ video} & \makecell[c]{UCF101 \\ audio + video} & \makecell[c]{UCF101 \\ video} & \makecell[c]{VGGSound \\ video} & \makecell[c]{VGGSound \\ audio + video} & \makecell[c]{WorldSense \\ audio + video} & \makecell[c]{WorldSense \\ video} & \textbf{Avg.} \\
\midrule
BidirLM/BidirLM-Omni-2.5B-Embedding & \textbf{58.10} & 18.21 & 15.09 & 56.97 & 56.09 & \textbf{56.88} & \textbf{51.53} & 13.37 & \textbf{93.89} & 94.19 & 47.36 & 49.85 & 67.92 & 64.76 & 56.89 \\
Haon-Chen/e5-omni-3B & 15.10 & 16.89 & 15.13 & 51.35 & 49.47 & 29.44 & 31.60 & 10.32 & 83.34 & 88.62 & 39.94 & 47.07 & 66.09 & 66.76 & 48.60 \\
Haon-Chen/e5-omni-7B & 48.67 & 17.58 & 15.33 & 53.40 & 51.81 & 29.65 & 32.92 & 13.06 & 91.13 & 91.91 & 42.58 & 52.84 & 67.62 & 65.04 & 52.23 \\
LCO-Embedding/LCO-Embedding-Omni-3B & 53.13 & 23.29 & 16.41 & \textbf{59.61} & 61.61 & 19.17 & 38.33 & 13.02 & 88.38 & 92.99 & \textbf{48.39} & 54.64 & 67.72 & 68.10 & 55.90 \\
LCO-Embedding/LCO-Embedding-Omni-7B & 56.07 & \textbf{25.10} & 15.77 & 59.15 & 58.73 & 28.75 & 43.19 & 17.23 & 88.80 & 94.27 & 48.14 & \textbf{56.80} & \textbf{68.96} & 69.34 & \textbf{57.43} \\
Qwen/Qwen2.5-Omni-3B & 0.52 & 19.68 & 14.25 & 12.90 & 7.56 & 21.81 & 23.33 & 0.59 & 14.31 & 14.26 & 0.32 & 4.75 & 42.70 & 35.24 & 14.19 \\
Qwen/Qwen2.5-Omni-7B & 0.58 & 14.50 & 12.40 & 15.06 & 8.91 & 28.61 & 24.38 & 0.59 & 15.74 & 15.87 & 0.32 & 0.32 & 37.25 & 43.27 & 14.17 \\
Qwen/Qwen3-VL-Embedding-2B & -- & -- & 15.33 & -- & 60.61 & -- & 45.28 & 13.46 & -- & 93.45 & 47.00 & -- & -- & 69.15 & 54.98 \\
Qwen/Qwen3-VL-Embedding-8B & -- & -- & 14.40 & -- & \textbf{62.37} & -- & 48.96 & \textbf{17.49} & -- & 94.46 & 48.30 & -- & -- & \textbf{70.49} & 56.81 \\
Tevatron/OmniEmbed-v0.1 & 46.80 & 19.68 & 14.50 & 50.41 & 52.05 & 44.72 & 36.46 & 14.68 & 90.11 & 92.17 & 42.88 & 50.74 & 67.43 & 65.52 & 53.33 \\
encord-team/ebind-audio-vision & 33.69 & 16.21 & 15.48 & 54.45 & 57.62 & 32.29 & 28.47 & 11.37 & 84.10 & 93.53 & 47.38 & 50.33 & 64.65 & 67.62 & 50.83 \\
encord-team/ebind-full & 33.69 & 16.21 & 15.48 & 54.45 & 57.62 & 32.29 & 28.47 & 11.37 & 84.10 & 93.52 & 47.38 & 50.33 & 64.65 & 67.62 & 50.83 \\
encord-team/ebind-points-vision & -- & -- & 15.48 & -- & 57.62 & -- & 28.47 & 11.37 & -- & 93.52 & 47.38 & -- & -- & 67.62 & 50.05 \\
facebook/pe-av-base & 45.25 & 15.23 & 13.77 & 55.04 & 55.92 & 35.90 & 36.53 & 6.91 & 81.13 & 84.54 & 44.69 & 53.53 & 64.46 & 66.38 & 50.13 \\
facebook/pe-av-large & 44.02 & 16.16 & 10.79 & 53.28 & 54.28 & 33.06 & 32.85 & 8.49 & 85.04 & 90.73 & 45.47 & 52.68 & 62.55 & 63.99 & 50.73 \\
facebook/pe-av-small & 44.41 & 15.04 & 13.87 & 53.46 & 54.75 & 38.47 & 32.99 & 6.69 & 84.81 & 88.17 & 43.85 & 52.21 & 64.18 & 65.90 & 50.21 \\
facebook/vjepa2-vitg-fpc32-384-diving48 & -- & -- & 15.72 & -- & 41.85 & -- & 39.31 & 14.86 & -- & 84.57 & 27.08 & -- & -- & 55.30 & 38.13 \\
facebook/vjepa2-vitg-fpc64-256 & -- & -- & 15.48 & -- & 41.97 & -- & 38.26 & 15.65 & -- & 83.29 & 26.31 & -- & -- & 53.39 & 37.70 \\
facebook/vjepa2-vitg-fpc64-384 & -- & -- & 15.33 & -- & 41.74 & -- & 36.11 & 15.63 & -- & 83.35 & 26.08 & -- & -- & 53.68 & 37.46 \\
facebook/vjepa2-vitg-fpc64-384-ssv2 & -- & -- & 15.33 & -- & 41.74 & -- & 36.11 & 15.63 & -- & 83.35 & 26.08 & -- & -- & 53.68 & 37.46 \\
facebook/vjepa2-vith-fpc64-256 & -- & -- & 14.85 & -- & 21.92 & -- & 34.72 & 13.98 & -- & 78.03 & 22.52 & -- & -- & 52.14 & 33.29 \\
facebook/vjepa2-vitl-fpc16-256-ssv2 & -- & -- & 16.16 & -- & 31.54 & -- & 33.89 & 14.29 & -- & 80.68 & 26.14 & -- & -- & 53.10 & 36.12 \\
facebook/vjepa2-vitl-fpc32-256-diving48 & -- & -- & 15.24 & -- & 32.59 & -- & 33.75 & 14.00 & -- & 81.28 & 25.76 & -- & -- & 52.52 & 35.98 \\
facebook/vjepa2-vitl-fpc64-256 & -- & -- & 16.02 & -- & 32.47 & -- & 29.72 & 13.76 & -- & 81.38 & 25.90 & -- & -- & 51.96 & 35.89 \\
jinaai/jina-embeddings-v5-omni-nano & 5.62 & 16.99 & 13.04 & 38.28 & 4.45 & 21.04 & 14.79 & 1.29 & 27.54 & 3.38 & 0.82 & 27.00 & 55.01 & 18.53 & 16.95 \\
jinaai/jina-embeddings-v5-omni-small & 3.29 & 17.58 & 10.79 & 38.98 & 5.04 & 22.43 & 15.49 & 0.72 & 31.38 & 3.07 & 0.32 & 26.95 & 59.40 & 20.43 & 17.20 \\
microsoft/xclip-base-patch16 & -- & -- & 15.19 & -- & 56.27 & -- & 45.49 & 8.60 & -- & 94.50 & 41.45 & -- & -- & 60.93 & 50.61 \\
microsoft/xclip-base-patch32 & -- & -- & 13.96 & -- & 49.35 & -- & 45.56 & 7.26 & -- & 91.97 & 38.64 & -- & -- & 61.13 & 47.44 \\
microsoft/xclip-large-patch14 & -- & -- & \textbf{16.89} & -- & 59.32 & -- & 47.78 & 9.53 & -- & \textbf{96.11} & 45.95 & -- & -- & 61.61 & 53.97 \\
VLM2Vec/VLM2Vec-V2.0 & -- & -- & 16.50 & -- & 49.53 & -- & 26.04 & 13.59 & -- & 88.65 & 41.17 & -- & -- & 64.08 & 45.98 \\
nvidia/omni-embed-nemotron-3b & 45.05 & 18.50 & 13.91 & 53.34 & 56.97 & 42.01 & 36.81 & 10.03 & 87.11 & 88.81 & 42.90 & 51.62 & 67.81 & 63.51 & 51.81 \\
zhibinlan/UME-R1-2B & -- & -- & 16.41 & -- & 51.06 & -- & 43.96 & 11.33 & -- & 91.08 & 44.71 & -- & -- & 67.43 & 50.26 \\
zhibinlan/UME-R1-7B & -- & -- & 14.99 & -- & 50.18 & -- & 40.83 & 13.67 & -- & 91.27 & 45.95 & -- & -- & 68.58 & 50.74 \\
\bottomrule
\end{tabular}%
}
\caption{\textbf{Classification} (28 tasks, accuracy), part 2 of 2.}
\label{tab:results-classification-b}
\end{table*}

\subsection{Zero-Shot Classification}
\label{appdx:results-zero-shot}

Per-task scores are in \autoref{tab:results-zero-shot-a} and \autoref{tab:results-zero-shot-b}. On the MVEB-curated zero-shot tasks (\autoref{tab:mveb-main-results}) eBind leads at 61.1, but on the full 26-task appendix average Qwen3-VL-Embedding-8B (59.8) edges ahead of LCO-Embedding-Omni-7B (58.3) and eBind (58.1) on the strength of Breakfast, HMDB51, HumAnimCartoon-video, MusicAVQA-video, and UCF101-video, where it tops every audio-capable model; per-column leadership on Kinetics, RAVDESS, VGGSound, and WorldSense still goes to audio-capable models (eBind, pe-av-large, BidirLM-Omni-2.5B), so Qwen3-VL-8B's lead is an across-column average rather than a clean sweep. The Qwen2.5-Omni generative-MLLM-as-embedder rows again collapse (3.8--16.5), confirming that hidden-state pooling without a text$\leftrightarrow$video contrastive stage cannot align video to label-text prompts.

\begin{table*}[ht]
\centering
\small
\setlength{\tabcolsep}{4pt}
\renewcommand{\arraystretch}{1.15}
\resizebox{\linewidth}{!}{%
\begin{tabular}{lcccccccccccccc}
\toprule
\textbf{Model} & \makecell[c]{AVEDataset \\ video} & \makecell[c]{AVEDataset \\ audio + video} & \makecell[c]{AVMeme \\ audio + video} & \makecell[c]{AVMeme \\ video} & \makecell[c]{Breakfast \\ video} & \makecell[c]{HMDB51 \\ video} & \makecell[c]{HumAnimCartoon \\ audio + video} & \makecell[c]{HumAnimCartoon \\ video} & \makecell[c]{Kinetics400 \\ audio + video} & \makecell[c]{Kinetics400 \\ video} & \makecell[c]{Kinetics600 \\ audio + video} & \makecell[c]{Kinetics600 \\ video} & \makecell[c]{Kinetics700 \\ audio + video} & \textbf{Avg.} \\
\midrule
BidirLM/BidirLM-Omni-2.5B-Embedding & 84.33 & 85.57 & 27.78 & 21.56 & 27.71 & 48.30 & 92.39 & 91.93 & 59.17 & 58.02 & 58.81 & 57.06 & 50.23 & 54.09 \\
Haon-Chen/e5-omni-3B & 76.87 & 64.93 & 19.78 & 12.78 & 31.64 & 40.78 & 85.40 & 89.75 & 46.31 & 55.84 & 43.16 & 54.30 & 33.43 & 43.18 \\
Haon-Chen/e5-omni-7B & 84.33 & 78.36 & 19.78 & 16.00 & 26.79 & 47.25 & 88.04 & 88.20 & 61.33 & 61.33 & 59.63 & 60.47 & 51.62 & 53.36 \\
LCO-Embedding/LCO-Embedding-Omni-3B & 88.81 & 91.29 & 27.78 & 21.89 & 43.19 & 52.94 & 90.99 & 91.61 & 65.43 & 69.31 & 63.90 & 67.65 & 55.70 & 55.56 \\
LCO-Embedding/LCO-Embedding-Omni-7B & 88.81 & 90.80 & \textbf{28.78} & 21.78 & 45.03 & 57.71 & \textbf{92.70} & 93.32 & 67.86 & 71.19 & 68.47 & 71.41 & \textbf{60.17} & 58.26 \\
Qwen/Qwen2.5-Omni-3B & 1.49 & 3.48 & 8.44 & 5.00 & 11.32 & 1.96 & 12.89 & 13.04 & 0.35 & 0.25 & 0.17 & 0.16 & 0.13 & 4.82 \\
Qwen/Qwen2.5-Omni-7B & 1.49 & 0.50 & 14.89 & 14.89 & 10.39 & 1.96 & 13.82 & 13.82 & 0.25 & 0.25 & 0.18 & 0.20 & 0.16 & 8.39 \\
Qwen/Qwen3-VL-Embedding-2B & 91.04 & -- & -- & 20.78 & 50.81 & 60.26 & -- & 94.41 & -- & 66.96 & -- & 66.53 & -- & 56.91 \\
Qwen/Qwen3-VL-Embedding-8B & 89.55 & -- & -- & 17.44 & \textbf{52.42} & \textbf{64.31} & -- & \textbf{95.19} & -- & 71.09 & -- & 70.92 & -- & \textbf{59.82} \\
Tevatron/OmniEmbed-v0.1 & 75.62 & 57.71 & 15.22 & 15.33 & 26.56 & 43.20 & 86.96 & 85.09 & 50.24 & 57.75 & 47.92 & 56.91 & 39.40 & 44.85 \\
encord-team/ebind-audio-vision & \textbf{92.54} & \textbf{94.03} & 26.00 & 21.67 & 34.41 & 56.60 & 90.37 & 93.79 & 66.83 & 72.59 & 65.52 & 72.43 & 58.11 & 58.11 \\
encord-team/ebind-full & \textbf{92.54} & \textbf{94.03} & 26.00 & 21.67 & 34.41 & 56.60 & 90.37 & 93.79 & 66.83 & 72.59 & 65.52 & 72.43 & 58.11 & 58.11 \\
encord-team/ebind-points-vision & \textbf{92.54} & -- & -- & 21.67 & 34.41 & 56.60 & -- & 93.79 & -- & 72.59 & -- & 72.43 & -- & 57.95 \\
facebook/pe-av-base & 90.80 & 91.54 & 25.33 & 19.89 & 27.48 & 39.22 & 89.75 & 90.53 & 69.89 & 71.76 & 69.89 & 72.48 & 59.31 & 52.73 \\
facebook/pe-av-large & 88.81 & 87.06 & 19.33 & 18.67 & 30.25 & 45.03 & 88.51 & 93.63 & 64.96 & 75.72 & 65.17 & \textbf{75.86} & 53.86 & 52.56 \\
facebook/pe-av-small & 85.57 & 90.55 & 20.11 & 15.22 & 32.10 & 38.95 & 87.58 & 89.60 & \textbf{70.51} & 72.49 & \textbf{70.86} & 72.78 & 58.91 & 53.51 \\
jinaai/jina-embeddings-v5-omni-nano & 3.98 & 60.95 & 22.56 & 13.89 & 10.16 & 2.48 & 39.60 & 16.30 & 6.48 & 0.33 & 5.30 & 0.16 & 4.33 & 14.73 \\
jinaai/jina-embeddings-v5-omni-small & 3.73 & 62.19 & 23.78 & 14.89 & 11.55 & 1.83 & 37.42 & 14.60 & 5.98 & 0.25 & 4.74 & 0.18 & 3.58 & 15.54 \\
microsoft/xclip-base-patch16 & 69.65 & -- & -- & 20.78 & 26.56 & 46.54 & -- & 84.63 & -- & 78.62 & -- & 60.56 & -- & 48.88 \\
microsoft/xclip-base-patch32 & 71.64 & -- & -- & 15.22 & 22.86 & 44.05 & -- & 85.09 & -- & 73.74 & -- & 58.03 & -- & 46.92 \\
microsoft/xclip-large-patch14 & 73.38 & -- & -- & 17.78 & 32.33 & 47.39 & -- & 86.34 & -- & \textbf{83.20} & -- & 65.51 & -- & 51.17 \\
VLM2Vec/VLM2Vec-V2.0 & 76.87 & -- & -- & 20.33 & 32.79 & 42.42 & -- & 84.94 & -- & 52.79 & -- & 50.39 & -- & 44.88 \\
nvidia/omni-embed-nemotron-3b & 59.45 & 62.44 & 17.78 & 14.78 & 31.41 & 35.03 & 87.58 & 83.85 & 36.80 & 35.89 & 33.89 & 31.12 & 28.40 & 38.52 \\
zhibinlan/UME-R1-2B & 83.08 & -- & -- & 18.78 & 36.95 & 51.11 & -- & 92.70 & -- & 62.75 & -- & 60.62 & -- & 52.21 \\
zhibinlan/UME-R1-7B & 85.82 & -- & -- & \textbf{22.00} & 39.26 & 50.39 & -- & 94.72 & -- & 65.96 & -- & 65.63 & -- & 53.99 \\
\bottomrule
\end{tabular}%
}
\caption{\textbf{Zero-shot classification} (26 tasks, accuracy), part 1 of 2.}
\label{tab:results-zero-shot-a}
\end{table*}
\begin{table*}[ht]
\centering
\small
\setlength{\tabcolsep}{4pt}
\renewcommand{\arraystretch}{1.15}
\resizebox{\linewidth}{!}{%
\begin{tabular}{lcccccccccccccc}
\toprule
\textbf{Model} & \makecell[c]{Kinetics700 \\ video} & \makecell[c]{MELD \\ audio + video} & \makecell[c]{MELD \\ video} & \makecell[c]{MusicAVQA \\ audio + video} & \makecell[c]{MusicAVQA \\ video} & \makecell[c]{RAVDESS \\ audio + video} & \makecell[c]{RAVDESS \\ video} & \makecell[c]{UCF101 \\ audio + video} & \makecell[c]{UCF101 \\ video} & \makecell[c]{VGGSound \\ audio + video} & \makecell[c]{VGGSound \\ video} & \makecell[c]{WorldSense \\ audio + video} & \makecell[c]{WorldSense \\ video} & \textbf{Avg.} \\
\midrule
BidirLM/BidirLM-Omni-2.5B-Embedding & 48.91 & 25.59 & 19.68 & 55.63 & 57.09 & 42.29 & \textbf{40.49} & 78.40 & 78.09 & 40.72 & 39.21 & 59.60 & 57.78 & 54.09 \\
Haon-Chen/e5-omni-3B & 45.31 & 11.77 & 7.86 & 52.99 & 57.21 & 28.68 & 22.01 & 66.98 & 70.37 & 28.28 & 31.48 & 20.82 & 24.07 & 43.18 \\
Haon-Chen/e5-omni-7B & 52.05 & \textbf{30.27} & 31.40 & 59.09 & 60.26 & 30.56 & 25.97 & 83.80 & 83.49 & 45.18 & 41.91 & 53.10 & 47.09 & 53.36 \\
LCO-Embedding/LCO-Embedding-Omni-3B & 59.79 & 21.29 & 15.28 & 60.49 & 62.66 & 21.53 & 20.83 & 78.19 & 81.12 & 46.50 & 44.84 & 52.53 & 49.00 & 55.56 \\
LCO-Embedding/LCO-Embedding-Omni-7B & 63.07 & 16.94 & 11.23 & \textbf{63.66} & 64.89 & 32.78 & 28.12 & \textbf{83.95} & 84.26 & 50.02 & 45.53 & 53.20 & 59.03 & 58.26 \\
Qwen/Qwen2.5-Omni-3B & 0.13 & 10.06 & 1.90 & 3.17 & 1.76 & 13.33 & 13.33 & 2.47 & 2.21 & 0.28 & 0.32 & 5.64 & 12.13 & 4.82 \\
Qwen/Qwen2.5-Omni-7B & 0.16 & 12.65 & \textbf{46.04} & 5.16 & 5.16 & 13.33 & 13.33 & 2.16 & 3.03 & 0.22 & 0.37 & 31.04 & 12.80 & 8.39 \\
Qwen/Qwen3-VL-Embedding-2B & 58.44 & -- & 38.62 & -- & 63.48 & -- & 26.46 & -- & 85.24 & -- & 44.60 & -- & 29.13 & 56.91 \\
Qwen/Qwen3-VL-Embedding-8B & 62.04 & -- & 24.80 & -- & \textbf{66.24} & -- & 35.35 & -- & \textbf{91.51} & -- & 47.39 & -- & 49.28 & \textbf{59.82} \\
Tevatron/OmniEmbed-v0.1 & 47.44 & 14.31 & 11.67 & 52.64 & 57.97 & 15.21 & 20.07 & 60.19 & 70.52 & 13.97 & 30.06 & 53.77 & 60.36 & 44.85 \\
encord-team/ebind-audio-vision & \textbf{65.52} & 12.35 & 17.53 & 63.31 & 65.47 & 29.79 & 27.64 & 75.41 & 79.37 & \textbf{52.29} & \textbf{49.58} & \textbf{65.52} & \textbf{62.18} & 58.11 \\
encord-team/ebind-full & \textbf{65.52} & 12.35 & 17.53 & 63.31 & 65.47 & 29.79 & 27.64 & 75.41 & 79.37 & \textbf{52.29} & \textbf{49.58} & \textbf{65.52} & \textbf{62.18} & 58.11 \\
encord-team/ebind-points-vision & \textbf{65.52} & -- & 17.53 & -- & 65.47 & -- & 27.64 & -- & 79.37 & -- & \textbf{49.58} & -- & \textbf{62.18} & 57.95 \\
facebook/pe-av-base & 61.46 & 18.90 & 17.48 & 56.68 & 57.27 & 25.35 & 29.17 & 70.83 & 77.21 & 41.72 & 37.61 & 29.89 & 29.61 & 52.73 \\
facebook/pe-av-large & 64.83 & 13.18 & 12.65 & 58.44 & 59.09 & 16.18 & 18.33 & 77.21 & 86.27 & 40.64 & 42.35 & 30.75 & 39.83 & 52.56 \\
facebook/pe-av-small & 61.60 & 19.09 & 26.22 & 59.14 & 58.79 & 36.18 & 19.24 & 80.25 & 79.89 & 39.17 & 36.48 & 31.90 & 38.20 & 53.51 \\
jinaai/jina-embeddings-v5-omni-nano & 0.20 & 28.76 & 17.53 & 39.10 & 8.03 & 18.06 & 6.67 & 15.48 & 1.59 & 15.15 & 0.25 & 37.44 & 8.31 & 14.73 \\
jinaai/jina-embeddings-v5-omni-small & 0.16 & 22.80 & 10.89 & 39.39 & 8.79 & 25.07 & 6.60 & 14.97 & 1.75 & 14.93 & 0.30 & 45.85 & 27.79 & 15.54 \\
microsoft/xclip-base-patch16 & 48.07 & -- & 16.16 & -- & 53.93 & -- & 34.38 & -- & 75.15 & -- & 29.95 & -- & 39.35 & 48.88 \\
microsoft/xclip-base-patch32 & 44.91 & -- & 10.74 & -- & 52.64 & -- & 28.06 & -- & 74.28 & -- & 27.95 & -- & 47.66 & 46.92 \\
microsoft/xclip-large-patch14 & 54.19 & -- & 21.00 & -- & 55.98 & -- & 34.03 & -- & 80.61 & -- & 33.09 & -- & 31.52 & 51.17 \\
VLM2Vec/VLM2Vec-V2.0 & 42.88 & -- & 7.96 & -- & 58.21 & -- & 7.22 & -- & 66.72 & -- & 32.29 & -- & 52.53 & 44.88 \\
nvidia/omni-embed-nemotron-3b & 26.58 & 10.60 & 9.91 & 31.13 & 26.44 & \textbf{46.04} & 32.92 & 59.88 & 56.64 & 22.86 & 19.27 & 50.05 & 50.72 & 38.52 \\
zhibinlan/UME-R1-2B & 52.44 & -- & 25.34 & -- & 61.31 & -- & 28.75 & -- & 76.23 & -- & 41.56 & -- & 39.26 & 52.21 \\
zhibinlan/UME-R1-7B & 56.99 & -- & 16.65 & -- & 62.43 & -- & 34.17 & -- & 83.02 & -- & 44.44 & -- & 34.38 & 53.99 \\
\bottomrule
\end{tabular}%
}
\caption{\textbf{Zero-shot classification} (26 tasks, accuracy), part 2 of 2.}
\label{tab:results-zero-shot-b}
\end{table*}

\subsection{Clustering}
\label{appdx:results-clustering}

Per-task scores are in \autoref{tab:results-clustering-a} and \autoref{tab:results-clustering-b}. Clustering remains universally hard on the curated MVEB pair. The strongest MVEB-curated clustering average is 27.3 ($v$-measure) by LCO-Embedding-Omni-7B, with runners-up in the 20--27 range; the bottleneck is the emotion-clustering task (MELDEmotion AV) where no model exceeds 8.8, while music-clustering (MusicAVQA AV) is comparatively easier (top scores in the 38--47 range). Across the wider 15-task appendix pool the picture changes: Qwen3-VL-Embedding-8B leads at 51.3 driven by strong UCF101 and HMDB51 splits where models reach 75--95. Adding the audio channel changes clustering scores very little on the curated pair, suggesting the headline-clustering bottleneck is the embedding geometry on fine-grained emotion labels rather than a missing modality.

\begin{table*}[ht]
\centering
\small
\setlength{\tabcolsep}{4pt}
\renewcommand{\arraystretch}{1.15}
\resizebox{\linewidth}{!}{%
\begin{tabular}{lccccccccc}
\toprule
\textbf{Model} & \makecell[c]{AVEDataset \\ audio + video} & \makecell[c]{AVEDataset \\ video} & \makecell[c]{HMDB51 \\ video} & \makecell[c]{MELDEmotion \\ audio + video} & \makecell[c]{MELDEmotion \\ video} & \makecell[c]{MELDSpeaker \\ audio + video} & \makecell[c]{MELDSpeaker \\ video} & \makecell[c]{MusicAVQA \\ audio + video} & \textbf{Avg.} \\
\midrule
BidirLM/BidirLM-Omni-2.5B-Embedding & 80.77 & 79.54 & 69.89 & 0.84 & 0.81 & 31.51 & 31.58 & 33.39 & 45.17 \\
Haon-Chen/e5-omni-3B & 82.66 & 79.43 & 69.56 & 1.66 & 1.01 & 27.04 & 30.05 & 41.17 & 43.63 \\
Haon-Chen/e5-omni-7B & 80.59 & 75.96 & 68.70 & 1.85 & 0.80 & 30.91 & 32.58 & 41.22 & 43.25 \\
LCO-Embedding/LCO-Embedding-Omni-3B & \textbf{87.38} & \textbf{85.49} & 72.83 & 7.30 & 0.96 & 20.03 & 31.06 & \textbf{46.73} & 46.65 \\
LCO-Embedding/LCO-Embedding-Omni-7B & 87.31 & 83.77 & 73.82 & \textbf{8.75} & 1.21 & 21.60 & 32.67 & 45.95 & 46.49 \\
Qwen/Qwen2.5-Omni-3B & 44.07 & 37.29 & 39.89 & 2.26 & \textbf{1.96} & 16.17 & 17.67 & 9.62 & 20.58 \\
Qwen/Qwen2.5-Omni-7B & 41.08 & 37.65 & 38.06 & 2.06 & 1.87 & 15.95 & 17.12 & 12.86 & 20.91 \\
Qwen/Qwen3-VL-Embedding-2B & -- & 83.18 & 72.58 & -- & 1.01 & -- & 28.78 & -- & 50.07 \\
Qwen/Qwen3-VL-Embedding-8B & -- & 84.02 & \textbf{75.47} & -- & 1.46 & -- & 31.33 & -- & \textbf{51.30} \\
Tevatron/OmniEmbed-v0.1 & 81.23 & 75.80 & 71.14 & 1.96 & 1.26 & 27.89 & 29.55 & 39.31 & 43.94 \\
encord-team/ebind-audio-vision & 83.68 & 83.73 & 72.04 & 1.78 & 1.15 & \textbf{36.72} & \textbf{41.51} & 38.42 & 45.34 \\
encord-team/ebind-full & 83.68 & 83.73 & 72.04 & 1.78 & 1.15 & \textbf{36.72} & \textbf{41.51} & 38.42 & 45.34 \\
encord-team/ebind-points-vision & -- & 83.73 & 72.04 & -- & 1.15 & -- & \textbf{41.51} & -- & 47.82 \\
facebook/pe-av-base & 82.48 & 80.79 & 58.41 & 2.11 & 1.85 & 28.27 & 22.48 & 43.04 & 41.91 \\
facebook/pe-av-large & 80.51 & 81.04 & 64.07 & 2.11 & 1.81 & 29.31 & 24.18 & 41.30 & 42.83 \\
facebook/pe-av-small & 81.18 & 77.13 & 57.09 & 2.07 & 1.76 & 28.49 & 22.71 & 43.74 & 41.52 \\
facebook/vjepa2-vitg-fpc32-384-diving48 & -- & 51.68 & 55.68 & -- & 1.34 & -- & 22.39 & -- & 32.66 \\
facebook/vjepa2-vitg-fpc64-256 & -- & 51.38 & 55.02 & -- & 1.39 & -- & 23.01 & -- & 32.54 \\
facebook/vjepa2-vitg-fpc64-384 & -- & 49.88 & 54.76 & -- & 1.07 & -- & 22.68 & -- & 31.80 \\
facebook/vjepa2-vitg-fpc64-384-ssv2 & -- & 49.88 & 54.76 & -- & 1.07 & -- & 22.68 & -- & 31.80 \\
facebook/vjepa2-vith-fpc64-256 & -- & 46.63 & 51.88 & -- & 1.07 & -- & 22.69 & -- & 29.12 \\
facebook/vjepa2-vitl-fpc16-256-ssv2 & -- & 57.77 & 58.07 & -- & 1.13 & -- & 23.61 & -- & 34.52 \\
facebook/vjepa2-vitl-fpc32-256-diving48 & -- & 57.90 & 56.53 & -- & 1.17 & -- & 23.73 & -- & 34.35 \\
facebook/vjepa2-vitl-fpc64-256 & -- & 57.01 & 55.76 & -- & 1.22 & -- & 24.00 & -- & 34.28 \\
jinaai/jina-embeddings-v5-omni-nano & 56.22 & 27.41 & 34.21 & 2.17 & 1.84 & 16.59 & 17.11 & 22.95 & 20.66 \\
jinaai/jina-embeddings-v5-omni-small & 60.81 & 28.58 & 33.63 & 1.63 & 1.48 & 16.53 & 17.14 & 25.81 & 22.28 \\
microsoft/xclip-base-patch16 & -- & 75.87 & 72.76 & -- & 1.07 & -- & 29.35 & -- & 45.62 \\
microsoft/xclip-base-patch32 & -- & 73.71 & 70.88 & -- & 1.09 & -- & 26.34 & -- & 43.89 \\
microsoft/xclip-large-patch14 & -- & 79.20 & 73.29 & -- & 1.25 & -- & 31.85 & -- & 46.20 \\
VLM2Vec/VLM2Vec-V2.0 & -- & 73.02 & 67.71 & -- & 0.95 & -- & 32.05 & -- & 42.19 \\
nvidia/omni-embed-nemotron-3b & 85.14 & 79.40 & 68.54 & 1.52 & 1.02 & 27.27 & 30.21 & 39.40 & 43.89 \\
zhibinlan/UME-R1-2B & -- & 78.02 & 69.47 & -- & 0.96 & -- & 32.40 & -- & 44.74 \\
zhibinlan/UME-R1-7B & -- & 78.19 & 68.90 & -- & 1.21 & -- & 33.61 & -- & 44.71 \\
\bottomrule
\end{tabular}%
}
\caption{\textbf{Clustering} (15 tasks, v-measure), part 1 of 2.}
\label{tab:results-clustering-a}
\end{table*}
\begin{table*}[ht]
\centering
\small
\setlength{\tabcolsep}{4pt}
\renewcommand{\arraystretch}{1.15}
\resizebox{\linewidth}{!}{%
\begin{tabular}{lcccccccc}
\toprule
\textbf{Model} & \makecell[c]{MusicAVQA \\ video} & \makecell[c]{RAVDESS \\ audio + video} & \makecell[c]{RAVDESS \\ video} & \makecell[c]{UCF101 \\ audio + video} & \makecell[c]{UCF101 \\ video} & \makecell[c]{WS-1m-Domain \\ audio + video} & \makecell[c]{WS-1m-Domain \\ video} & \textbf{Avg.} \\
\midrule
BidirLM/BidirLM-Omni-2.5B-Embedding & 32.75 & 18.95 & 21.57 & \textbf{93.09} & 94.05 & \textbf{46.67} & 42.09 & 45.17 \\
Haon-Chen/e5-omni-3B & 34.66 & 7.27 & 7.96 & 87.53 & 91.33 & 44.45 & 48.70 & 43.63 \\
Haon-Chen/e5-omni-7B & 36.86 & 3.71 & 3.45 & 90.55 & 91.58 & 45.26 & 44.69 & 43.25 \\
LCO-Embedding/LCO-Embedding-Omni-3B & 48.44 & 7.41 & 17.60 & 88.44 & 93.75 & 46.02 & 46.32 & 46.65 \\
LCO-Embedding/LCO-Embedding-Omni-7B & 39.93 & 7.90 & 20.80 & 88.37 & 94.14 & 44.96 & 46.19 & 46.49 \\
Qwen/Qwen2.5-Omni-3B & 8.63 & 6.62 & 7.92 & 38.92 & 38.54 & 27.56 & 11.60 & 20.58 \\
Qwen/Qwen2.5-Omni-7B & 9.72 & 9.18 & 7.35 & 37.81 & 36.58 & 27.26 & 19.06 & 20.91 \\
Qwen/Qwen3-VL-Embedding-2B & 44.15 & -- & \textbf{24.89} & -- & 93.81 & -- & \textbf{52.14} & 50.07 \\
Qwen/Qwen3-VL-Embedding-8B & \textbf{48.95} & -- & 22.16 & -- & \textbf{95.14} & -- & 51.90 & \textbf{51.30} \\
Tevatron/OmniEmbed-v0.1 & 35.88 & 10.97 & 10.65 & 90.23 & 92.39 & 45.45 & 45.34 & 43.94 \\
encord-team/ebind-audio-vision & 44.33 & 11.20 & 1.37 & 82.86 & 93.06 & 42.89 & 45.39 & 45.34 \\
encord-team/ebind-full & 44.33 & 11.20 & 1.37 & 82.86 & 93.06 & 42.89 & 45.39 & 45.34 \\
encord-team/ebind-points-vision & 44.33 & -- & 1.37 & -- & 93.06 & -- & 45.39 & 47.82 \\
facebook/pe-av-base & 43.58 & 7.09 & 10.67 & 79.72 & 80.74 & 42.75 & 44.66 & 41.91 \\
facebook/pe-av-large & 42.52 & 6.03 & 11.53 & 85.88 & 87.70 & 41.13 & 43.33 & 42.83 \\
facebook/pe-av-small & 41.32 & 6.81 & 8.99 & 82.00 & 84.64 & 41.76 & 43.07 & 41.52 \\
facebook/vjepa2-vitg-fpc32-384-diving48 & 21.44 & -- & 5.42 & -- & 76.47 & -- & 26.87 & 32.66 \\
facebook/vjepa2-vitg-fpc64-256 & 21.93 & -- & 6.24 & -- & 75.72 & -- & 25.66 & 32.54 \\
facebook/vjepa2-vitg-fpc64-384 & 20.83 & -- & 5.04 & -- & 74.60 & -- & 25.53 & 31.80 \\
facebook/vjepa2-vitg-fpc64-384-ssv2 & 20.83 & -- & 5.04 & -- & 74.60 & -- & 25.53 & 31.80 \\
facebook/vjepa2-vith-fpc64-256 & 9.84 & -- & 5.95 & -- & 71.42 & -- & 23.48 & 29.12 \\
facebook/vjepa2-vitl-fpc16-256-ssv2 & 25.43 & -- & 6.86 & -- & 78.72 & -- & 24.57 & 34.52 \\
facebook/vjepa2-vitl-fpc32-256-diving48 & 25.69 & -- & 7.25 & -- & 78.54 & -- & 23.98 & 34.35 \\
facebook/vjepa2-vitl-fpc64-256 & 25.70 & -- & 6.95 & -- & 78.01 & -- & 25.58 & 34.28 \\
jinaai/jina-embeddings-v5-omni-nano & 5.74 & 15.59 & 0.96 & 46.53 & 29.80 & 27.14 & 5.61 & 20.66 \\
jinaai/jina-embeddings-v5-omni-small & 5.37 & \textbf{22.32} & 1.16 & 52.60 & 31.22 & 30.10 & 5.82 & 22.28 \\
microsoft/xclip-base-patch16 & 44.89 & -- & 10.62 & -- & 93.84 & -- & 36.57 & 45.62 \\
microsoft/xclip-base-patch32 & 40.62 & -- & 11.14 & -- & 92.38 & -- & 34.96 & 43.89 \\
microsoft/xclip-large-patch14 & 48.60 & -- & 8.17 & -- & 94.64 & -- & 32.63 & 46.20 \\
VLM2Vec/VLM2Vec-V2.0 & 32.41 & -- & 0.17 & -- & 88.91 & -- & 42.31 & 42.19 \\
nvidia/omni-embed-nemotron-3b & 40.96 & 8.19 & 10.92 & 88.66 & 90.87 & 45.53 & 40.70 & 43.89 \\
zhibinlan/UME-R1-2B & 29.62 & -- & 14.93 & -- & 90.27 & -- & 42.26 & 44.74 \\
zhibinlan/UME-R1-7B & 30.15 & -- & 8.79 & -- & 91.27 & -- & 45.54 & 44.71 \\
\bottomrule
\end{tabular}%
}
\caption{\textbf{Clustering} (15 tasks, v-measure), part 2 of 2.}
\label{tab:results-clustering-b}
\end{table*}

\subsection{Pair Classification}
\label{appdx:results-pair-cls}

Per-task scores are in \autoref{tab:results-pair-cls-a} and \autoref{tab:results-pair-cls-b}. Pair classification is the easiest category for top models, with the leader (LCO-Embedding-Omni-3B at 80.7 max-AP) just above LCO-Embedding-Omni-7B (79.6) and Perception Encoder large (75.2). The MLLM-embed and audio-visual contrastive families cluster within four points of each other. On the 12-task appendix average, Qwen3-VL-Embedding-2B/8B lead at 75.0/74.5 because they consistently score high on the video-only pair-classification splits where audio is unavailable. Smaller multimodal-binding (eBind) and audio-visual contrastive checkpoints (PE-AV-small/base) are noticeably behind the MLLM-embed leaders on this category.

\begin{table*}[ht]
\centering
\small
\setlength{\tabcolsep}{4pt}
\renewcommand{\arraystretch}{1.15}
\resizebox{\linewidth}{!}{%
\begin{tabular}{lccccccc}
\toprule
\textbf{Model} & \makecell[c]{AVEDataset \\ audio + video} & \makecell[c]{AVEDataset \\ video} & \makecell[c]{AVSpeakerBench \\ video} & \makecell[c]{HumAnimCartoon \\ audio + video} & \makecell[c]{HumAnimCartoon \\ video} & \makecell[c]{MELD \\ audio + video} & \textbf{Avg.} \\
\midrule
BidirLM/BidirLM-Omni-2.5B-Embedding & 96.11 & 95.80 & 95.89 & \textbf{84.93} & 85.70 & 53.44 & 73.85 \\
Haon-Chen/e5-omni-3B & 96.14 & 94.52 & 97.09 & 77.29 & 82.14 & 51.19 & 72.03 \\
Haon-Chen/e5-omni-7B & 94.70 & 93.90 & 98.01 & 80.71 & 84.29 & \textbf{54.34} & 72.38 \\
LCO-Embedding/LCO-Embedding-Omni-3B & 95.60 & \textbf{97.03} & 97.46 & 82.75 & 87.74 & 53.40 & 74.31 \\
LCO-Embedding/LCO-Embedding-Omni-7B & 95.21 & 96.11 & 97.43 & 81.39 & 85.91 & 53.69 & 73.91 \\
Qwen/Qwen2.5-Omni-3B & 68.69 & 61.45 & 78.62 & 52.44 & 54.49 & 53.69 & 57.99 \\
Qwen/Qwen2.5-Omni-7B & 61.64 & 57.79 & 75.73 & 51.50 & 53.11 & 51.57 & 56.47 \\
Qwen/Qwen3-VL-Embedding-2B & -- & 96.16 & 96.65 & -- & \textbf{88.35} & -- & \textbf{75.02} \\
Qwen/Qwen3-VL-Embedding-8B & -- & 95.32 & 97.07 & -- & 86.75 & -- & 74.45 \\
Tevatron/OmniEmbed-v0.1 & 94.45 & 93.39 & 98.04 & 75.94 & 81.81 & 52.82 & 72.32 \\
encord-team/ebind-audio-vision & \textbf{96.25} & 95.67 & \textbf{99.15} & 73.35 & 78.18 & 49.27 & 71.93 \\
encord-team/ebind-full & \textbf{96.25} & 95.67 & \textbf{99.15} & 73.35 & 78.18 & 49.27 & 71.93 \\
encord-team/ebind-points-vision & -- & 95.67 & \textbf{99.15} & -- & 78.18 & -- & 72.51 \\
facebook/pe-av-base & 90.83 & 92.82 & 98.01 & 76.32 & 75.24 & 52.59 & 72.12 \\
facebook/pe-av-large & 90.69 & 92.67 & 98.23 & 76.45 & 79.24 & 54.24 & 72.01 \\
facebook/pe-av-small & 91.56 & 93.24 & 98.23 & 76.02 & 76.54 & 53.10 & 71.99 \\
facebook/vjepa2-vitg-fpc32-384-diving48 & -- & 78.76 & 94.08 & -- & 68.60 & -- & 66.27 \\
facebook/vjepa2-vitg-fpc64-256 & -- & 77.60 & 93.38 & -- & 68.08 & -- & 66.16 \\
facebook/vjepa2-vitg-fpc64-384 & -- & 75.42 & 92.84 & -- & 67.56 & -- & 65.28 \\
facebook/vjepa2-vitg-fpc64-384-ssv2 & -- & 75.42 & 92.84 & -- & 67.56 & -- & 65.28 \\
facebook/vjepa2-vith-fpc64-256 & -- & 73.52 & 94.14 & -- & 68.76 & -- & 64.52 \\
facebook/vjepa2-vitl-fpc16-256-ssv2 & -- & 81.53 & 93.36 & -- & 70.05 & -- & 66.25 \\
facebook/vjepa2-vitl-fpc32-256-diving48 & -- & 81.82 & 93.49 & -- & 71.38 & -- & 66.61 \\
facebook/vjepa2-vitl-fpc64-256 & -- & 82.84 & 93.80 & -- & 72.85 & -- & 67.19 \\
jinaai/jina-embeddings-v5-omni-nano & 80.88 & 53.25 & 71.96 & 55.29 & 51.47 & 49.88 & 57.35 \\
jinaai/jina-embeddings-v5-omni-small & 76.29 & 54.96 & 80.49 & 54.09 & 52.38 & 50.43 & 57.87 \\
microsoft/xclip-base-patch16 & -- & 91.61 & 96.94 & -- & 74.86 & -- & 72.80 \\
microsoft/xclip-base-patch32 & -- & 90.92 & 96.36 & -- & 74.74 & -- & 72.27 \\
microsoft/xclip-large-patch14 & -- & 90.64 & 97.31 & -- & 77.33 & -- & 73.07 \\
VLM2Vec/VLM2Vec-V2.0 & -- & 91.96 & 97.06 & -- & 81.30 & -- & 71.48 \\
nvidia/omni-embed-nemotron-3b & 94.54 & 92.91 & 97.56 & 79.28 & 82.86 & 53.16 & 72.49 \\
zhibinlan/UME-R1-2B & -- & 95.16 & 97.37 & -- & 82.12 & -- & 73.51 \\
zhibinlan/UME-R1-7B & -- & 96.04 & 98.29 & -- & 84.05 & -- & 73.75 \\
\bottomrule
\end{tabular}%
}
\caption{\textbf{Pair classification} (12 of 13 MVEB+ tasks, max-AP; VideoConPairClassification is in the MVEB+ pool but has no evaluations in this release), part 1 of 2.}
\label{tab:results-pair-cls-a}
\end{table*}
\begin{table*}[ht]
\centering
\small
\setlength{\tabcolsep}{4pt}
\renewcommand{\arraystretch}{1.15}
\resizebox{\linewidth}{!}{%
\begin{tabular}{lccccccc}
\toprule
\textbf{Model} & \makecell[c]{MELD \\ video} & \makecell[c]{MusicAVQA \\ audio + video} & \makecell[c]{MusicAVQ \\ video} & \makecell[c]{RAVDESSAV \\ audio + video} & \makecell[c]{RAVDESSAV \\ video} & \makecell[c]{Vinoground \\ video} & \textbf{Avg.} \\
\midrule
BidirLM/BidirLM-Omni-2.5B-Embedding & 51.74 & 72.57 & 72.23 & \textbf{62.39} & 63.59 & 51.83 & 73.85 \\
Haon-Chen/e5-omni-3B & 51.13 & 76.73 & 72.14 & 56.78 & 56.83 & 52.36 & 72.03 \\
Haon-Chen/e5-omni-7B & 52.82 & 73.92 & 71.15 & 55.11 & 56.67 & 52.92 & 72.38 \\
LCO-Embedding/LCO-Embedding-Omni-3B & 50.61 & \textbf{78.64} & 79.98 & 55.55 & 59.83 & 53.10 & 74.31 \\
LCO-Embedding/LCO-Embedding-Omni-7B & 50.69 & 77.87 & 76.18 & 56.30 & 62.57 & 53.52 & 73.91 \\
Qwen/Qwen2.5-Omni-3B & 52.66 & 54.91 & 55.05 & 56.50 & 56.30 & 51.06 & 57.99 \\
Qwen/Qwen2.5-Omni-7B & 53.03 & 54.98 & 55.09 & 57.71 & 54.70 & 50.77 & 56.47 \\
Qwen/Qwen3-VL-Embedding-2B & 52.19 & -- & 78.55 & -- & 60.18 & 53.02 & \textbf{75.02} \\
Qwen/Qwen3-VL-Embedding-8B & 52.79 & -- & 77.18 & -- & 59.77 & 52.27 & 74.45 \\
Tevatron/OmniEmbed-v0.1 & 51.00 & 73.41 & 72.84 & 61.38 & 59.57 & 53.14 & 72.32 \\
encord-team/ebind-audio-vision & 50.54 & 77.43 & 77.20 & 59.34 & 55.45 & 51.36 & 71.93 \\
encord-team/ebind-full & 50.54 & 77.43 & 77.20 & 59.34 & 55.45 & 51.36 & 71.93 \\
encord-team/ebind-points-vision & 50.54 & -- & 77.20 & -- & 55.45 & 51.36 & 72.51 \\
facebook/pe-av-base & \textbf{54.62} & 75.17 & 76.26 & 58.11 & 62.45 & 52.99 & 72.12 \\
facebook/pe-av-large & 54.20 & 74.04 & 76.77 & 57.56 & 57.52 & 52.55 & 72.01 \\
facebook/pe-av-small & 52.72 & 74.61 & 75.62 & 59.52 & 58.89 & 53.87 & 71.99 \\
facebook/vjepa2-vitg-fpc32-384-diving48 & 52.75 & -- & 62.81 & -- & 56.22 & 50.68 & 66.27 \\
facebook/vjepa2-vitg-fpc64-256 & 51.91 & -- & 62.88 & -- & 54.15 & \textbf{55.09} & 66.16 \\
facebook/vjepa2-vitg-fpc64-384 & 51.76 & -- & 61.80 & -- & 53.78 & 53.82 & 65.28 \\
facebook/vjepa2-vitg-fpc64-384-ssv2 & 51.76 & -- & 61.80 & -- & 53.78 & 53.82 & 65.28 \\
facebook/vjepa2-vith-fpc64-256 & 52.35 & -- & 55.33 & -- & 54.35 & 53.21 & 64.52 \\
facebook/vjepa2-vitl-fpc16-256-ssv2 & 51.73 & -- & 62.83 & -- & 54.52 & 49.72 & 66.25 \\
facebook/vjepa2-vitl-fpc32-256-diving48 & 51.70 & -- & 63.26 & -- & 54.91 & 49.72 & 66.61 \\
facebook/vjepa2-vitl-fpc64-256 & 51.51 & -- & 63.52 & -- & 54.20 & 51.60 & 67.19 \\
jinaai/jina-embeddings-v5-omni-nano & 51.23 & 62.85 & 49.63 & 60.66 & 50.67 & 50.38 & 57.35 \\
jinaai/jina-embeddings-v5-omni-small & 51.05 & 63.17 & 49.51 & 60.02 & 50.52 & 51.49 & 57.87 \\
microsoft/xclip-base-patch16 & 49.98 & -- & 78.18 & -- & \textbf{63.66} & 54.37 & 72.80 \\
microsoft/xclip-base-patch32 & 50.35 & -- & 76.46 & -- & 63.29 & 53.80 & 72.27 \\
microsoft/xclip-large-patch14 & 49.42 & -- & \textbf{81.33} & -- & 62.31 & 53.11 & 73.07 \\
VLM2Vec/VLM2Vec-V2.0 & 51.49 & -- & 69.64 & -- & 54.46 & 54.48 & 71.48 \\
nvidia/omni-embed-nemotron-3b & 52.21 & 76.04 & 73.26 & 58.96 & 57.69 & 51.38 & 72.49 \\
zhibinlan/UME-R1-2B & 51.41 & -- & 72.60 & -- & 62.56 & 53.37 & 73.51 \\
zhibinlan/UME-R1-7B & 51.30 & -- & 72.33 & -- & 60.85 & 53.39 & 73.75 \\
\bottomrule
\end{tabular}%
}
\caption{\textbf{Pair classification} (12 of 13 MVEB+ tasks, max-AP; VideoConPairClassification is in the MVEB+ pool but has no evaluations in this release), part 2 of 2.}
\label{tab:results-pair-cls-b}
\end{table*}

\subsection{Question Answering}
\label{appdx:results-qa}

Per-task scores are in \autoref{tab:results-qa-a} and \autoref{tab:results-qa-b}. LCO-Embedding-Omni-7B leads QA at 57.0, with the smaller LCO-Embedding-Omni-3B close behind at 56.4 and e5-omni-3B at 45.4. Qwen3-VL-Embedding-8B and BidirLM-Omni-2.5B are next strongest among the newer MLLM-embed entrants. The Perception Encoder audio-visual family scores 27.8--30.6 here, well behind the MLLM-embed leaders. This is a paradigm gap rather than a scale gap, since PE-AV-large (2.2B) loses to LCO-Embedding-Omni-3B (4.7B) by more than 25 points. Video-centric QA benefits more from large language-model context than from larger video-only feature extractors.

\begin{table*}[ht]
\centering
\small
\setlength{\tabcolsep}{4pt}
\renewcommand{\arraystretch}{1.15}
\resizebox{\linewidth}{!}{%
\begin{tabular}{lccccccccccc}
\toprule
\textbf{Model} & \makecell[c]{AVMemeExam \\ audio + video} & \makecell[c]{AVMemeExam \\ video} & \makecell[c]{AVQA \\ audio + video} & \makecell[c]{AVQA \\ video} & \makecell[c]{AVSpeakerBench \\ audio + video} & \makecell[c]{AVSpeakerBench \\ video} & \makecell[c]{DailyOmni \\ audio + video} & \makecell[c]{DailyOmni \\ video} & \makecell[c]{EgoSchema \\ video} & \makecell[c]{NExTQA \\ video} & \textbf{Avg.} \\
\midrule
BidirLM/BidirLM-Omni-2.5B-Embedding & 36.44 & 31.89 & 79.15 & 77.96 & 29.17 & 28.58 & 27.84 & 27.43 & 55.80 & 46.32 & 40.36 \\
Haon-Chen/e5-omni-3B & 37.44 & 35.67 & 72.64 & 75.46 & 29.61 & 29.67 & 32.86 & 31.86 & 45.40 & 51.76 & 42.65 \\
Haon-Chen/e5-omni-7B & 41.67 & 38.00 & 79.59 & 77.96 & 32.16 & 31.35 & 34.87 & 32.94 & 43.80 & 58.31 & 46.22 \\
LCO-Embedding/LCO-Embedding-Omni-3B & 40.00 & 38.56 & 83.50 & 82.30 & \textbf{34.00} & 33.41 & 42.89 & \textbf{36.96} & 56.40 & 66.47 & 49.77 \\
LCO-Embedding/LCO-Embedding-Omni-7B & \textbf{42.22} & \textbf{40.11} & \textbf{85.67} & 83.17 & 33.84 & \textbf{34.34} & \textbf{43.90} & 36.71 & 57.00 & \textbf{70.80} & \textbf{51.81} \\
Qwen/Qwen2.5-Omni-3B & 26.67 & 27.33 & 24.86 & 24.32 & 26.99 & 25.96 & 24.08 & 23.33 & 14.40 & 21.15 & 25.19 \\
Qwen/Qwen2.5-Omni-7B & 23.78 & 23.00 & 24.00 & 26.06 & 24.66 & 25.28 & 23.49 & 23.58 & 13.20 & 19.13 & 23.91 \\
Qwen/Qwen3-VL-Embedding-2B & -- & 38.11 & -- & 81.98 & -- & 32.04 & -- & 29.43 & \textbf{63.40} & 55.29 & 44.76 \\
Qwen/Qwen3-VL-Embedding-8B & -- & 37.44 & -- & \textbf{83.50} & -- & 32.56 & -- & 31.02 & 56.80 & 58.61 & 46.09 \\
Tevatron/OmniEmbed-v0.1 & 32.56 & 33.44 & 67.32 & 68.73 & 28.71 & 29.14 & 28.85 & 27.59 & 50.20 & 41.99 & 37.53 \\
encord-team/ebind-audio-vision & 32.44 & 30.67 & 80.02 & 73.18 & 29.08 & 29.76 & 26.92 & 26.67 & 31.40 & 43.30 & 36.37 \\
encord-team/ebind-full & 32.44 & 30.67 & 80.02 & 73.18 & 29.08 & 29.76 & 26.92 & 26.67 & 31.40 & 43.30 & 36.37 \\
encord-team/ebind-points-vision & -- & 30.67 & -- & 73.18 & -- & 29.76 & -- & 26.67 & 31.40 & 43.30 & 35.91 \\
facebook/pe-av-base & 27.22 & 27.67 & 49.29 & 49.19 & 26.21 & 26.68 & 28.76 & 28.34 & 30.60 & 34.04 & 30.48 \\
facebook/pe-av-large & 29.78 & 28.89 & 41.04 & 42.56 & 26.25 & 26.40 & 28.18 & 27.01 & 29.20 & 33.84 & 30.74 \\
facebook/pe-av-small & 27.78 & 27.44 & 45.71 & 46.80 & 26.56 & 26.87 & 27.34 & 26.92 & 27.80 & 35.45 & 30.01 \\
jinaai/jina-embeddings-v5-omni-nano & 32.78 & 28.78 & 50.38 & 26.28 & 26.81 & 24.44 & 25.50 & 27.51 & 24.40 & 32.63 & 29.72 \\
jinaai/jina-embeddings-v5-omni-small & 38.11 & 22.67 & 47.77 & 28.77 & 26.43 & 21.51 & 25.92 & 25.17 & 21.20 & 25.18 & 28.30 \\
microsoft/xclip-base-patch16 & -- & 31.22 & -- & 65.91 & -- & 27.40 & -- & 26.42 & 28.40 & 42.60 & 33.85 \\
microsoft/xclip-base-patch32 & -- & 29.67 & -- & 64.71 & -- & 25.47 & -- & 26.76 & 30.80 & 41.49 & 34.69 \\
microsoft/xclip-large-patch14 & -- & 31.44 & -- & 70.79 & -- & 25.16 & -- & 25.75 & 29.20 & 43.20 & 34.83 \\
VLM2Vec/VLM2Vec-V2.0 & -- & 33.89 & -- & 72.31 & -- & 31.10 & -- & 27.26 & 55.80 & 44.91 & 41.72 \\
nvidia/omni-embed-nemotron-3b & 32.89 & 29.89 & 62.54 & 65.91 & 29.61 & 29.05 & 30.60 & 29.10 & 46.60 & 37.36 & 38.25 \\
zhibinlan/UME-R1-2B & -- & 36.89 & -- & 81.65 & -- & 29.05 & -- & 27.76 & 57.60 & 54.48 & 43.47 \\
zhibinlan/UME-R1-7B & -- & 37.56 & -- & 80.35 & -- & 30.70 & -- & 30.69 & 60.80 & 58.71 & 46.97 \\
\bottomrule
\end{tabular}%
}
\caption{\textbf{Question answering} (20 tasks, accuracy), part 1 of 2.}
\label{tab:results-qa-a}
\end{table*}
\begin{table*}[ht]
\centering
\small
\setlength{\tabcolsep}{4pt}
\renewcommand{\arraystretch}{1.15}
\resizebox{\linewidth}{!}{%
\begin{tabular}{lccccccccccc}
\toprule
\textbf{Model} & \makecell[c]{OmniVideoBench \\ audio + video} & \makecell[c]{OmniVideoBench \\ video} & \makecell[c]{PerceptionTest \\ audio + video} & \makecell[c]{PerceptionTest \\ video} & \makecell[c]{VideoMME-Short \\ audio + video} & \makecell[c]{VideoMME-Short \\ video} & \makecell[c]{WorldQA \\ audio + video} & \makecell[c]{WorldQA \\ video} & \makecell[c]{WS-1m \\ audio + video} & \makecell[c]{WS-1m \\ video} & \textbf{Avg.} \\
\midrule
BidirLM/BidirLM-Omni-2.5B-Embedding & 26.92 & 26.92 & 44.46 & 44.67 & 40.89 & 41.11 & -- & 38.37 & 32.00 & 30.95 & 40.36 \\
Haon-Chen/e5-omni-3B & 27.31 & 28.29 & 42.54 & 47.76 & 48.89 & 53.78 & 48.30 & 50.21 & 32.86 & 30.75 & 42.65 \\
Haon-Chen/e5-omni-7B & 27.11 & 26.52 & 54.48 & 53.30 & 54.22 & 55.44 & 53.87 & 54.17 & 38.01 & 36.68 & 46.22 \\
LCO-Embedding/LCO-Embedding-Omni-3B & \textbf{32.81} & \textbf{32.81} & 54.80 & 55.76 & 61.33 & 59.22 & 56.33 & 55.66 & 36.68 & 35.44 & 49.77 \\
LCO-Embedding/LCO-Embedding-Omni-7B & 31.63 & 30.84 & \textbf{59.91} & \textbf{61.51} & \textbf{64.67} & \textbf{62.22} & \textbf{59.93} & \textbf{57.77} & \textbf{41.07} & \textbf{38.97} & \textbf{51.81} \\
Qwen/Qwen2.5-Omni-3B & 23.58 & 24.36 & 34.54 & 35.50 & 22.78 & 23.33 & 25.67 & 27.89 & 23.21 & 23.88 & 25.19 \\
Qwen/Qwen2.5-Omni-7B & 25.74 & 25.15 & 36.57 & 33.58 & 22.11 & 21.11 & 18.12 & 18.36 & 26.27 & 24.93 & 23.91 \\
Qwen/Qwen3-VL-Embedding-2B & -- & 26.72 & -- & 46.05 & -- & 46.56 & -- & 43.12 & -- & 29.70 & 44.76 \\
Qwen/Qwen3-VL-Embedding-8B & -- & 27.31 & -- & 51.60 & -- & 52.22 & -- & 44.37 & -- & 31.52 & 46.09 \\
Tevatron/OmniEmbed-v0.1 & 26.52 & 25.93 & 42.00 & 40.41 & 40.22 & 40.11 & 34.23 & 37.85 & 27.98 & 26.74 & 37.53 \\
encord-team/ebind-audio-vision & 23.77 & 22.20 & 45.42 & 43.82 & 34.11 & 34.00 & 35.11 & 34.04 & 25.41 & 25.98 & 36.37 \\
encord-team/ebind-full & 23.77 & 22.20 & 45.42 & 43.82 & 34.11 & 34.00 & 35.11 & 34.04 & 25.41 & 25.98 & 36.37 \\
encord-team/ebind-points-vision & -- & 22.20 & -- & 43.82 & -- & 34.00 & -- & 34.04 & -- & 25.98 & 35.91 \\
facebook/pe-av-base & 24.95 & 25.15 & 36.67 & 36.35 & 28.44 & 28.00 & 25.94 & 28.17 & 24.07 & 23.78 & 30.48 \\
facebook/pe-av-large & 24.56 & 26.33 & 42.32 & 43.50 & 28.78 & 29.00 & 26.40 & 27.44 & 25.79 & 27.51 & 30.74 \\
facebook/pe-av-small & 23.77 & 23.58 & 33.79 & 33.69 & 29.11 & 30.11 & 27.38 & 28.81 & 25.41 & 25.98 & 30.01 \\
jinaai/jina-embeddings-v5-omni-nano & 23.97 & 26.33 & 41.58 & 41.05 & 27.89 & 26.56 & 27.47 & 30.79 & 26.46 & 22.92 & 29.72 \\
jinaai/jina-embeddings-v5-omni-small & 25.34 & 24.36 & 41.90 & 36.89 & 27.67 & 22.56 & 29.66 & 24.76 & 27.89 & 22.16 & 28.30 \\
microsoft/xclip-base-patch16 & -- & 22.40 & -- & 39.87 & -- & 30.33 & -- & 31.03 & -- & 26.74 & 33.85 \\
microsoft/xclip-base-patch32 & -- & 26.52 & -- & 42.11 & -- & 29.89 & -- & 34.65 & -- & 29.51 & 34.69 \\
microsoft/xclip-large-patch14 & -- & 24.16 & -- & 42.43 & -- & 31.33 & -- & 34.90 & -- & 24.74 & 34.83 \\
VLM2Vec/VLM2Vec-V2.0 & -- & 29.08 & -- & 42.75 & -- & 47.78 & -- & 40.07 & -- & 34.00 & 41.72 \\
nvidia/omni-embed-nemotron-3b & 26.13 & 25.15 & 40.73 & 41.26 & 40.33 & 40.22 & 48.20 & 48.33 & 29.51 & 31.52 & 38.25 \\
zhibinlan/UME-R1-2B & -- & 25.15 & -- & 43.60 & -- & 45.11 & -- & 44.40 & -- & 32.47 & 43.47 \\
zhibinlan/UME-R1-7B & -- & 26.72 & -- & 54.16 & -- & 56.22 & -- & 47.20 & -- & 33.62 & 46.97 \\
\bottomrule
\end{tabular}%
}
\caption{\textbf{Question answering} (20 tasks, accuracy), part 2 of 2.}
\label{tab:results-qa-b}
\end{table*}

\section{Data Contamination Analysis}
\label{appdx:contamination}

We follow \citet{chung2025maintainingmteblongterm} and report disclosed train-evaluation overlap per model in \autoref{tab:contamination}. Each model's declared training datasets (in its MTEB \texttt{ModelMeta}) are cross-referenced against MVEB task names. Of the 33 evaluated models, 11 report no overlap with MVEB tasks, 7 do not disclose at dataset granularity, and 15 have disclosed overlap on at least one task; the worst case is 65\%. Qwen2.5-Omni, Qwen3-VL-Embedding, Jina-Embeddings-v5-omni, and VLM2Vec-V2.0 do not disclose their training data at dataset granularity, so we cannot certify their MVEB results as zero-shot; their rows in \autoref{tab:contamination} carry \texttt{NA} in the Zero-shot~\% and \#~Overlap columns rather than the optimistic 100\% / 0 that would otherwise appear by default.

\begin{table*}[t]
    \centering
    \footnotesize
    \setlength{\tabcolsep}{6pt}
    \renewcommand{\arraystretch}{1.2}
    \begin{tabular}{l c c >{\raggedright\arraybackslash}p{0.50\textwidth}}
    \toprule
    \textbf{Model} & \textbf{Zero-shot \%} & \textbf{\# Overlap} & \textbf{Overlapping tasks} \\
    \midrule
    BidirLM-Omni-2.5B-Embedding           & 100\% & 0 & --- \\
    e5-omni-3B                            & 100\% & 0 & --- \\
    e5-omni-7B                            & 100\% & 0 & --- \\
    LCO-Embedding-Omni-3B                 & 100\% & 0 & --- \\
    LCO-Embedding-Omni-7B                 & 100\% & 0 & --- \\
    Qwen2.5-Omni-3B$^{\dagger}$           & NA    & NA & --- \\
    Qwen2.5-Omni-7B$^{\dagger}$           & NA    & NA & --- \\
    Qwen3-VL-Embedding-2B$^{\dagger}$     & NA    & NA & --- \\
    Qwen3-VL-Embedding-8B$^{\dagger}$     & NA    & NA & --- \\
    Tevatron OmniEmbed-v0.1               & 100\% & 0 & --- \\
    VLM2Vec-V2.0$^{\dagger}$              & NA    & NA & --- \\
    ebind-audio-vision                    &  83\% & 4 & AudioCapsAVVA2T, AudioCapsAVVT2A, VALOR32KT2VA, VGGSoundAVA2V \\
    ebind-full                            &  83\% & 4 & AudioCapsAVVA2T, AudioCapsAVVT2A, VALOR32KT2VA, VGGSoundAVA2V \\
    ebind-points-vision                   & 100\% & 0 & --- \\
    pe-av-base                            &  65\% & 8 & ActivityNetCaptionsT2V, AudioCapsAVVA2T, AudioCapsAVVT2A, HMDB51ZeroShot, Kinetics700VA, MSVDT2V, VALOR32KT2VA, VGGSoundAVA2V \\
    pe-av-large                           &  65\% & 8 & ActivityNetCaptionsT2V, AudioCapsAVVA2T, AudioCapsAVVT2A, HMDB51ZeroShot, Kinetics700VA, MSVDT2V, VALOR32KT2VA, VGGSoundAVA2V \\
    pe-av-small                           &  65\% & 8 & ActivityNetCaptionsT2V, AudioCapsAVVA2T, AudioCapsAVVT2A, HMDB51ZeroShot, Kinetics700VA, MSVDT2V, VALOR32KT2VA, VGGSoundAVA2V \\
    vjepa2-vitg-fpc32-384-diving48        &  96\% & 1 & Kinetics700VA \\
    vjepa2-vitg-fpc64-256                 &  96\% & 1 & Kinetics700VA \\
    vjepa2-vitg-fpc64-384                 &  96\% & 1 & Kinetics700VA \\
    vjepa2-vitg-fpc64-384-ssv2            &  96\% & 1 & Kinetics700VA \\
    vjepa2-vith-fpc64-256                 &  96\% & 1 & Kinetics700VA \\
    vjepa2-vitl-fpc16-256-ssv2            &  96\% & 1 & Kinetics700VA \\
    vjepa2-vitl-fpc32-256-diving48        &  96\% & 1 & Kinetics700VA \\
    vjepa2-vitl-fpc64-256                 &  96\% & 1 & Kinetics700VA \\
    jina-embeddings-v5-omni-nano$^{\dagger}$  & NA    & NA & --- \\
    jina-embeddings-v5-omni-small$^{\dagger}$ & NA    & NA & --- \\
    xclip-base-patch16                    & 100\% & 0 & --- \\
    xclip-base-patch32                    & 100\% & 0 & --- \\
    xclip-large-patch14                   & 100\% & 0 & --- \\
    omni-embed-nemotron-3b                & 100\% & 0 & --- \\
    UME-R1-2B                             &  65\% & 8 & BreakfastClassification, HMDB51ZeroShot, Kinetics700VA, MSVDT2V, UCF101VideoAudioClassification, VATEXV2A, VATEXVA2T, YouCook2T2VA \\
    UME-R1-7B                             &  65\% & 8 & BreakfastClassification, HMDB51ZeroShot, Kinetics700VA, MSVDT2V, UCF101VideoAudioClassification, VATEXV2A, VATEXVA2T, YouCook2T2VA \\
    \bottomrule
    \end{tabular}
    \caption{Disclosed train-evaluation overlap on MVEB. $^{\dagger}$Training data not disclosed at dataset granularity; we cannot certify zero-shot status for these models, so the Zero-shot~\% and \#~Overlap columns are reported as NA rather than 100\% / 0.}
    \label{tab:contamination}
\end{table*}
\section{Extended Analyses on Temporal Context Scaling (\# Frames)}
\label{appdx:frame-scaling}

\autoref{analyses_test_time_scaling_frames} discusses the aggregate effects
of test-time frame scaling. This appendix details the experimental setup
(task and model subsets, what varies, what is held fixed) and provides the
task- and model-level breakdowns that the main-text plot aggregates over.

\subsection{Experimental Setup}

\paragraph{What varies.} We sweep the number of uniformly sampled video
frames $N \in \{1, 8, 16, 32, 64\}$ at test time, for each (task, model)
pair in the subsets below.

\paragraph{What is held fixed.} Every other input the model sees is
unchanged from its default MVEB run: the audio track (when the model
accepts audio) is passed at the model's declared sampling rate and duration
cap, instructions are passed only for models that declare
\texttt{use\_instructions=True}, embeddings are taken from each model's
default output layer, and the per-task metric is the same one used in the
main leaderboard (nDCG@10 for retrieval, accuracy for classification and
QA, $v$-measure for clustering). No fine-tuning or pooling-strategy
changes are introduced.

\paragraph{Task subset.} We restrict the analysis to MVEB tasks whose
average source clip is longer than 15~seconds, where multi-frame sampling
has room to matter. \autoref{tab:frame-scaling-subsets} lists the seven
tasks (top block) and the five models (bottom block) used.

\paragraph{Model subset.} The five models span the major embedding
paradigms in our roster and bracket the parameter range (847M to 7B),
covering both architectures that integrate frames via temporal attention
and those that pool frame embeddings post-hoc. Models with fixed
clip-length training (X-CLIP, the 16-frame PE-AV variants) are excluded
because their positional embeddings are tied to a single budget.

\begin{table}[H]
    \centering
    \footnotesize
    \setlength{\tabcolsep}{6pt}
    \begin{tabular}{lll}
    \toprule
    \textbf{Entry} & \textbf{Identifier} & \textbf{Category / family} \\
    \midrule
    Task  & Breakfast                & Classification \\
    Task  & MusicAVQACLS             & Clustering \\
    Task  & OmniVideoBench           & QA \\
    Task  & WorldSense1Min           & QA \\
    Task  & MSRVTT T2V               & Retrieval \\
    Task  & VATEX T2VA               & Retrieval \\
    Task  & VATEX V2A                & Retrieval \\
    \midrule
    Model & LCO-Embedding-Omni-3B    & MLLM Embed \\
    Model & e5-omni-3B               & MLLM Embed \\
    Model & omni-embed-nemotron-3b   & MLLM Embed \\
    Model & ebind-full               & Multi-Bind \\
    Model & pe-av-small              & Aud-Vis Contr \\
    \bottomrule
    \end{tabular}
    \caption{Frame-scaling experimental subsets. The seven tasks are MVEB entries with average source-clip duration $>$~15~s; the five models span the MLLM-based, multimodal-binding, and audio-visual contrastive paradigms.}
    \label{tab:frame-scaling-subsets}
\end{table}

\subsection{Task-Level Performance Scaling}
\label{appdx:frame-scaling-task}

\begin{figure}[h]
    \centering
    \includegraphics[width=1.0\linewidth]{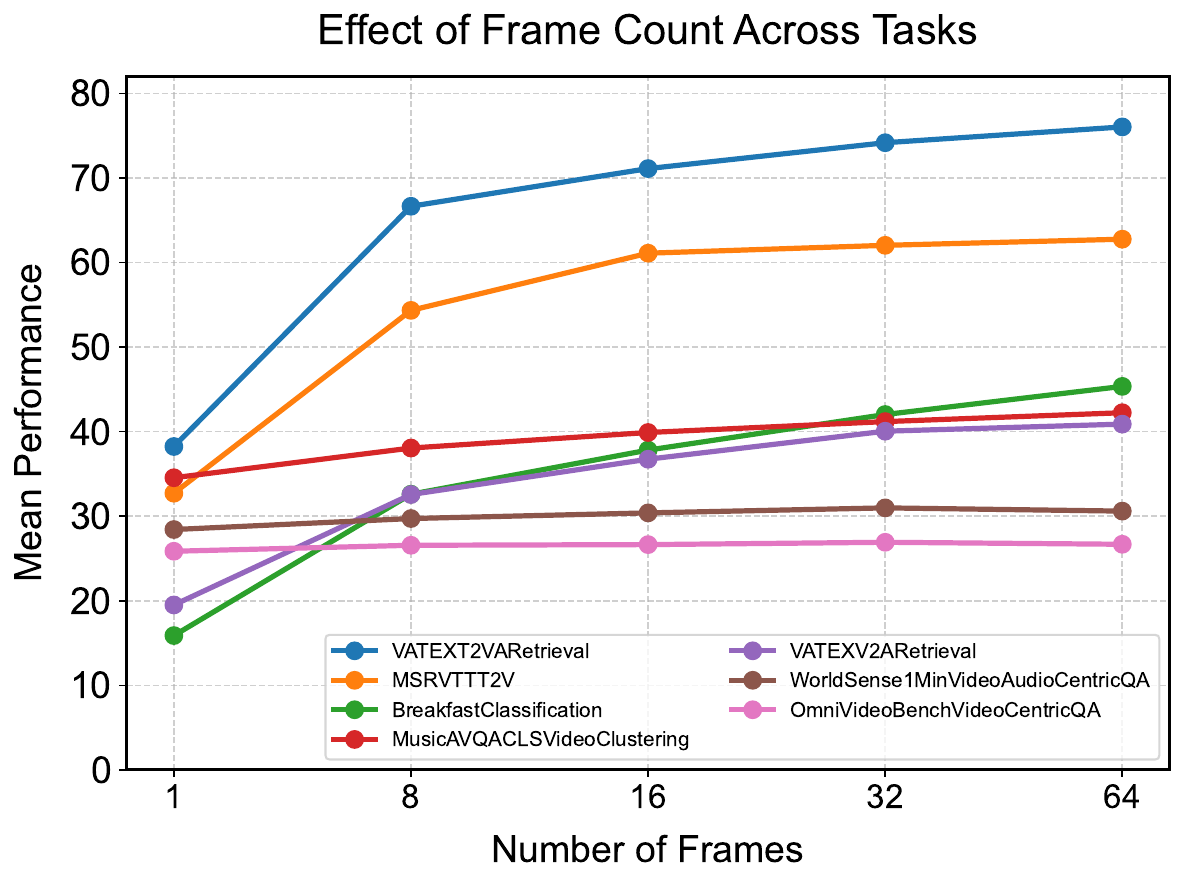}
    \caption{Per-task mean performance as a function of sampled frame count.}
    \label{fig:task_frame_scaling}
\end{figure}

The benefit of frame scaling depends sharply on the task. Tasks with
temporally rich content show steep gains: Breakfast improves from
15.88 ($N{=}1$) to 45.35 ($N{=}64$), and VATEX T2VA scales from 38.24 to
76.03. Other tasks barely move: OmniVideoBench QA shifts from 25.85
($N{=}1$) to 26.68 ($N{=}64$), and WorldSense1Min QA from 28.42 to 30.60.
The flat tasks share two properties: they either carry complementary
audio/text context that already saturates the model, or they admit a
strong single-frame solution (e.g., scene-level QA that can be answered
from a representative still). Per-task curves are in
\autoref{fig:task_frame_scaling}.

\subsection{Model-Level Performance Scaling}
\label{appdx:frame-scaling-model}

Scaling trajectories differ across architectures
(\autoref{fig:model_performance_scaling_with_overall}, main text).
\texttt{ebind-full} plateaus early: it reaches 45.62 at $N{=}8$ and only
46.68 at $N{=}64$, suggesting its frame-pooling integrator extracts most
of its signal from a small budget. The MLLM-based and audio-visual
contrastive models scale more smoothly through $N{=}64$;
\texttt{omni-embed-nemotron-3b} reaches 44.67 and \texttt{pe-av-small}
reaches 44.32 at the top of the sweep.
\section{Accessibility and Reproducibility}
\label{appendix:reproducibility}

While previous works \cite{enevoldsen2025mmteb, enevoldsen2024scandinavianembeddingbenchmarkscomprehensive, chung2025maintainingmteblongterm} drastically improve transparency, reproducibility and submission protocols for MTEB, we here describe further improvements to \textsc{mteb} that support reproducibility and long-term usability of the benchmark:

\paragraph{Experiment tracking:}
\textsc{mteb} now records the full set of evaluation run settings (model configuration and task parameters) alongside each result file.
Package version is tracked at the granularity of individual dataset subsets rather than per result file, enabling reliable merging of results obtained across different \textsc{mteb} versions. Thus reducing the need to rerun the full results when expanding existing evaluations to include new splits or languages, while maintain a full reproducible traceback. 
Results can be submitted to the community repository directly from the local cache via a standardised API.
Submissions can additionally be scoped to a named experiment, keeping ablations and non-standard runs separate from canonical model results, thus allow for experiments on embedding compression, reduced vector size and similar.

\paragraph{Runtime tracking and evaluation efficiency:}
Results are now written incrementally per dataset subset rather than once per completed task, so that only the interrupted subset must be re-run following a crash or preemption.
Runtime is also recorded per evaluation phase, supporting transparency into wall-clock costs alongside quality metrics.

\paragraph{Model metadata extensions:}
\textsc{ModelMeta} has been extended with an \texttt{active\_parameters} field, recording the number of parameters engaged during inference rather than the total parameter count, which is more relevant for efficiency comparisons especially for smaller models where the vocabulary can drastically influence the parameters count in the embeddings whithout influencing the runtime cost.
A \texttt{required\_dependencies} field allows model authors to declare additional package requirements at registration time, reducing friction and ensuring reproducibility when running third-party models.

\paragraph{Leaderboard transparency:}
A performance-over-time visualization traces the progression of state-of-the-art scores across benchmarks over the history of model submissions, enabling researchers to contextualize new results against the trajectory of the field.

\subsection{Per-model sampling configuration}
\label{appdx:sampling-config}

\autoref{tab:sampling-config} lists each model's declared video and audio sampling configuration. We honor these declared settings on every task rather than forcing a benchmark-wide budget; the rationale is in \S\ref{sec:eval-protocol}.

\begin{table}[h]
    \centering
    \footnotesize
    \setlength{\tabcolsep}{4pt}
    \renewcommand{\arraystretch}{1.1}
    \begin{tabular}{lll}
    \toprule
    \textbf{Model family} & \textbf{Video} & \textbf{Audio cap} \\
    \midrule
    \multicolumn{3}{l}{\textit{Variable-length: fps=2, max\_frames=64}} \\
    \midrule
    LCO-Embedding-Omni 3B/7B            & fps=2, max=64 & uncapped$^{\dagger}$ \\
    e5-omni 3B/7B                       & fps=2, max=64 & uncapped$^{\dagger}$ \\
    BidirLM-Omni-2.5B                   & fps=2, max=64 & 30 s \\
    OmniEmbed-v0.1 (Tevatron)           & fps=2, max=64 & uncapped$^{\dagger}$ \\
    OmniEmbed-Nemotron-3B               & fps=2, max=64 & $\sim$128 s \\
    Qwen2.5-Omni 3B/7B                  & fps=2, max=64 & 300 s \\
    Jina-Embeddings-v5-omni nano/small  & fps=2, max=64 & 30 s \\
    UME-R1 2B/7B                        & fps=2, max=64 & --- (no audio) \\
    VLM2Vec-V2.0                        & fps=2, max=64 & --- (no audio) \\
    Qwen3-VL-Embedding 2B/8B            & fps=2, max=64 & --- (no audio) \\
    PE-AV (variable-length)             & fps=2, max=64 & 30 s \\
    \midrule
    \multicolumn{3}{l}{\textit{Fixed-length: uniform sampling}} \\
    \midrule
    X-CLIP base-patch16/32, large-patch14 & 8 frames  & --- (no audio) \\
    eBind (full / audio-vision / points)  & 8 frames  & uncapped$^{\dagger}$ \\
    PE-AV (16-frame)$^{\ddagger}$         & 16 frames & 30 s \\
    V-JEPA-2 (fpc16/32/64 variants)       & 16/32/64 frames & --- (no audio) \\
    \bottomrule
    \end{tabular}
    \caption{Per-model video and audio sampling configuration. ``Variable-length'' models sample at \texttt{fps=2} with a hard cap of \texttt{max\_frames=64} to bound peak GPU memory; ``fixed-length'' models sample the exact number of frames their training pipeline expects. $^{\dagger}$Uncapped models delegate audio truncation to the model's own processor. $^{\ddagger}$The fixed-length 16-frame PE-AV variants are registered in the MTEB model registry but not among the 33 models evaluated on MVEB (\autoref{tab:all-models}); only the variable-length PE-AV checkpoints are evaluated. We list them here so the registry's model coverage is visible.}
    \label{tab:sampling-config}
\end{table}      
\section{Task correlation}
\label{appdx:task_correlation}

\begin{figure*}[t]
    \centering
    \includegraphics[width=\linewidth]{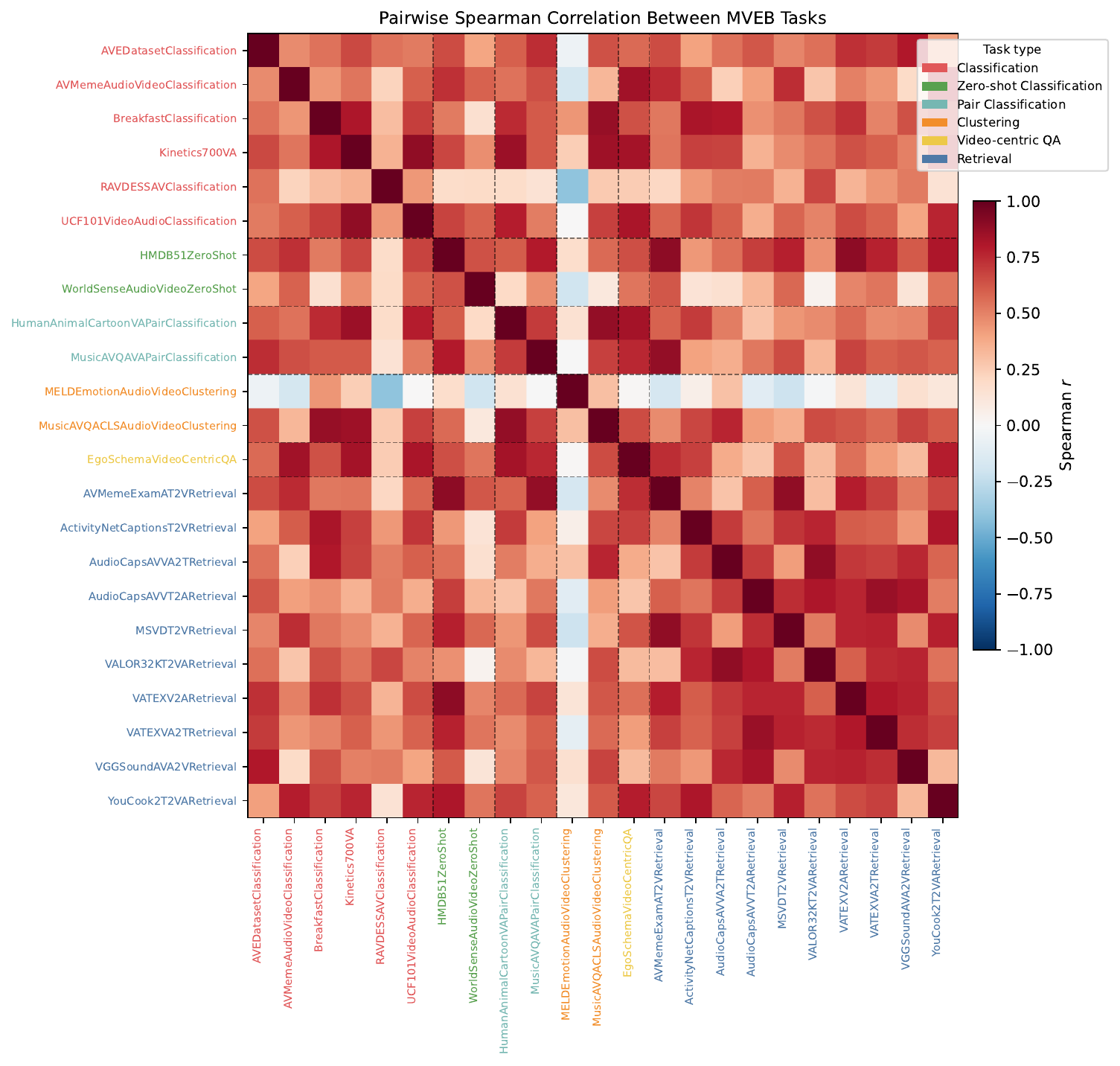}
    \caption{Pairwise rank correlation across MVEB+ tasks computed over the model roster. Block structure along the diagonal indicates clusters of tasks that rank models similarly (e.g., the eight retrieval directions cluster together; classification splits within a dataset family cluster together). This is the matrix from which the redundancy-removal step in \S\ref{sec:benchmark-construction} drops a task when $\rho > 0.85$ to a retained one.}
    \label{fig:task_correlation}
\end{figure*}

\begin{figure*}[t]
    \centering
    \includegraphics[width=\linewidth]{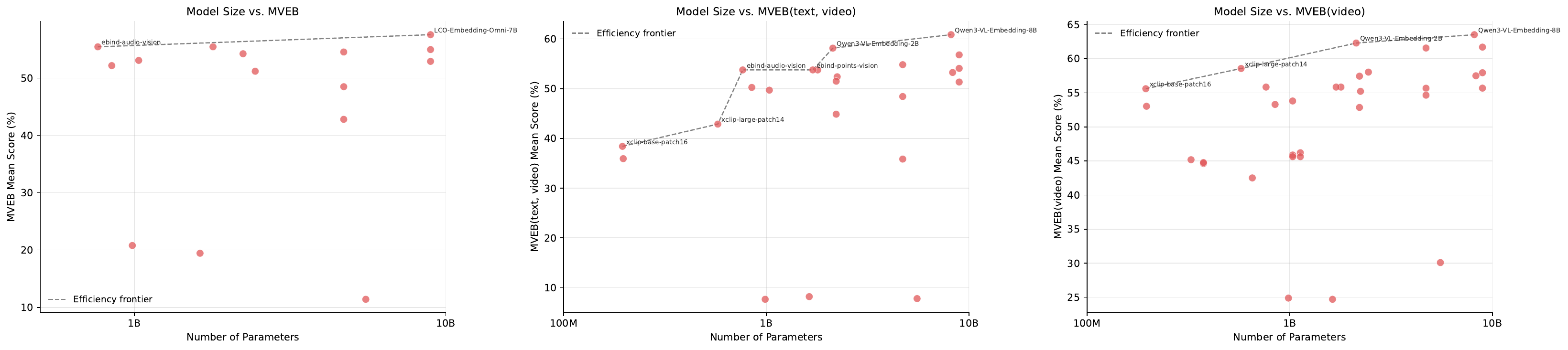}
    \caption{Parameter count vs.\ MVEB performance, broken out by task family. Complements the headline scaling view in \autoref{fig:param_vs_borda_rank} (which uses overall MVEB mean): per-family curves show where extra parameters convert into score (e.g., retrieval and QA) versus where they plateau (clustering, pair classification).}
    \label{fig:size_vs_performance}
\end{figure*}

\begin{figure*}[t]
    \centering
    \includegraphics[width=\linewidth]{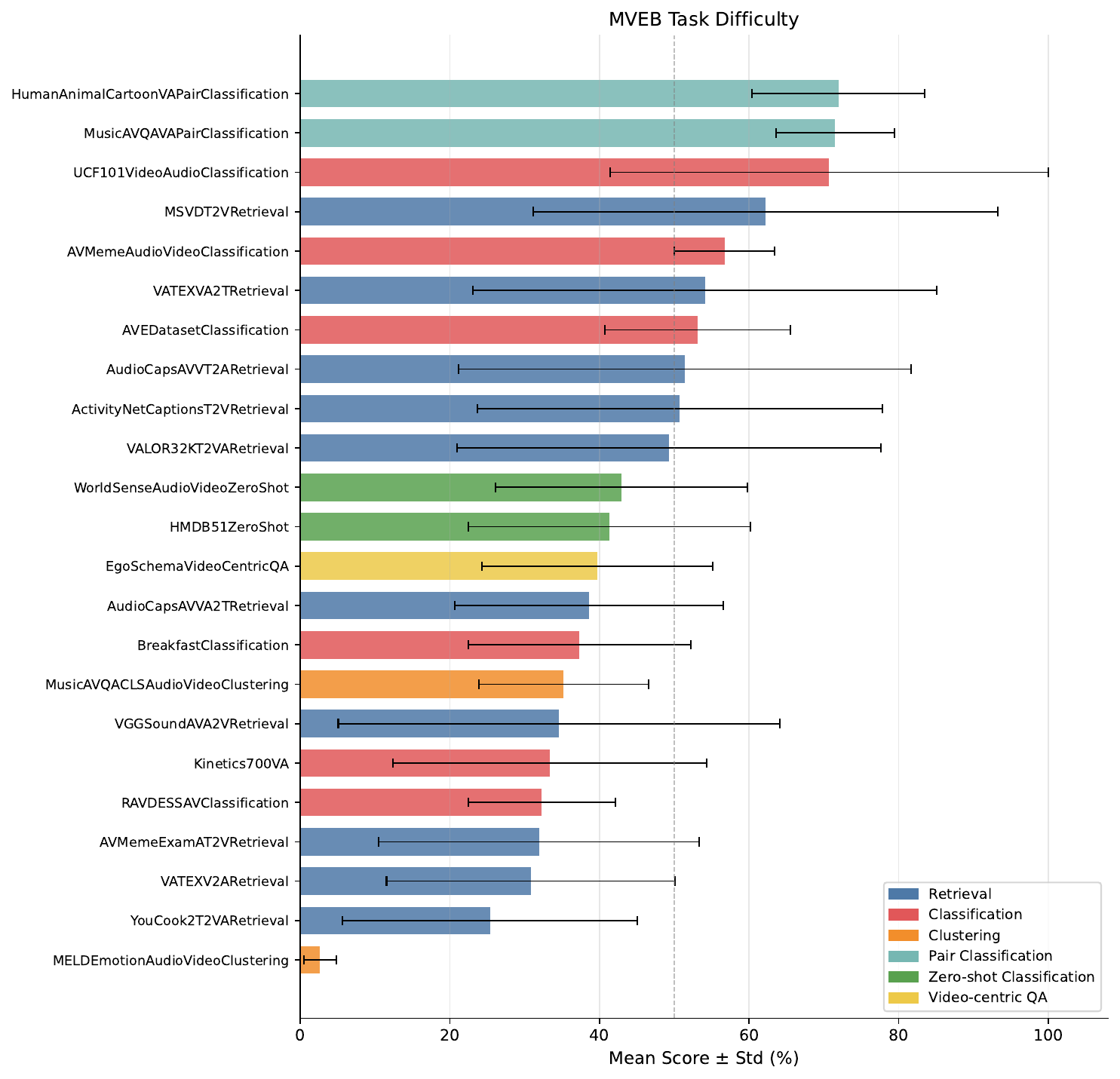}
    \caption{Per-task mean score across the model roster, sorted ascending. Low scores reflect a mix of intrinsic task difficulty and dataset label-quality issues; the emotion-recognition splits (MELD, RAVDESS) sit low primarily because of known label noise rather than embedding-quality limits (see \autoref{appdx:annotation-quality}).}
    \label{fig:task_difficulty}
\end{figure*}

This appendix groups three diagnostic plots over the MVEB task pool: how tasks correlate with one another across models (\autoref{fig:task_correlation}), how model parameter count relates to MVEB performance broken out by task (\autoref{fig:size_vs_performance}), and how the per-task mean score is distributed across the roster (\autoref{fig:task_difficulty}). Together they support the redundancy-pruning and dataset-coverage decisions described in \S\ref{sec:benchmark-construction}.

\section{Audio Contribution: Per-Dataset and Per-Model Breakdowns}
\label{appdx:audio-contribution}

This appendix expands the audio-contribution analysis from
\S\ref{sec:audio-contribution}. Table~\ref{tab:audio_per_dataset}
reports the per-dataset audio delta sorted within each annotation
group; Table~\ref{tab:audio_cross_modal} reports cross-modal retrieval
alignment for the datasets that have both v2a and a2v task variants;
Table~\ref{tab:audio_per_model} reports the per-model mean audio delta
across all 48 paired task groups.

\begin{table*}[p]
\small
\centering
\caption{Per-dataset audio delta ($\Delta = \text{score}_{\text{va}} - \text{score}_{\text{v}}$), sorted by mean delta within each annotation group. AV-grounded datasets had labels produced from both audio and visual content; V-grounded datasets had labels produced from visuals alone, even though the source clips often carry audio (see \autoref{tab:av-provenance}). $N$ counts model--dataset pairs with results for both task variants.}
\label{tab:audio_per_dataset}
\begin{tabular}{llrrr}
\toprule
Dataset & Task type & $\bar\Delta$ & $\sigma$ & $N$ \\
\midrule
\multicolumn{5}{l}{\textit{AV-grounded datasets}} \\
\midrule
AudioCaps-AV              & Retrieval          & $+0.151$ & $0.093$ & $14$ \\
AVMeme-Exam               & Classification     & $+0.126$ & $0.041$ & $14$ \\
AVMeme-Exam               & Retrieval          & $+0.063$ & $0.090$ & $14$ \\
VGGSound                  & Classification     & $+0.062$ & $0.031$ & $14$ \\
YouCook2                  & Retrieval          & $+0.056$ & $0.107$ & $14$ \\
AVMeme-Exam               & ZS Classification  & $+0.040$ & $0.024$ & $14$ \\
MELD                      & Classification     & $+0.036$ & $0.026$ & $14$ \\
RAVDESS-AV                & ZS Classification  & $+0.030$ & $0.061$ & $14$ \\
AVE-Dataset               & Clustering         & $+0.029$ & $0.023$ & $14$ \\
AVE-Dataset               & Classification     & $+0.025$ & $0.038$ & $14$ \\
Daily-Omni                & QA                 & $+0.016$ & $0.022$ & $14$ \\
AVMeme-Exam               & QA                 & $+0.014$ & $0.016$ & $14$ \\
WorldSense                & Clustering         & $+0.013$ & $0.054$ & $14$ \\
MUSIC-AVQA                & Clustering         & $+0.008$ & $0.039$ & $14$ \\
AVQA                      & QA                 & $+0.006$ & $0.032$ & $14$ \\
VGGSound                  & ZS Classification  & $+0.004$ & $0.052$ & $14$ \\
WorldSense                & QA                 & $+0.003$ & $0.013$ & $14$ \\
OmniVideoBench            & QA                 & $+0.002$ & $0.009$ & $14$ \\
WorldSense                & Classification     & $+0.001$ & $0.035$ & $14$ \\
AV-SpeakerBench           & QA                 & $-0.000$ & $0.006$ & $14$ \\
MUSIC-AVQA                & Classification     & $-0.000$ & $0.030$ & $14$ \\
PerceptionTest            & QA                 & $-0.001$ & $0.020$ & $14$ \\
WorldSense                & ZS Classification  & $-0.001$ & $0.071$ & $14$ \\
Video-MME                 & QA                 & $-0.001$ & $0.017$ & $14$ \\
RAVDESS-AV                & Classification     & $-0.003$ & $0.078$ & $14$ \\
WorldQA                   & QA                 & $-0.006$ & $0.016$ & $13$ \\
VALOR-32K                 & Retrieval          & $-0.008$ & $0.071$ & $14$ \\
MUSIC-AVQA                & ZS Classification  & $-0.011$ & $0.024$ & $14$ \\
Shot2Story20K             & Retrieval          & $-0.012$ & $0.040$ & $14$ \\
MELD                      & ZS Classification  & $-0.012$ & $0.104$ & $14$ \\
AVE-Dataset               & ZS Classification  & $-0.014$ & $0.064$ & $14$ \\
RAVDESS-AV                & Clustering         & $-0.014$ & $0.062$ & $14$ \\
MELD                      & Clustering         & $-0.019$ & $0.053$ & $14$ \\
VGGSound-AV-Ret.          & Retrieval          & $-0.035$ & $0.087$ & $14$ \\
\midrule
\multicolumn{5}{l}{\textit{V-grounded datasets}} \\
\midrule
Human-Animal-Cartoon      & ZS Classification  & $-0.010$ & $0.024$ & $14$ \\
Human-Animal-Cartoon      & Classification     & $-0.027$ & $0.043$ & $14$ \\
MSR-VTT                   & Retrieval          & $-0.030$ & $0.057$ & $14$ \\
VATEX                     & Retrieval          & $-0.032$ & $0.044$ & $14$ \\
UCF101-51VA               & Clustering         & $-0.033$ & $0.035$ & $14$ \\
Kinetics-400              & ZS Classification  & $-0.034$ & $0.039$ & $14$ \\
UCF101-51VA               & Classification     & $-0.037$ & $0.032$ & $14$ \\
Kinetics-600              & ZS Classification  & $-0.037$ & $0.045$ & $14$ \\
Kinetics-700-2020         & ZS Classification  & $-0.039$ & $0.045$ & $14$ \\
Kinetics-600              & Classification     & $-0.059$ & $0.070$ & $14$ \\
Kinetics-400              & Classification     & $-0.065$ & $0.069$ & $14$ \\
Kinetics-700-2020         & Classification     & $-0.070$ & $0.098$ & $14$ \\
DiDeMo                    & Retrieval          & $-0.072$ & $0.049$ & $14$ \\
Panda-70M                 & Retrieval          & $-0.102$ & $0.089$ & $13$ \\
\bottomrule
\end{tabular}
\end{table*}

\begin{table}[t]
\centering
\caption{Cross-modal retrieval scores ($\bar{s}$) averaged across 14 models, for datasets that have v2a (video-to-audio) and a2v (audio-to-video) retrieval task variants. High scores indicate that video and audio embeddings are well-aligned, making explicit audio information more likely redundant. \textit{YouCook2} has the lowest cross-modal alignment and gains $\Delta = +0.056$ in the paired v/va comparison, consistent with spoken narration providing information not recoverable from visual frames alone. By contrast, \textit{VGGSound-AV-Ret.} has high alignment yet yields $\Delta = -0.035$, indicating the video embedding already captures the acoustic structure of these clips.}
\label{tab:audio_cross_modal}
\begin{tabular}{llrrr}
\toprule
Dataset & Dir. & $\bar{s}$ & $\sigma$ & $N$ \\
\midrule
AVMeme-Exam        & a2v & $0.259$ & $0.130$ & $14$ \\
AVMeme-Exam        & v2a & $0.249$ & $0.150$ & $14$ \\
AudioCaps-AV       & a2v & $0.381$ & $0.296$ & $14$ \\
AudioCaps-AV       & v2a & $0.410$ & $0.271$ & $14$ \\
DiDeMo             & a2v & $0.285$ & $0.211$ & $14$ \\
DiDeMo             & v2a & $0.304$ & $0.190$ & $14$ \\
MSR-VTT            & a2v & $0.350$ & $0.160$ & $14$ \\
MSR-VTT            & v2a & $0.332$ & $0.153$ & $14$ \\
Shot2Story20K      & a2v & $0.449$ & $0.327$ & $14$ \\
Shot2Story20K      & v2a & $0.404$ & $0.290$ & $14$ \\
VALOR-32K          & a2v & $0.246$ & $0.201$ & $14$ \\
VALOR-32K          & v2a & $0.250$ & $0.165$ & $14$ \\
VATEX              & a2v & $0.394$ & $0.204$ & $14$ \\
VATEX              & v2a & $0.351$ & $0.164$ & $14$ \\
VGGSound-AV-Ret.   & a2v & $0.395$ & $0.283$ & $14$ \\
VGGSound-AV-Ret.   & v2a & $0.413$ & $0.251$ & $14$ \\
YouCook2           & a2v & $0.162$ & $0.121$ & $14$ \\
YouCook2           & v2a & $0.133$ & $0.113$ & $14$ \\
\bottomrule
\end{tabular}
\end{table}

\begin{table}[t]
\centering
\caption{Per-model mean audio delta ($\bar\Delta$) across 48 paired task groups for the 14 audio-capable models in our roster. A positive $\bar\Delta$ indicates the model benefits on average from the audio track; a negative value indicates the audio channel hurts performance. $N$ counts dataset--task pairs evaluated.}
\label{tab:audio_per_model}
\begin{tabular}{lrrr}
\toprule
Model & $\bar\Delta$ & $\sigma$ & $N$ \\
\midrule
BidirLM-Omni-2.5B-Embedding    & $+0.023$ & $0.036$ & $46$ \\
omni-embed-nemotron-3b         & $+0.020$ & $0.042$ & $48$ \\
e5-omni-7B                     & $+0.019$ & $0.052$ & $48$ \\
Qwen2.5-Omni-3B                & $+0.013$ & $0.042$ & $48$ \\
pe-av-small                    & $+0.008$ & $0.062$ & $48$ \\
Qwen2.5-Omni-7B                & $+0.007$ & $0.063$ & $48$ \\
pe-av-base                     & $+0.006$ & $0.058$ & $48$ \\
LCO-Embedding-Omni-7B          & $+0.002$ & $0.082$ & $48$ \\
LCO-Embedding-Omni-3B          & $-0.002$ & $0.080$ & $48$ \\
OmniEmbed-v0.1                 & $-0.016$ & $0.074$ & $48$ \\
pe-av-large                    & $-0.021$ & $0.069$ & $48$ \\
ebind-audio-vision             & $-0.023$ & $0.072$ & $48$ \\
ebind-full                     & $-0.023$ & $0.072$ & $48$ \\
e5-omni-3B                     & $-0.039$ & $0.099$ & $48$ \\
\bottomrule
\end{tabular}
\end{table}

\section{Cross-Modal Retrieval Direction Structure: Extended Analysis}
\label{appdx:retrieval-correlation}

This appendix expands the cross-modal retrieval correlation analysis in
\S\ref{sec:retrieval-direction-structure}.
\autoref{tab:retrieval-pairwise} reports every pairwise Spearman
correlation across the eight retrieval directions, computed over the
16 audio-capable models in our roster.

\begin{table*}[t]
\centering
\small
\caption{Pairwise Spearman correlation $\rho$ across the eight retrieval directions, computed across the audio-capable models in our roster. Each cell reports $\rho$ over the models that have results in both directions. Higher $\rho$ means models that rank well on one direction also rank well on the other (the two directions measure overlapping capabilities). Bold cells: $|\rho| > 0.85$.}
\label{tab:retrieval-pairwise}
\begin{tabular}{lcccccccc}
\toprule
 & $T\!\to\!V$ & $V\!\to\!T$ & $A\!\to\!V$ & $V\!\to\!A$ & $AT\!\to\!V$ & $VA\!\to\!T$ & $VT\!\to\!A$ & $T\!\to\!VA$ \\
\midrule
$T\!\to\!V$ & $\,1.00$ & $\,0.74$ & $\,0.50$ & $\,0.59$ & $\,0.53$ & $\,0.70$ & \textbf{$\,0.88$} & \textbf{$\,0.92$} \\
$V\!\to\!T$ &  & $\,1.00$ & $\,0.71$ & \textbf{$\,0.86$} & $\,0.81$ & \textbf{$\,0.96$} & $\,0.82$ & $\,0.66$ \\
$A\!\to\!V$ &  &  & $\,1.00$ & \textbf{$\,0.87$} & \textbf{$\,0.90$} & $\,0.63$ & $\,0.65$ & $\,0.38$ \\
$V\!\to\!A$ &  &  &  & $\,1.00$ & \textbf{$\,0.94$} & $\,0.75$ & $\,0.74$ & $\,0.43$ \\
$AT\!\to\!V$ &  &  &  &  & $\,1.00$ & $\,0.72$ & $\,0.64$ & $\,0.39$ \\
$VA\!\to\!T$ &  &  &  &  &  & $\,1.00$ & $\,0.73$ & $\,0.68$ \\
$VT\!\to\!A$ &  &  &  &  &  &  & $\,1.00$ & $\,0.77$ \\
$T\!\to\!VA$ &  &  &  &  &  &  &  & $\,1.00$ \\
\bottomrule
\end{tabular}
\end{table*}

\paragraph{Three latent capability axes.} The 28 pairwise correlations
resolve into three latent capability groups:

\begin{description}
    \item[Text-as-query] $T\!\to\!V$, $T\!\to\!VA$, $VT\!\to\!A$.
        Internal $\rho \in [0.77, 0.92]$. These directions all start
        from a textual query and retrieve into some non-text modality;
        the strongest pair is $T\!\to\!V$ vs.\ $T\!\to\!VA$
        ($\rho = 0.92$), which differ only in whether the target
        includes audio.
    \item[Audio-as-conditioning-into-video] $A\!\to\!V$, $AT\!\to\!V$,
        $V\!\to\!A$. Internal $\rho \in [0.87, 0.94]$. The directions
        that involve audio paired with video as either query or target;
        $V\!\to\!A$ correlates as tightly with $AT\!\to\!V$
        ($\rho = 0.94$) as it does with $A\!\to\!V$ ($\rho = 0.87$),
        suggesting these three measure essentially the same capability
        of an audio-video joint encoder.
    \item[Text-as-target] $V\!\to\!T$, $VA\!\to\!T$. A tight pair at
        $\rho = 0.96$; whether the query includes audio or not, the
        model's text-target ranking is nearly invariant.
\end{description}

\paragraph{What this means for benchmarking.} The directions that
\textit{cross} group boundaries are the most decoupled: $T\!\to\!VA$
vs.\ $A\!\to\!V$ ($\rho = 0.38$) and $T\!\to\!VA$ vs.\ $AT\!\to\!V$
($\rho = 0.39$) are the lowest pairs in the table. Both contrasts
involve audio: in one case $T\!\to\!VA$ has audio as part of the
\textit{target} (joined with video), while $A\!\to\!V$/$AT\!\to\!V$
have audio as part of the \textit{query}. Models that handle one of
these well are not guaranteed to handle the other; this is consistent
with the audio-track analysis in \S\ref{sec:audio-contribution} that audio
contributes a genuinely distinct signal whose encoding is
implementation-specific.

This decoupling is the contrast that
MMEB-V3~\cite{huang2026mmebv3measuringperformancegaps} cannot expose:
by treating audio strictly as a separate retrieval modality with only
$A \to V$ and $V \to A$ directions and never paired audio$+$video
inputs, it does not include $T \to VA$ at all, and so the third
capability group (audio as part of a joint retrieval target) is
invisible.

\paragraph{Implication for the headline leaderboard.} The
within-group correlations above $0.85$ mean a reader could
collapse the eight directions into three target-grouped scores
(text-target, video-target, audio-target) without losing rank
information. We keep all eight in the per-direction appendix tables
(\autoref{appdx:results-retrieval}) for transparency, but the MVEB
\textbf{Retr} column already mean-pools them into a single score that
aligns with the dominant capability axis.
\section{MVEB(text, video): Extended Analysis}
\label{appdx:text-video-leaderboard}

This appendix expands the discussion of MVEB(text, video) from
\S\ref{sec:key-findings}. The full 25-model leaderboard is in
\autoref{tab:mveb-text-video-results}
(Appendix~\ref{appdx:results-scope-variants}); we report three
observations here.

\paragraph{Breadth of the Qwen3-VL family's lead.}
Qwen3-VL-Embedding-8B and -2B~\cite{qwen3vlembedding} take ranks 1
and 2 on MVEB(text, video) (60.9 and 58.1 mean). On the per-category
breakdown, Qwen3-VL-Embedding-8B leads retrieval (69.2), classification
(57.5), and zero-shot classification (59.2); Qwen3-VL-Embedding-2B
leads clustering (24.9) and pair classification (88.3). LCO-Embedding-Omni-3B
retains QA (32.8). Qwen3-VL-Embedding-2B at 58.1 mean already exceeds
LCO-Embedding-Omni-7B (56.8) at roughly one-quarter the parameter
count. We report these numbers; we do not attempt to attribute the gap
to a specific cause, since the Qwen3-VL checkpoints differ from the
omni embedders in training recipe, training-data composition, and
declared input modalities simultaneously.

\paragraph{What MVEB(text, video) is not.} MVEB(text, video) excludes
the audio-conditioned retrieval directions ($A\to V$, $V\to A$,
$AT\to V$, $VT\to A$, $T\to VA$, $VA\to T$) and the
\texttt{audio+video} task variants by construction; a high rank here
is not evidence that a model would generalize to audio-bearing tasks,
and the headline MVEB leaderboard remains the appropriate surface for
audio-capable models. We expose MVEB(text, video) so that models that
do not accept the full audio+video+text input surface can still be
evaluated on a principled subset of the same task pool.

\paragraph{Relation to MMEB-V2.}
MMEB-V2~\cite{meng2025vlm2vecv2} is the closest existing text-video
evaluation surface: 18 video tasks across retrieval, classification,
VQA, grounding, and instance retrieval, evaluating text-video models
in isolation. The structural difference with MVEB(text, video) is that
the latter is the text-video projection of the same task pool we use
for the headline MVEB and the audio-only-encoder variant, so the same
model can be compared on the same tasks with and without an audio
path. We do not claim a head-to-head comparison of model rankings
between MMEB-V2 and MVEB(text, video) here, since the two benchmarks
draw their tasks from different pools.
\section{Annotation Quality Notes}
\label{appdx:annotation-quality}

We expand on the annotation-quality limitation introduced in the
Limitations section. The most affected MVEB source datasets are the
emotion-recognition splits; similar issues likely apply elsewhere but
emotion datasets surface them most visibly.

\paragraph{MELD.} MELD assigns a single discrete emotion to short
utterance clips drawn from a sitcom. Two annotation patterns recur:
(i) multi-sentiment utterances are squashed to one label, since a
single clip can carry a surprise reaction, a brief rebuke, and a
laugh track all in a few seconds, yet only one emotion tag is
assigned; (ii) sitcom-specific context cues (laugh tracks, sarcastic
delivery, the listener's reaction shot) can pull the gold label
away from the literal text content. As an illustrative example of
the first pattern, an utterance that contains an initial exclamation
followed by a mild verbal complaint may carry an \textit{anger} label
even though a reasonable reader would split the same clip into
several distinct affect segments. The single-label formulation is a
property of the dataset, not the MVEB task wrapper, and we have not
re-annotated for this paper.

\paragraph{RAVDESS.} RAVDESS uses scripted emotion-acting performed
by 24 professional actors at two intensity levels. Compound or
transition states (e.g., a clip whose first half is neutral and
second half is angry) are not captured by the single-emotion label,
and intensity is a separate axis rather than a per-clip secondary
label.

\paragraph{AVE-Dataset.} AVE-Dataset annotates audio-visual events
with class labels and bounding times. Some events co-occur within a
clip but only the dominant event is annotated, which introduces
unavoidable noise in clip-level classification tasks that operate
on the whole clip rather than on the bounded event spans.

\paragraph{Why we keep the as-released labels.} (a) These datasets
are widely used and re-annotating them would break score
comparability with the substantial body of prior published results.
(b) MVEB's goal is to measure embedding quality on the field's
standard evaluation surfaces, including their imperfections; an
embedding model that perfectly fits a noisy single-label gold
standard may be over-fitting to the annotation noise itself.

\paragraph{Mitigation paths for the community.} Three directions are
tractable: (i) soft-label or multi-label re-releases that publish
per-clip distributions over labels rather than a single argmax, with
evaluation tasks that consume the distribution; (ii) human-model
agreement audits in the style of
HUME~\cite{assadi2025humemeasuringhumanmodelperformance} that sample
clips where current models disagree with the as-released label and
check which side a fresh human annotator takes; (iii) targeted
re-annotation on the most affected splits, with the as-released
labels retained under a versioned task so prior results remain
reproducible.

\end{document}